\documentclass{article}

\usepackage[preprint]{neurips_2023}

\usepackage{amsmath,amsfonts,bm}

\def\eqref#1{equation~\ref{#1}}

\def\1{\bm{1}}

\def\va{{\bm{a}}}
\def\vb{{\bm{b}}}

\def\vx{{\bm{x}}}

\def\mA{{\bm{A}}}

\def\mE{{\bm{E}}}

\def\mH{{\bm{H}}}

\def\mK{{\bm{K}}}

\def\mP{{\bm{P}}}
\def\mQ{{\bm{Q}}}

\def\mV{{\bm{V}}}
\def\mW{{\bm{W}}}

\DeclareMathAlphabet{\mathsfit}{\encodingdefault}{\sfdefault}{m}{sl}
\SetMathAlphabet{\mathsfit}{bold}{\encodingdefault}{\sfdefault}{bx}{n}

\newcommand{\R}{\mathbb{R}}

\usepackage{hyperref}
\usepackage{url}

\usepackage{booktabs} 
\usepackage{adjustbox}
\usepackage{graphicx}
\usepackage{subcaption}
\usepackage{multirow}
\usepackage{tabularx}
\usepackage{booktabs}
\usepackage{svg}
\usepackage{natbib}

\usepackage{tabularx}
\usepackage{diagbox}
\usepackage{booktabs}
\usepackage{tikz}
\newcommand{\tikzxmark}{%
\textcolor{red}{
\tikz[scale=0.23] {
    \draw[line width=0.7,line cap=round] (0,0) to [bend left=6] (1,1);
    \draw[line width=0.7,line cap=round] (0.2,0.95) to [bend right=3] (0.8,0.05);
}}}
\newcommand{\tikzcmark}{%
\textcolor{ao}{
\tikz[scale=0.23] {
    \draw[line width=0.7,line cap=round] (0.25,0) to [bend left=10] (1,1);
    \draw[line width=0.8,line cap=round] (0,0.35) to [bend right=1] (0.23,0);
}}}
\usepackage[utf8]{inputenc} 
\usepackage[T1]{fontenc}    
\usepackage{hyperref}       
\usepackage{url}            
\usepackage{booktabs}       
\usepackage{amsfonts}       
\usepackage{nicefrac}       
\usepackage{microtype}      
\usepackage{xcolor}         

\usepackage{array}
\usepackage{makecell}
\usepackage{multicol}

\usepackage{microtype}
\usepackage{xcolor}
\usepackage{amsfonts} 
\usepackage[ruled,vlined]{algorithm2e}
\usepackage{algpseudocode}

\definecolor{ao}{rgb}{0.0, 0.5, 0.0}
\definecolor{awesome}{rgb}{0.9, 0.17, 0.31}
\definecolor{ameth}{rgb}{0.6, 0.4, 0.8}
\definecolor{_green}{rgb}{0.0, 0.4, 0.65}
\usepackage[utf8]{inputenc}

\title{Ahead-of-Time P-Tuning}

\author{Daniil Gavrilov, Nikita Balagansky \\
    Tinkoff \\
  \texttt{d.gavrilov@tinkoff.ai}, \texttt{n.n.balaganskiy@tinkoff.ai}}

\begin{document}

\maketitle

\begin{abstract}

In this paper, we propose Ahead-of-Time (AoT) P-Tuning, a novel parameter-efficient fine-tuning method for pre-trained Language Models (LMs) that adds input-dependent bias before each Transformer layer. We evaluate AoT P-Tuning on GLUE and SuperGLUE benchmarking datasets using RoBERTa and DeBERTa models, showing that it outperforms BitFit and is comparable or better than other baseline methods for efficient fine-tuning. Additionally, we assess the inference overhead of AoT P-Tuning and demonstrate that it introduces negligible overhead compared to established baseline methods. Our method enables multi-task inference with a single backbone LM, making it a practical solution for real-world applications.
\end{abstract}

\section{Introduction}
As pre-trained Language Models (LMs) grow in size \citep{gpt, bert, t5}, developing new methods for handling these large models becomes increasingly important. One potential solution is parameter-efficient fine-tuning, where only a subset of the model's parameters are optimized from scratch or using existing weights \citep{v1, v2, adapters, lora, bitfit, google-pt}.

Typically, the success of these parameter-efficient fine-tuning methods is judged based on the number of optimized parameters and the model's performance on downstream tasks. However, in this paper, we also consider two additional factors: the inference overhead that results from fine-tuning and the model's ability to perform multi-task inference. The latter refers to using a single LM to handle multiple tasks during inference, which is advantageous for real-world applications \citep{google-pt}.

Most current methods, such as P-Tuning \citep{v1, v2, google-pt}, Adapters \citep{adapters}, and LoRA \citep{lora}, have a trade-off between introducing inference overhead and the ability to perform multi-task inference. BitFit \citep{bitfit} does not introduce any overhead and can handle multi-task inference, but its performance is lacking compared to other methods.

In this paper, we propose a simple parameter-efficient fine-tuning method for LMs called Ahead-of-Time (AoT) P-Tuning. This method involves adding input-dependent bias before each Transformer layer \citep{transformer}. Furthermore, AoT P-Tuning can be used in multi-task inference setups with a single backbone LM.

Our contributions in this paper can be summarized as follows:
\begin{enumerate}
\item{We introduce Ahead-of-Time (AoT) P-Tuning, a parameter-efficient fine-tuning method that adds input-dependent bias before each Transformer layer.}
\item{We test the proposed method on GLUE and SuperGLUE benchmarking datasets \citep{glue, superglue} using RoBERTa \citep{roberta} and DeBERTa \citep{deberta} models. Our experiments show that AoT P-Tuning outperforms BitFit and is comparable or better than other baseline methods for efficient fine-tuning.}
\item{We measure the inference overhead of AoT P-Tuning and demonstrate that it introduces negligible overhead compared to established baseline methods.}
\end{enumerate}

\begin{figure*}
\centering
  \includegraphics[width=0.8\textwidth]{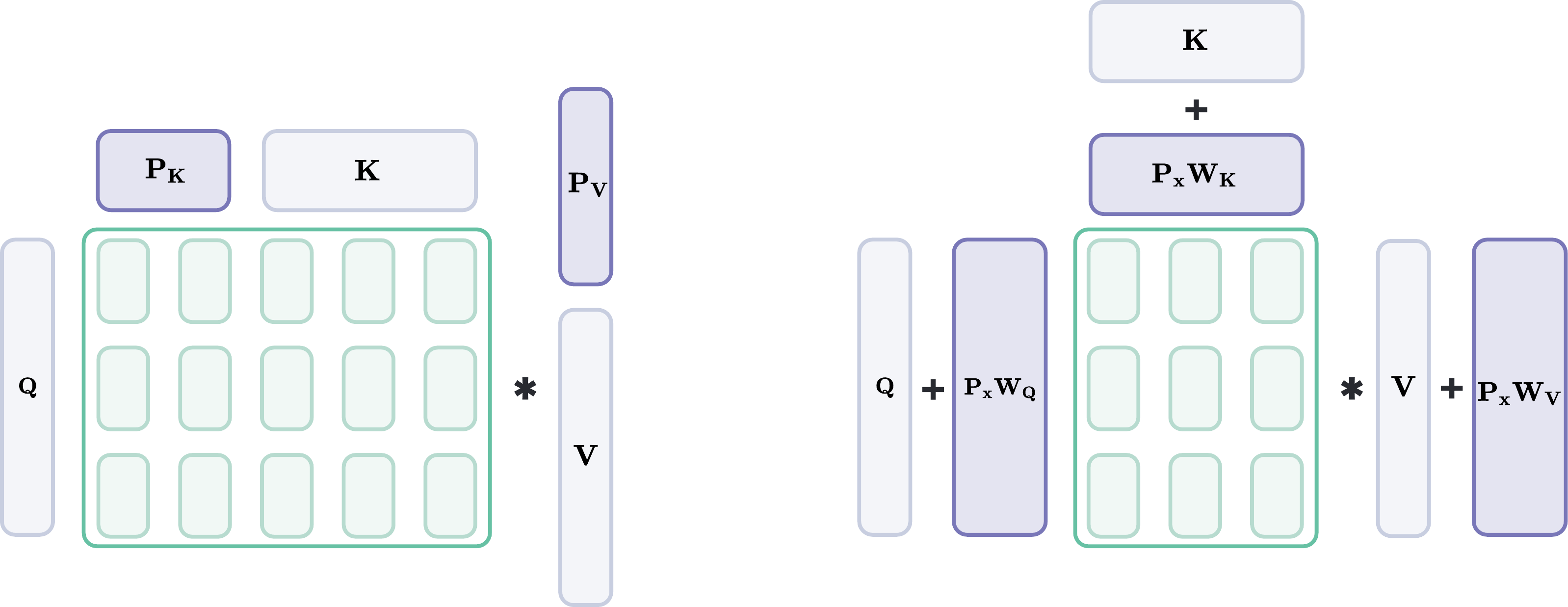}
  \caption{Schematic comparison of P-Tuning v2 (left), and AoT P-Tuning (right). Since the sequence length is not increased, AoT P-Tuning takes significantly less time to evaluate, only requiring the overhead of adding biases to the input sequence (See Section \ref{section-speed-results} for experiments with inference speed).}
  \label{figure-schematic}
\end{figure*}

\begin{table*}
\small
\centering
\begin{tabular}{r|c|c|c}
\toprule
Method         & Parameter Efficient & Zero-Cost & Multi-Task Inference \\
\toprule
Fine-Tuning    & $\tikzxmark$              & $\tikzcmark$                            & $\tikzxmark$               \\
\midrule
LoRA           & $\tikzcmark$              & $\tikzxmark$                            & $\tikzcmark$               \\
LoRA Fused           & $\tikzcmark$              & $\tikzcmark$                            & $\tikzxmark$               \\
Adapters       & $\tikzcmark$              & $\tikzxmark$                            & $\tikzcmark$               \\
BitFit   & $\tikzcmark$              & $\tikzcmark$                            & $\tikzcmark$              \\
P-Tuning v1/v2 & $\tikzcmark$              & $\tikzxmark$                            & $\tikzcmark$               \\
AoT P-Tuning (ours)   & $\tikzcmark$              & $\tikzcmark$                            & $\tikzcmark$              \\
\toprule
\end{tabular}
\caption{Schematic comparison of recent fine-tuning methods with AoT P-Tuning. Recent fine-tuning approaches either allow inference with no computational overhead or multi-task inference. See Section \ref{section-overhead} for details.}
\label{table-schematic}
\end{table*}

\section{Recent Works}

Currently, a wide range of different methods could be referenced with P-Tuning. \citet{v1} proposed to add soft prompts to the embeddings of GPT-2's input sequence \citep{gpt} to train it on classification tasks. \citet{google-pt} proposed a scheme similar to the one used in \citet{v1}, but trained a T5 model \citep{t5} with P-Tuning to show how the performance of the method changes with the increased scale of the backbone model.

Recently, \citet{mixture, prefix, v2} proposed to add prefixes not only to input embeddings but also at each layer of the Transformer model. In addition, \citet{v2} suggested training a linear classification head on top of the backbone model instead of utilizing a LM head to obtain classification results.

Due to this range of similar methods, we will follow the naming used by \citet{v2} and refer to Prompt-Tuning (adding soft prompts to the input embeddings) as P-Tuning v1 and to Prefix-Tuning (adding soft prefixes at each layer of Transformer backbone) as P-Tuning v2.

\citet{lora} proposed to train low-rank changes of attention weights, while \citet{adapters} fine-tuned additional model layers, which can also be considered parameter-efficient. \citet{bitfit} proposed to fine-tune only bias terms of the model.

\section{Ahead-of-Time P-Tuning}
\label{section-aot}

\subsection{Motivation}
\label{section-overhead}

 Parameter-efficient fine-tuning methods are crucial for reducing computational and memory requirements, enabling deploying of advanced AI models in resource-constrained environments and real-world applications. In discussions of parameter-efficient fine-tuning methods, recent research often refers to the number of parameters utilized for optimization \citep{v1, v2, adapters, lora, bitfit, google-pt}. A method can be considered superior to another if it uses fewer trained parameters and achieves better performance in terms of metrics on downstream tasks.

In this study, we categorize methods based on two additional characteristics: the inference overhead produced by the parameter-efficient fine-tuning method and the capability to perform multi-task inference after fine-tuning multiple models for distinct tasks. While minimizing inference overhead is beneficial for faster model evaluation, the practicality of multi-task inference is debatable. However, multi-task inference is desirable for high-demand systems deployed in real-world scenarios. It can enable more efficient resource allocation, reduce the overall number of models required, and facilitate dynamic workload adjustment based on different tasks as needed. Using the same backbone model for various tasks promotes more streamlined serving, as all workers share the same model in memory. This approach enables dynamic workload adjustment based on different tasks as needed. In contrast, without multi-task inference, some workers might be underutilized depending on external conditions, resulting in suboptimal resource allocation that could have been used for other tasks.

Table \ref{table-schematic} compares recent methods concerning these properties. P-Tuning v1/v2 has inference overhead because the length of sequences passed to the model increases, while adapters introduce overhead by adding additional layers to the model computation. LoRA's analysis is more complex, as it can be evaluated in two modes: one in which we fuse the weights of LoRA \citep{lora}, resulting in no inference overhead since the model is now identical to the original pre-trained model, and another where we do not fuse weights, leading to inference overhead due to the need for additional matrix multiplications during evaluation.

In terms of multi-task inference capability, P-Tuning v1/v2 allows for this, as different prompts can be stacked in a batch and propagated through the backbone model. For Adapters, weights can also be stacked in a batch and evaluated using batched matrix multiplication. However, it is important to note that these layers must have the same shape, which is a tunable hyperparameter. While multi-task inference with a fused LoRA is theoretically possible, similar to Adapters, it demands a substantial amount of GPU memory. For example, for the DeBERTa-XL model with layers of hidden size d = 1024 and l = 48, passing a batch with b sequences necessitates storing 1024 * 1024 * 48 * 4 * b parameters for the model, where 4 represents the number of parameter matrices of the Attention module. Notably, with b = 4, the number of original parameters in the model is already exceeded, which is impractical. To address this issue, one can opt not to fuse LoRA weights (to work with factorized weights with low rank, thus reducing the total number of parameters passed as a batch) for multi-task inference, although this can introduce computational overhead.

In contrast, BitFit \citep{bitfit} is advantageous in terms of both criteria: it only modifies biases of the model, thereby not altering its structure or introducing computational overhead, and can easily be parallelized for multi-task inference by stacking biases for different tasks in a batch. However, based on our experimental results in subsequent sections, its performance is inferior to other methods.

In light of these considerations, this paper aims to answer the following research question: \textit{Can we develop a parameter-efficient fine-tuning method that combines the rapid evaluation and multi-task inference capabilities of BitFit while outperforming it on downstream tasks?}

\subsection{Proposed Mechanism}
 With AoT P-Tuning, we propose to augment each Transformer layer with a simple procedure. We define trainable matrices $\mP^i \in \R^{|\mV| \times d}$ for each layer. Then, before the evaluation of the $i$-th layer, we modify the hidden states as follows:

\begin{equation} 
\label{equation-aot-pt} 
\begin{split} & \mH^{\prime i} = \mH^i + {\mP^i_{x_1}, \dots, \mP^i_{x_n}} \in \R^{n\times d}, 
\end{split} \end{equation}

where $\mP^i_{x_j} \in \R^{d}$ is a lookup of $x_j$-th prompt embedding from $\mP^i$. Such a scheme saves us significant time during evaluation since AoT P-Tuning does not imply an increase in sequence length and involves only adding biases during the evaluation, which introduces negligible computational overhead.

While $\mP^i$ in naive implementation will require much memory to store parameters, we describe reparametrizations that make training more tractable in the following subsections.

Note that AoT P-Tuning could be evaluated in parallel with several tasks in a batch due to the fact that performing look-up from $\mP$ can be easily parallelized.
As for other methods for parameter-efficient fine-tuning, we only optimize parameters of $\mP$ and Classification Head during fine-tuning.

\subsection{On the Parameter Efficiency of AoT P-Tuning}
\label{section-efficiency}
It is notable that, in most cases, it is not feasible to optimize the weight $\mP \in \R^{|\mV| \times d}$ for each layer. If we consider training RoBERTa-Large with such a scheme (which has $|\mV| = 50265$, $d = 1024$ and $l = 24$), then storing all biases $\mP$ will exceed $1.2$B parameters, while the model itself has roughly $350$M parameters.

To overcome this limitation, we propose two reparametrizations of $\mP$ so that it can use fewer parameters during training.

The first is based on the Kronecker product (namely, \textbf{Kronecker AoT P-Tuning}). More specifically, we reparametrize $\mP$ as

\begin{equation}
\label{equation-aot-kron}
    \mP = (\mW_L \otimes \mW_M)\mW_R,
\end{equation}

where $\mW_L \in \R^{a \times r}$, $\mW_M \in \R^{b \times r}$, $\mW_R \in \R^{r^2 \times d}$, $a$ and $b$ are selected in such a way so $a * b = |\mV|$, $r$ is the factorization rank which is a hyperparameter to tune, and $\otimes$ denotes the Kronecker product.

With this reparametrization, training AoT P-Tuning becomes tractable. E.g., for RoBERTa-Large, with $a = 256$, $b = 200$, and $r = 20$, $\mP$ will contain roughly $10$M parameters, which is less than $3\%$ of the total number of parameters in the model\footnote{One may note that $256 * 200 = 51200 \neq 50265$. However, $50265$ is difficult to factorize efficiently since $50265 = 1117 * 3^2 * 5$. Because of this, we chose to mostly factorize $\mP$ in such a way as to make it slightly larger than the original vocabulary size. Doing so allows us to select more appropriate $a$ and $b$ from the perspective of parameter and computational efficiency.}.

The second approach to work with $\mP$, which we used in our experiments, is based on passing the embeddings matrix $\mE$ through a learnable Fully Connected network (namely, \textbf{FC AoT P-Tuning}). Thus, we reparametrize $\mP$ as

\begin{equation}
\label{equation-aot-fc}
    \mP = f(\mE\mW_1 + \vb_1)\mW_2 + \vb_2,
\end{equation}

where $\mW_1 \in \R^{d \times r}$,  $\vb_1 \in \R^{r}$, $\mW_2 \in \R^{r \times d}$, $\vb_2 \in \R^{d}$, $f$ is a non-linearity, and $r$ is the mapping rank, which is also a hyperparameter to tune, same as for Kronecker AoT P-Tuning. 

With FC AoT P-Tuning, we utilize knowledge stored in the pre-trained embeddings matrix $\mE$, which should hypothetically perform better than training $\mP$ from scratch as Kronecker AoT P-Tuning.

Note that for both Kronecker and FC AoT P-Tuning, we can evaluate only specific rows $\{\mP_{x_i}, \dots, \mP_{x_n}\}$ for input sequence $\{x_1, \dots, x_n\}$, making training more efficient.

For both reparametrizations, $\mP$ could be fused once training is complete, and thus the rank of factorization $r$ does not affect inference speed. During the evaluation, there is no need to store the full $\mP$ in GPU memory. Instead, it could be stored in RAM, and only rows of these matrices should be placed in GPU memory to be added to the hidden states before each layer. 

From a certain perspective, choosing between AoT P-Tuning and P-Tuning is a trade-off between evaluation speed and RAM consumption during inference. If RAM is limited, then usual P-Tuning could be used at the cost of slower inference. In other cases, AoT P-Tuning is viable if there is enough RAM and inference speed is crucial. Although, in most cases, $\mP$ matrices for different tasks could be easily stored in the RAM. For RoBERTa-Large, a single task parameter will require roughly $2.4$Gb if stored in half-precision.

\subsection{Intuition Behind AoT P-Tuning and Connection to Other Methods}
\label{section-intuition}

Having $\mH^\prime$, after passing through $\mW_Q$, $\mW_K$, and $\mW_V$ we obtain $\mQ^\prime$, $\mK^\prime$, and $\mV^\prime$. Note that
$\mV^\prime = \mH\mW_V + \{\mP_{x_1}, \dots, \mP_{x_n}\}\mW_V \stackrel{\text{def}}{=} \mV + \mP_x\mW_V$.

The result of evaluating Attention with AoT P-Tuning could be seen as:

\begin{equation}
\label{equation-aot-pt-dis}
    \begin{split}
    \mA_{i}^\prime = &\sum_{j=1}^n \va_j(\mQ_i^\prime, \mK^\prime)\mP_{x_j}\mW_V + \sum_{j=1}^n \va_j(\mQ_i^\prime, \mK^\prime)\mV_j.
    \end{split}
\end{equation}

From such a perspective, there is a clear connection between AoT P-Tuning (Equation \ref{equation-aot-pt-dis}) and P-Tuning v2 (Equation \ref{equation-ptv2-dis}) with the following changes:

\begin{enumerate}
    \item For AoT P-Tuning, attention weights $\va_j$, $j \in \overline{1, l}$ are used for both terms in Equation \ref{equation-aot-pt-dis}.
    \item For AoT P-Tuning, attention is evaluated on modified $\mQ^\prime$. In addition, there is a difference in the form of dependency of $\mK^\prime$ and $\mV^\prime$ on prefix weight. For AoT P-Tuning, we add prefixes to $\mK$ and $\mV$, while for P-Tuning v2, prefixes are concatenated to these matrices.
    \item For AoT P-Tuning, the first term of Equation \ref{equation-aot-pt-dis} implies evaluation of Attention with a prompt which is dependent on the input text, while for P-Tuning v2, the prompt $\mP_{V}$ is constant.
\end{enumerate}

Considering Equation \ref{equation-aot-pt-dis}, AoT can be seen as a form of the P-Tuning method, for which we embed prefixes before evaluating the attention layer\footnote{It is possible to think of AoT P-Tuning as a method which adds bias \textbf{after} the evaluation of the Transformer layer. In this case, it could be seen as a method that directly models the result of the evaluation of P-Tuning v2 with a slightly different computation order. However, we believe that this way is more difficult to consider.}.

Also, one may note that AoT P-Tuning is highly related to BitFit \citep{bitfit}. Both methods are identical in their practical properties as they are zero-cost during the inference and allow multi-task inference (See Table \ref{table-schematic}). Although, there is a clear difference in the form of biases added during model evaluation: BitFit adds constant bias to each element in the sequence as $\mH^{\prime i} = \mH^i + \{\vb, \dots, \vb\}$. Such bias leads to implied Attention results in the form of:

\begin{equation}
\label{equation-aot-bitfit-dis}
    \begin{split}
    \mA_{i_{BitFit}}^\prime = \vb\mW_V + \sum_{j=1}^n \va_j(\mQ_i^\prime, \mK^\prime)\mV_j ,
    \end{split}
\end{equation}

which no longer modifies the attention map with input-dependent bias. In the latter sections, we will show that \textbf{such discrepancy leads to poor performance of BitFit during fine-tuning compared to AoT P-Tuning}.

\section{Experiments}
\subsection{Experimental Details}
\label{section-glue-experiments}

We compared AoT P-Tuning (Kronecker and FC reparametrizations of $\mP$) with other fine-tuning methods capable of performing multi-task inference: P-Tuning v1, P-Tuning v2 on GLUE and SuperGLUE \citep{glue, superglue} Benchmarking Datasets. We also evaluated plain fine-tuning, LoRA, Adapters, and BitFit for reference. For each fine-tuning approach, we experimented with the RoBERTa-Base, RoBERTa-Large, and DeBERTa-XL backbone models.

For each task, we performed a grid hyperparameter search (see Appendix Table 2 for hyperparameter ranges). For RoBERTa models, we evaluated each hyperparameter set with $5$ different seed values and reported median and std score values for each task. For DeBERTa-XL, we used to assess each hyperparameter assignment with a single seed due to longer training time. See Appendix Table 1 for a list of metrics used for each task.

\begin{figure*}[h!]
  \centering
  
    \medskip
        \begin{subfigure}[t]{.4\linewidth}
    \centering\includegraphics[width=\linewidth]{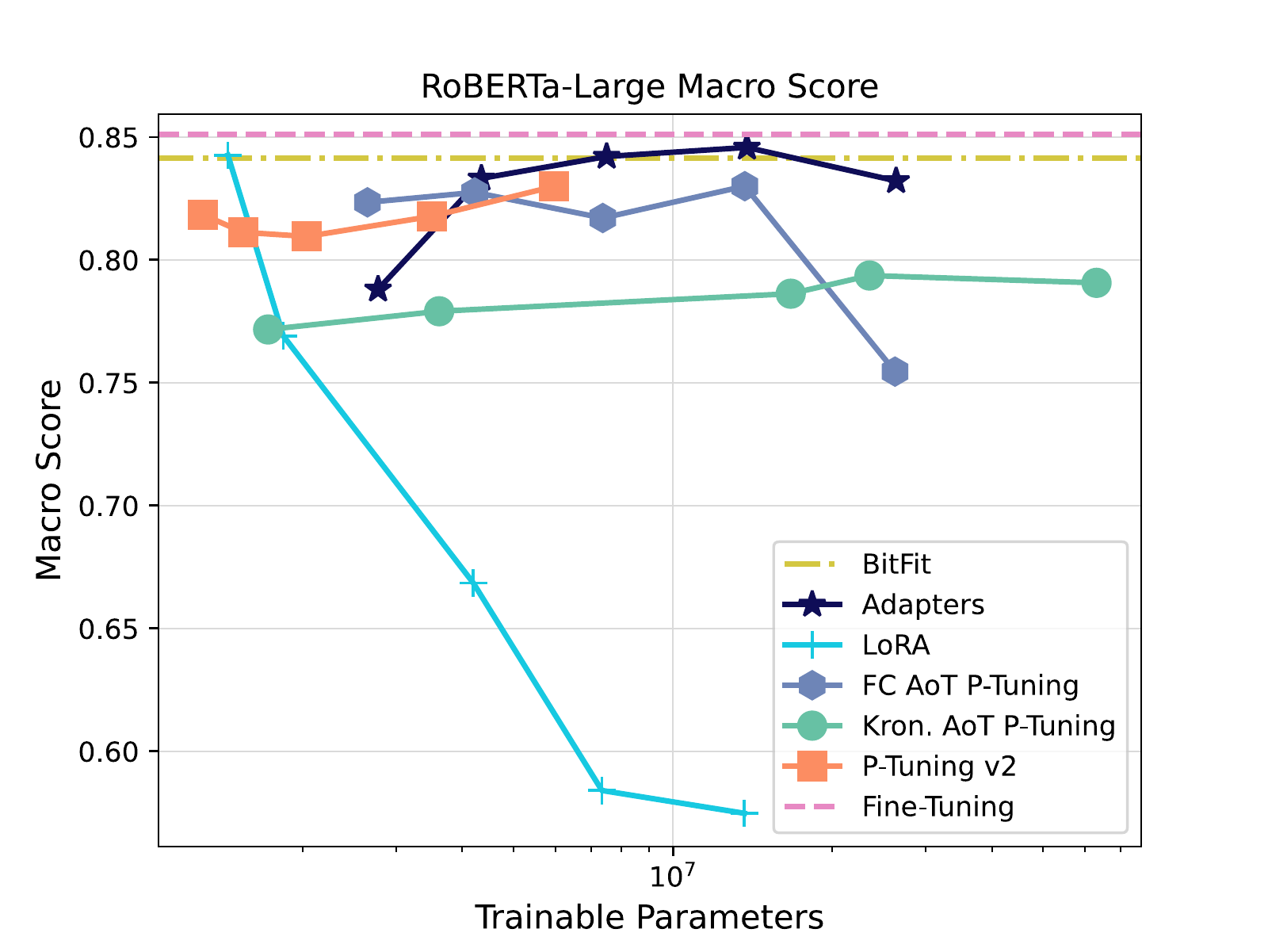}
    \caption{}
  \end{subfigure}
      \begin{subfigure}[t]{.4\linewidth}
    \centering\includegraphics[width=\linewidth]{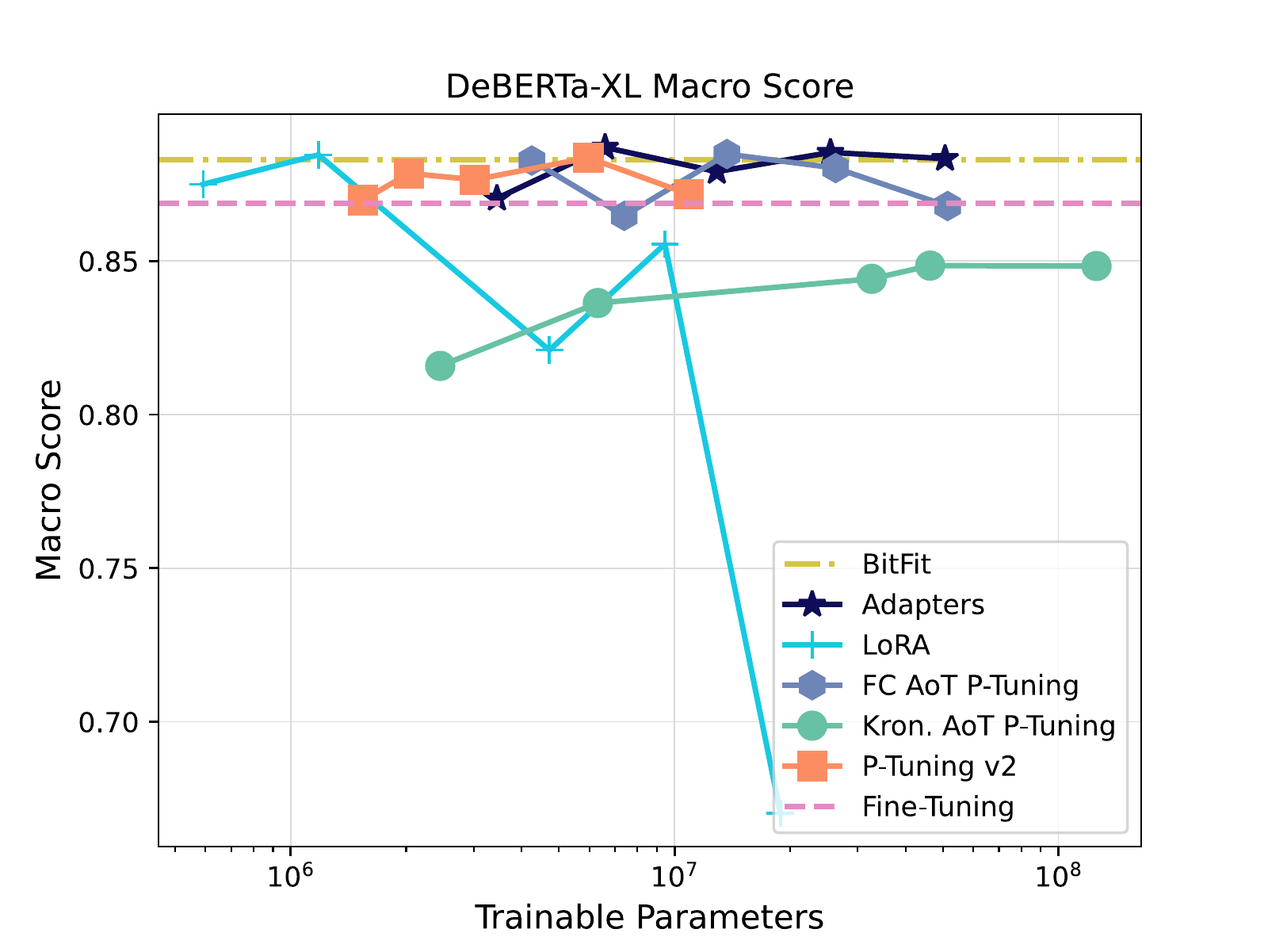}
    \caption{}
  \end{subfigure}

  \caption{SuperGLUE macro score for RoBERTa-Base (a) and DeBERTa-XL (b) models. See Section \ref{section-glue-results} for details.}
  \label{figure-macro}
\end{figure*}

We used the Adam \citep{adam} optimizer with a constant learning rate for each task. We stopped training once the validation metric stopped increasing (see the "patience" parameter in Appendix Table 4). 

For Kronecker AoT P-Tuning with RoBERTa models, we parametrized the matrix $\mP = (\mW_L \otimes \mW_M)\mW_R$ with $a = 256$, and $b = 200$, while for DeBERTa, we used $a = b = 360$. $\mW_L$ and $\mW_M$ were initialized randomly, while $\mW_R$ was initialized as a zero matrix. For FC AoT P-Tuning, we initialized $\mW_1$ randomly, while $\mW_2$, $\vb_1$, and $\vb_2$ were initialized with zeros. For Kronecker AoT P-Tuning, we applied dropout \citep{dropout} to the $\mP_x$ with a fixed probability equal to $0.1$. In contrast, for FC AoT P-Tuning, we applied dropout to $\mE$ before multiplying it with $\mW_1$.

Each experiment was run on a single NVIDIA A100 GPU with a total computation time of roughly $1200$ days.

\begin{table*}[ht!]
\small
\centering
\begin{tabular}{r|cccc}
\toprule
\multicolumn{5}{c}{RoBERTa-Large} \\
\toprule
Model & RTE & COPA & WSC & WiC \\
\toprule
Fine-Tuning & 88.1 ± 1.5 & {87.0 ± 10.2} & {80.8 ± 6.3} & {73.8 ± 1.6} \\
\midrule
Adapters & \underline{87.7 ± 18.5} & \underline{89.0 ± 10.1} & \underline{77.9 ± 9.8} & \textbf{73.5 ± 1.0} \\
LoRA & 87.4 ± 1.6 & \textbf{91.0 ± 8.5} &  \textbf{79.8 ± 10.6} & 71.9 ± 1.2 \\
BitFit & \underline{87.7 ± 0.8} & \textbf{91.0 ± 2.3} & 71.2 ± 6.7 & 71.3 ± 9.5 \\
\midrule

P-Tuning v1 & 62.8 ± 2.3 & 75.0 ± 4.3 & 66.3 ± 1.3 & 64.1 ± 0.9 \\
P-Tuning v2 & 87.4 ± 1.5 & {87.0 ± 6.3} & 75.0 ± 7.7 & 70.8 ± 1.5 \\
\midrule
Kron. AoT P-Tuning (ours) & 84.8 ± 1.3 & 72.0 ± 9.1 & 67.3 ± 3.0 & 71.0 ± 1.0 \\
FC AoT P-Tuning (ours) & \textbf{{88.4 ± 0.9}} & 85.0 ± 10.1 & \textbf{79.8 ± 4.1} & \underline{72.1 ± 1.5} \\
\toprule
 & MultiRC & CB & BoolQ & Macro \\
 \toprule
Fine-Tuning & 83.3 ± 1.1 & 97.3 ± 2.8 & 85.6 ± 0.3 & 85.1 \\
\midrule
Adapters & \textbf{83.7 ± 20.3} & \textbf{100.0 ± 0.0} & \textbf{85.7 ± 10.6} & \textbf{85.4} \\
LoRA & 75.7 ± 17.4 & \textbf{100.0 ± 2.6} & 84.6 ± 0.6 & 84.3 \\
BitFit & 82.5 ± 0.6 &  \textbf{100.0 ± 0.7} & 85.4 ± 1.0 & 84.2 \\
\midrule

P-Tuning v1 & 54.3 ± 2.9 & 81.4 ± 3.0 & 64.3 ± 1.2 & 66.9 \\
P-Tuning v2 & 82.4 ± 0.6 & \textbf{100.0 ± 0.8} & 85.0 ± 0.6 & 83.9 \\
\midrule
Kron. AoT P-Tuning (ours) & \underline{82.8 ± 0.8} & \underline{97.3 ± 2.3} & 84.8 ± 0.5 & 80.0 \\
FC AoT P-Tuning (ours) & 82.7 ± 19.3 & \textbf{100.0 ± 0.0} & \underline{85.5 ± 10.3} & \underline{84.8} \\
\toprule
\multicolumn{5}{c}{DeBERTa-XL} \\
\toprule
Model & RTE & COPA & WSC & WiC \\
\toprule
Fine-Tuning & 89.9 & 96.0 & 76.9 & 75.9 \\
\midrule
Adapters & 90.3 & 96.0 & \underline{89.4} & \textbf{77.3} \\
LoRA & 90.3 & \underline{97.0} & \underline{89.4} & 75.5 \\

BitFit & 89.2 & \underline{97.0} & 86.5 & 73.7 \\
\midrule
P-Tuning v1 & 78.3 & 90.0 & 67.3 & 66.8 \\
P-Tuning v2 & \underline{90.6} & \underline{97.0} & \underline{89.4} & \underline{76.5} \\
\midrule
Kron. AoT P-Tuning (ours) & 88.8 & 96.0 & 87.5 & 71.8 \\
FC AoT P-Tuning (ours) & \textbf{{91.0}} & \textbf{{98.0}} & \textbf{{94.2}} & 74.1 \\
\toprule
 & MultiRC & CB & BoolQ & Macro \\
 \toprule
Fine-Tuning & 84.3 & 98.4 & 86.7 & 86.9 \\
\midrule
Adapters & \underline{86.7} & \underline{97.3} & \textbf{88.9} & \underline{89.4} \\
LoRA & 86.0 & \textbf{100.0} & \underline{88.3} & \textbf{89.5} \\
BitFit & 85.2 & \textbf{100.0} & 86.5 & 88.3 \\
\midrule
P-Tuning v1 & 82.1 & 93.8 & 79.4 & 79.7 \\
P-Tuning v2 & \textbf{{87.1}} & \underline{97.3} & 87.0 & {89.3} \\
\midrule
Kron. AoT P-Tuning (ours) & 86.3 & 83.1 & 87.3 & 85.8 \\
FC AoT P-Tuning (ours) & {86.5} & 92.3 & {88.1} & 89.2 \\
\toprule
\end{tabular}
\caption{Results on the SuperGLUE Dev set. For RoBERTa-Large, each result is median and std across several seeds, and the Macro column is a mean score across all tasks. For DeBERTa-XL, we evaluated each hyperparameter assignment with a single seed and reported its metric score. We bolded the best results and underlined the second best results. Fine-tuning is omitted from comparison with other methods and was not bolded for visibility. See Section \ref{section-glue-results} for details.}
\label{table-superglue}
\end{table*}

\subsection{Results}
\label{section-glue-results}
Refer to Table \ref{table-superglue} and Appendix Table 3 for the trained model results.

Our observations indicate that FC AoT P-Tuning outperforms Kronecker AoT P-Tuning. This outcome mainly results from FC reparametrization using a pre-trained embedding matrix rather than learning biases from scratch.

We also observed that FC AoT P-Tuning performs better or is similar to P-Tuning v1/v2 and LoRA measured by Macro score. Additionally, FC AoT P-Tuning primarily yields comparable results to Adapters. In cases where Adapters outperform AoT P-Tuning, the gains are usually marginal based on our experiments. Furthermore, AoT P-Tuning surpasses BitFit in most instances. Only for the RoBERTa-Base backbone evaluated with GLUE Macro score, BitFit performs on par with AoT P-Tuning. We noticed that both AoT P-Tuning reparametrizations predominantly exhibit a lower variance of metrics across different seeds.

Note that the results presented in Table \ref{table-superglue} and Appendix Table 3 were obtained using different training hyperparameters, including distinct factorization ranks for LoRA and Adapters. This implies that if we consider having the same rank for different tasks for LoRA or Adapters, which is essential for multi-task inference (see Section \ref{section-overhead} for details), these results could represent an upper bound of performance. Refer to Figure \ref{figure-macro} for SuperGLUE macro scores with varying prefix lengths $p$, and prefix ranks $r$ for different methods. Also, remember that while we included AoT P-Tuning in these figures for reference, AoT P-Tuning does not require the same rank for multi-task inference since weights are fused during evaluation, and $r$ no longer affects any shape. When considering the same rank for Adapters and LoRA, these methods only marginally outperform BitFit. We also observed that LoRA struggled with large $r$.

Based on our experimental results, if the ability to perform multi-task inference without computational overhead is essential, then AoT P-Tuning will not significantly diminish performance compared to LoRA and Adapters and could be utilized.

\begin{figure*}[h!]
  \centering
      \medskip

        \begin{subfigure}[t]{.32\linewidth}
    \centering\includegraphics[width=\linewidth]{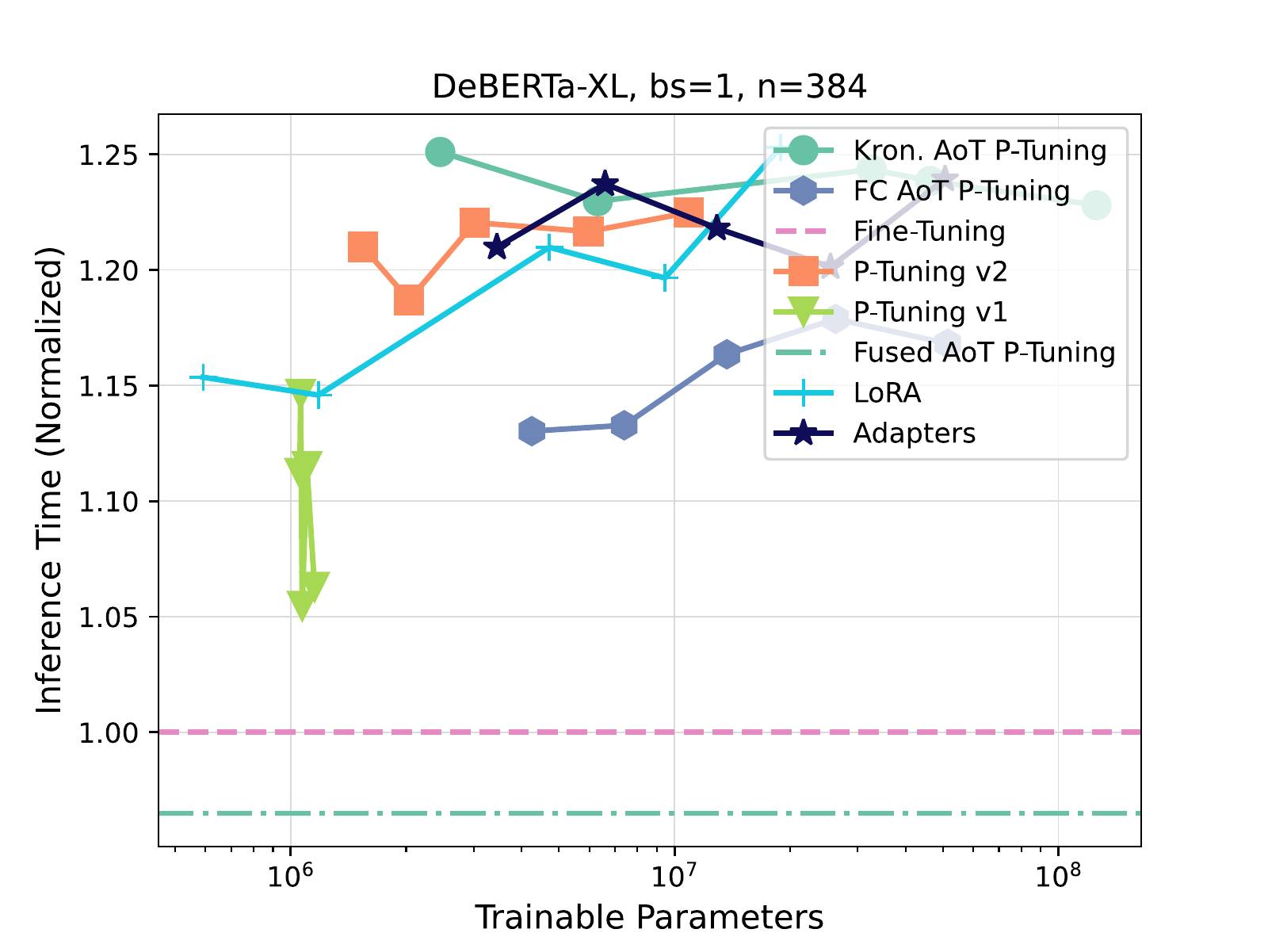}
    \caption{}
  \end{subfigure}
    \begin{subfigure}[t]{.32\linewidth}
    \centering\includegraphics[width=\linewidth]{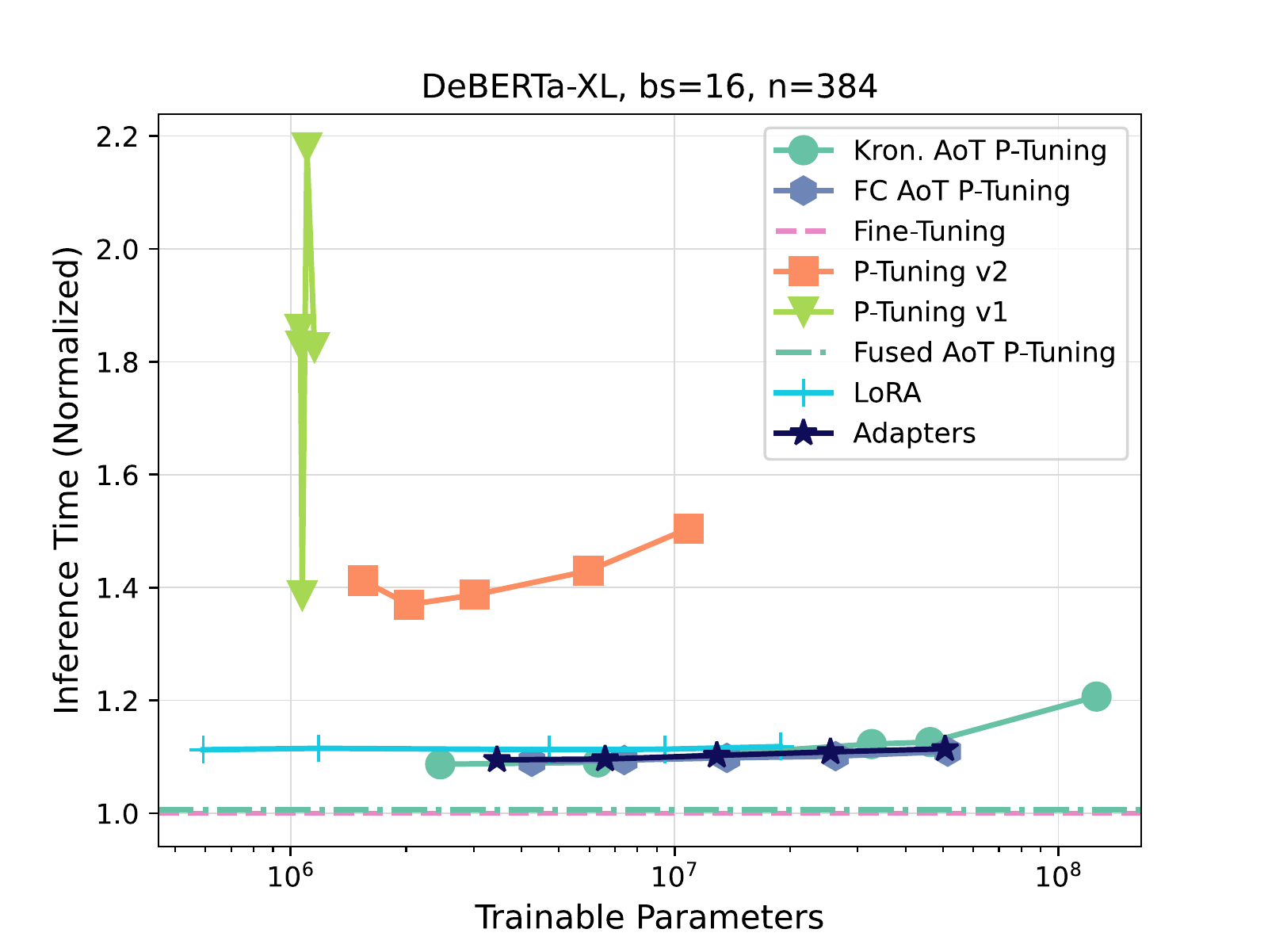}
    \caption{}
  \end{subfigure}
        \begin{subfigure}[t]{.32\linewidth}
    \centering\includegraphics[width=\linewidth]{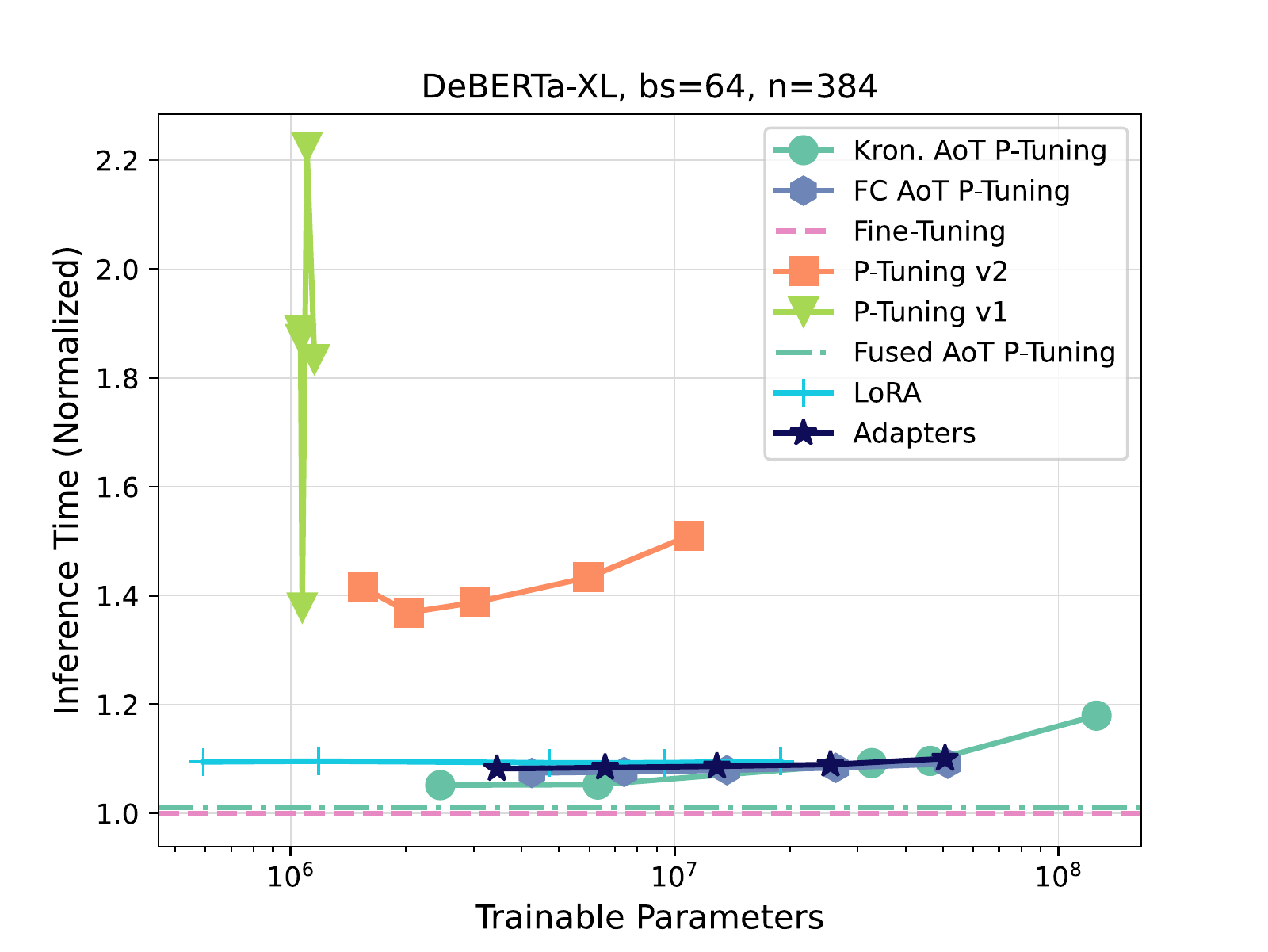}
    \caption{}
  \end{subfigure}
  
  \caption{Speed measurements for baseline methods with sequence length equal to $384$ for DeBERTa-XL. See Appendix Figures 5, 6 for results with other sequence lengths and Section \ref{section-speed-results} for details.}
  \label{figure-speed}
\end{figure*}

With per-task Expected Validation Performance (EVP) \citep{evp}, we observed that AoT P-Tuning highly depends on the number of hyperparameter assignments (see Appendix Figures 2, 4). Although, in most cases, using less than $100$ hyperparameter assignments for AoT P-Tuning is enough for it to outperform P-Tuning v2, which is not crucial in most cases.

\subsection{Analysis of Trained Weights}
\label{section-analysis}
We investigated trained $\mP$ matrices for WSC, COPA, CB, and RTE tasks with the DeBERTa-XL model. Since FC AoT P-Tuning performed better than Kronecker factorization, we selected this reparametrization method to report the results.

More specifically, we sorted rows of $\mP$ matrices for each layer measured by the $L_2$ norm and reported the appropriate tokens for these rows. See Appendix Tables 5, 6, 7, 8 for results.

For the WSC task, there is a clear interpretation of trained rows for $\mP$, since rows with a large $L_2$ norm represent tokens responsible for pronouns and names, which is crucial for solving WSC. For the COPA task, we observed that the model tends to assign large norms for verb tokens. For the RTE and CB tasks, $\mP$ also assigns large norms for name tokens, which often occur in the training data, while CB primarily modifies adverbs for later layers.

\subsection{Inference Speed Overhead}
\label{section-speed-results}
In Figure \ref{figure-speed} and Appendix Figures 5, 6, we investigated the computational overhead of AoT P-Tuning compared to other baselines.

We estimated inference time for RoBERTa-Base, RoBERTa-Large, and DeBERTa-XL models with batch sizes $\in [1, 16, 64]$ and sequence lengths $\in [64, 128, 384]$. For batch size equal to $1$, we evaluated the model $300$ times, and $100$ times for other values. For evaluation, we used A100 80Gb GPU. We report mean values of inference time normalized by the inference time of the vanilla model (i.e., plain fine-tuning).

We evaluated AoT P-Tuning on two setups. For the first setup, we fused $\mP$ so that the model can perform on its top speed. We did not perform fusing for the second setup. While a lack of fusing is not required to use AoT P-Tuning, we report these results for reference. We also report LoRA results for the experiments in which we did not fuse weights. This setup makes it possible to use LoRA in a multi-task setup (see Section \ref{section-overhead} for details). 

We observed that the proposed method performed differently depending on the experiment's parameters. AoT performed with computational overhead for a small model (i.e., RoBERTa-Base), small sequence length, and small batch size. We observed $12\%$ overhead on inference speed compared to plain fine-tuning for RoBERTa-Base with batch size $1$ and sequence length $64$. However, AoT P-Tuning still performed faster than other baselines by a large margin (i.e., LoRA for RoBERTa-Base with batch size $1$ and sequence length $384$ added $50-70\%$ computational overhead compared to fine-tuning).

Once model size or input size is increased, we observed that the overhead of adding biases for AoT P-Tuning becomes negligible. The proposed method performed the same as plain fine-tuning (for some experiments, we observed $1-2\%$ overhead, while for others, AoT P-Tuning performed even faster than fine-tuning, which is a variation in inference time measurements). For the most practical setup with large models (DeBERTa-XL), small batch size ($=1$) and long sequence length ($=384$), AoT P-Tuning performed slightly faster than fine-tuning, \textbf{while other methods performed} $12-25\%$ \textbf{slower}. LoRA and Adapters showed $10\%$ overhead for larger batch sizes, while AoT P-Tuning still performed with negligible overhead.

\section{Conclusion and Future Work}
This paper focused on parameter-efficient fine-tuning methods emphasizing inference speed and multi-task capabilities. We observed that widely used methods typically involve a trade-off between fast and multi-task inference, while those that can perform both tend to struggle in downstream task evaluations. To address this challenge, we introduced a novel method called AoT P-Tuning, which has negligible inference overhead while maintaining multi-task performance.

Our experiments demonstrated that AoT P-Tuning outperforms BitFit, which also does not introduce any overhead and could be evaluated in a multi-task manner. Furthermore, AoT P-Tuning is competitive with other methods across various downstream tasks and pre-trained models, offering faster inference times.

Although we investigated two reparametrizations based on the Kronecker product and fully connected (FC) network, further exploration of alternative reparametrizations for weight $\mP$ may enhance the performance of our proposed method. Moreover, while our approach is straightforward, implementing various architectural modifications could improve AoT P-Tuning's performance and decrease the need for hyperparameter tuning.




\bibliography{custom}

\begin{thebibliography}{19}
\providecommand{\natexlab}[1]{#1}
\providecommand{\url}[1]{\texttt{#1}}
\expandafter\ifx\csname urlstyle\endcsname\relax
  \providecommand{\doi}[1]{doi: #1}\else
  \providecommand{\doi}{doi: \begingroup \urlstyle{rm}\Url}\fi

\bibitem[Ben~Zaken et~al.(2022)Ben~Zaken, Goldberg, and Ravfogel]{bitfit}
Elad Ben~Zaken, Yoav Goldberg, and Shauli Ravfogel.
\newblock {B}it{F}it: Simple parameter-efficient fine-tuning for
  transformer-based masked language-models.
\newblock In \emph{Proceedings of the 60th Annual Meeting of the Association
  for Computational Linguistics (Volume 2: Short Papers)}, pp.\  1--9, Dublin,
  Ireland, May 2022. Association for Computational Linguistics.
\newblock \doi{10.18653/v1/2022.acl-short.1}.
\newblock URL \url{https://aclanthology.org/2022.acl-short.1}.

\bibitem[Devlin et~al.(2019)Devlin, Chang, Lee, and Toutanova]{bert}
Jacob Devlin, Ming-Wei Chang, Kenton Lee, and Kristina Toutanova.
\newblock {BERT}: Pre-training of deep bidirectional transformers for language
  understanding.
\newblock In \emph{Proceedings of the 2019 Conference of the North {A}merican
  Chapter of the Association for Computational Linguistics: Human Language
  Technologies, Volume 1 (Long and Short Papers)}, pp.\  4171--4186,
  Minneapolis, Minnesota, June 2019. Association for Computational Linguistics.
\newblock \doi{10.18653/v1/N19-1423}.
\newblock URL \url{https://aclanthology.org/N19-1423}.

\bibitem[Dodge et~al.(2019)Dodge, Gururangan, Card, Schwartz, and Smith]{evp}
Jesse Dodge, Suchin Gururangan, Dallas Card, Roy Schwartz, and Noah~A. Smith.
\newblock Show your work: Improved reporting of experimental results.
\newblock In \emph{Proceedings of the 2019 Conference on Empirical Methods in
  Natural Language Processing and the 9th International Joint Conference on
  Natural Language Processing (EMNLP-IJCNLP)}, pp.\  2185--2194, Hong Kong,
  China, November 2019. Association for Computational Linguistics.
\newblock \doi{10.18653/v1/D19-1224}.
\newblock URL \url{https://aclanthology.org/D19-1224}.

\bibitem[He et~al.(2020)He, Liu, Gao, and Chen]{deberta}
Pengcheng He, Xiaodong Liu, Jianfeng Gao, and Weizhu Chen.
\newblock Deberta: Decoding-enhanced bert with disentangled attention.
\newblock \emph{CoRR}, abs/2006.03654, 2020.
\newblock URL
  \url{http://dblp.uni-trier.de/db/journals/corr/corr2006.html#abs-2006-03654}.

\bibitem[Houlsby et~al.(2019)Houlsby, Giurgiu, Jastrzebski, Morrone,
  De~Laroussilhe, Gesmundo, Attariyan, and Gelly]{adapters}
Neil Houlsby, Andrei Giurgiu, Stanislaw Jastrzebski, Bruna Morrone, Quentin
  De~Laroussilhe, Andrea Gesmundo, Mona Attariyan, and Sylvain Gelly.
\newblock Parameter-efficient transfer learning for {NLP}.
\newblock In Kamalika Chaudhuri and Ruslan Salakhutdinov (eds.),
  \emph{Proceedings of the 36th International Conference on Machine Learning},
  volume~97 of \emph{Proceedings of Machine Learning Research}, pp.\
  2790--2799. PMLR, 09--15 Jun 2019.
\newblock URL \url{https://proceedings.mlr.press/v97/houlsby19a.html}.

\bibitem[Hu et~al.(2022)Hu, Shen, Wallis, Allen-Zhu, Li, Wang, Wang, and
  Chen]{lora}
Edward~J. Hu, Yelong Shen, Phillip Wallis, Zeyuan Allen-Zhu, Yuanzhi Li, Shean
  Wang, Lu~Wang, and Weizhu Chen.
\newblock Lora: Low-rank adaptation of large language models.
\newblock In \emph{ICLR 2022}, April 2022.
\newblock URL
  \url{https://www.microsoft.com/en-us/research/publication/lora-low-rank-adaptation-of-large-language-models/}.

\bibitem[Kingma \& Ba(2015)Kingma and Ba]{adam}
Diederik~P. Kingma and Jimmy Ba.
\newblock Adam: {A} method for stochastic optimization.
\newblock In Yoshua Bengio and Yann LeCun (eds.), \emph{3rd International
  Conference on Learning Representations, {ICLR} 2015, San Diego, CA, USA, May
  7-9, 2015, Conference Track Proceedings}, 2015.
\newblock URL \url{http://arxiv.org/abs/1412.6980}.

\bibitem[Lester et~al.(2021)Lester, Al-Rfou, and Constant]{google-pt}
Brian Lester, Rami Al-Rfou, and Noah Constant.
\newblock The power of scale for parameter-efficient prompt tuning.
\newblock In \emph{Proceedings of the 2021 Conference on Empirical Methods in
  Natural Language Processing}, pp.\  3045--3059, Online and Punta Cana,
  Dominican Republic, November 2021. Association for Computational Linguistics.
\newblock \doi{10.18653/v1/2021.emnlp-main.243}.
\newblock URL \url{https://aclanthology.org/2021.emnlp-main.243}.

\bibitem[Li \& Liang(2021)Li and Liang]{prefix}
Xiang~Lisa Li and Percy Liang.
\newblock Prefix-tuning: Optimizing continuous prompts for generation.
\newblock In \emph{Proceedings of the 59th Annual Meeting of the Association
  for Computational Linguistics and the 11th International Joint Conference on
  Natural Language Processing (Volume 1: Long Papers)}, pp.\  4582--4597,
  Online, August 2021. Association for Computational Linguistics.
\newblock \doi{10.18653/v1/2021.acl-long.353}.
\newblock URL \url{https://aclanthology.org/2021.acl-long.353}.

\bibitem[Liu et~al.(2021{\natexlab{a}})Liu, Ji, Fu, Du, Yang, and Tang]{v2}
Xiao Liu, Kaixuan Ji, Yicheng Fu, Zhengxiao Du, Zhilin Yang, and Jie Tang.
\newblock P-tuning v2: Prompt tuning can be comparable to fine-tuning
  universally across scales and tasks.
\newblock \emph{CoRR}, abs/2110.07602, 2021{\natexlab{a}}.
\newblock URL \url{https://arxiv.org/abs/2110.07602}.

\bibitem[Liu et~al.(2021{\natexlab{b}})Liu, Zheng, Du, Ding, Qian, Yang, and
  Tang]{v1}
Xiao Liu, Yanan Zheng, Zhengxiao Du, Ming Ding, Yujie Qian, Zhilin Yang, and
  Jie Tang.
\newblock Gpt understands, too.
\newblock \emph{arXiv:2103.10385}, 2021{\natexlab{b}}.

\bibitem[Liu et~al.(2019)Liu, Ott, Goyal, Du, Joshi, Chen, Levy, Lewis,
  Zettlemoyer, and Stoyanov]{roberta}
Yinhan Liu, Myle Ott, Naman Goyal, Jingfei Du, Mandar Joshi, Danqi Chen, Omer
  Levy, Mike Lewis, Luke Zettlemoyer, and Veselin Stoyanov.
\newblock Roberta: A robustly optimized bert pretraining approach, 2019.
\newblock URL \url{http://arxiv.org/abs/1907.11692}.
\newblock cite arxiv:1907.11692.

\bibitem[Qin \& Eisner(2021)Qin and Eisner]{mixture}
Guanghui Qin and Jason Eisner.
\newblock Learning how to ask: Querying {LM}s with mixtures of soft prompts.
\newblock In \emph{Proceedings of the 2021 Conference of the North American
  Chapter of the Association for Computational Linguistics: Human Language
  Technologies}, pp.\  5203--5212, Online, June 2021. Association for
  Computational Linguistics.
\newblock \doi{10.18653/v1/2021.naacl-main.410}.
\newblock URL \url{https://aclanthology.org/2021.naacl-main.410}.

\bibitem[Radford et~al.(2019)Radford, Wu, Child, Luan, Amodei, and
  Sutskever]{gpt}
Alec Radford, Jeff Wu, Rewon Child, David Luan, Dario Amodei, and Ilya
  Sutskever.
\newblock Language models are unsupervised multitask learners.
\newblock 2019.

\bibitem[Raffel et~al.(2020)Raffel, Shazeer, Roberts, Lee, Narang, Matena,
  Zhou, Li, and Liu]{t5}
Colin Raffel, Noam Shazeer, Adam Roberts, Katherine Lee, Sharan Narang, Michael
  Matena, Yanqi Zhou, Wei Li, and Peter~J. Liu.
\newblock Exploring the limits of transfer learning with a unified text-to-text
  transformer.
\newblock \emph{Journal of Machine Learning Research}, 21\penalty0
  (140):\penalty0 1--67, 2020.
\newblock URL \url{http://jmlr.org/papers/v21/20-074.html}.

\bibitem[Srivastava et~al.(2014)Srivastava, Hinton, Krizhevsky, Sutskever, and
  Salakhutdinov]{dropout}
Nitish Srivastava, Geoffrey Hinton, Alex Krizhevsky, Ilya Sutskever, and Ruslan
  Salakhutdinov.
\newblock Dropout: A simple way to prevent neural networks from overfitting.
\newblock \emph{Journal of Machine Learning Research}, 15\penalty0
  (56):\penalty0 1929--1958, 2014.
\newblock URL \url{http://jmlr.org/papers/v15/srivastava14a.html}.

\bibitem[Vaswani et~al.(2017)Vaswani, Shazeer, Parmar, Uszkoreit, Jones, Gomez,
  Kaiser, and Polosukhin]{transformer}
Ashish Vaswani, Noam Shazeer, Niki Parmar, Jakob Uszkoreit, Llion Jones,
  Aidan~N Gomez, {\L}ukasz Kaiser, and Illia Polosukhin.
\newblock Attention is all you need.
\newblock In \emph{Advances in Neural Information Processing Systems}, pp.\
  5998--6008, 2017.

\bibitem[Wang et~al.(2018)Wang, Singh, Michael, Hill, Levy, and Bowman]{glue}
Alex Wang, Amapreet Singh, Julian Michael, Felix Hill, Omer Levy, and Samuel~R.
  Bowman.
\newblock Glue: A multi-task benchmark and analysis platform for natural
  language understanding, 2018.
\newblock URL \url{http://arxiv.org/abs/1804.07461}.
\newblock cite arxiv:1804.07461Comment: https://gluebenchmark.com/.

\bibitem[Wang et~al.(2019)Wang, Pruksachatkun, Nangia, Singh, Michael, Hill,
  Levy, and Bowman]{superglue}
Alex Wang, Yada Pruksachatkun, Nikita Nangia, Amanpreet Singh, Julian Michael,
  Felix Hill, Omer Levy, and Samuel Bowman.
\newblock Superglue: A stickier benchmark for general-purpose language
  understanding systems.
\newblock In H.~Wallach, H.~Larochelle, A.~Beygelzimer, F.~d\textquotesingle
  Alch\'{e}-Buc, E.~Fox, and R.~Garnett (eds.), \emph{Advances in Neural
  Information Processing Systems}, volume~32. Curran Associates, Inc., 2019.
\newblock URL
  \url{https://proceedings.neurips.cc/paper/2019/file/4496bf24afe7fab6f046bf4923da8de6-Paper.pdf}.

\end{thebibliography}
\bibliographystyle{iclr_bib}

\newpage
\appendix

\section{Background}
\label{section-bacground}
\subsection{Evaluation of Transformer}

Having an input sequence $\vx = \{x_1, \dots, x_n\}$, where $x_i$ is token index, the embeddings of input texts are evaluated as $\mH^0 = \{\mE_{x_1}, \dots, \mE_{x_n}\}$, where $\mE \in \R^{|\mV| \times d}$ is the embeddings matrix, $|\mV|$ is the vocabulary size, $d$ is the size of the hidden state of the model, and $\mE_{x_i}$ is an embedding of the token $x_i$. Hidden states $\mH^i$ are then passed to the $(i+1)$-th layer of the Transformer to evaluate $\mH^{i+1}$ with a total $l$ number of layers. To do so, $\mH^i$ are first mapped through three matrices $\mW_Q$, $\mW_K$, $\mW_V \in \R^{d \times d}$ to get $\mQ$, $\mK$ and $\mV$, which are then used to evaluate the  attention layer's results as:

\begin{equation}
    \begin{split}
    \mA &= \text{attention}(\mQ, \mK, \mV) = \text{softmax}(\frac{\mQ \mK^T}{\sqrt{d}})\mV \in \R^{n\times d}.
    \end{split}
\end{equation}

After $\mA$ is evaluated, it is passed through the remaining layers\footnote{In fact, Transformer architecture implies evaluation of multi-head Attention. We omit this in this paper for simplicity since all derivations could be easily extended on the multi-head case.}, including residual connections and FC layers to get $\mH^{i + 1}$. Here and later, we omit the layer index $i$ for attention result $\mA$ for visibility.

\subsection{P-Tuning v1}
Having a pre-trained Transformer LM with parameters $\Theta$, instead of fine-tuning all parameters of this model on a downstream task, it is possible to define soft prompts $\mP \in \R^{p \times d}$, where $p$ is the length of prompt. $\mP$ is then concatenated to input sequence embeddings as:

\begin{equation}
\mH^{\prime 0} = \text{concat}(\mP, \mH^0) \in \R^{(p + n) \times d}.
\end{equation}

Then, only $\mP$ and Classification Head are fine-tuned on a downstream task, while $\Theta$ remains frozen\footnote{Original implementation of P-Tuning v1 implied utilizing the LM Head of a pre-trained model instead of training a Classification Head.}. Such  parametrization of fine-tuning makes it possible to perform multi-task inference. 

\subsection{P-Tuning v2}
\label{section-ptv2}
Instead of concatenation of a single prompt $\mP$ to the $\mH^0$, P-Tuning v2 proposed to concatenate soft prefixes at each layer of the Transformer model. To apply P-Tuning v2, soft prefixes $\mP_K, \mP_V \in \R^{p \times d}$ are defined for each layer and concatenated to the $\mK$ and $\mV$ matrices before evaluating the attention $\mK^\prime = \text{concat}(\mP_K, \mK)$, $\mV^\prime = \text{concat}(\mP_V, \mV)$. Then, Attention is evaluated as follows:

\begin{equation}
\label{equation-ptv2}
\begin{split}
    & \mA^\prime = \text{attention}(\mQ, \mK^\prime, \mV^\prime),
\end{split}
\end{equation}

where $i$-th component of $\mA^\prime$ could be then written as:

\begin{equation}
\label{equation-ptv2-dis}
\begin{split}
    \mA_{i}^\prime = &\sum_{j=1}^p \va_j(\mQ_{i}, \mK^\prime)\mP_{\mV_j} + \sum_{k=1}^{n} \va_{k + p}(\mQ_{i}, \mK^\prime) \mV_{k}.
\end{split}
\end{equation}

Note that $\va \in \R^{p + n}$ are attention weights for the $i$-th token (we omit the $i$-th index for simplicity) and thus $\sum_{j=1}^{p + n} \va_j = 1$.

As for P-Tuning v1, only parameters of soft prefixes $\mP_K, \mP_V$ and Classification Head are optimized on a downstream task while freezing the parameters of a backbone model.

\begin{table*}[!htb]
\centering
\begin{tabular}{rc|rc}
\toprule
Task  & Metric                                       & Task    & Metric                                  \\ \midrule
CoLA  & Mattews Correlation                          & BoolQ   & Accuracy                                \\ \midrule
MRPC & $\frac{\text{Accuracy} + \text{F}1}{2}$ & CB & $\frac{\text{Accuracy} + \text{F}1}{2}$ \\ \midrule
RTE   & Accuracy                                     & RTE     & Accuracy                                \\ \midrule
SST-2 & Accuracy                                     & COPA    & Accuracy                                \\ \midrule
MNLI  & Accuracy                                     & MultiRC & $\frac{\text{Accuracy} + \text{F}1}{2}$ \\ \midrule
QNLI  & Accuracy                                     & WSC     & Accuracy \\ \midrule
QQP   & $\frac{\text{Accuracy} + \text{F}1}{2}$      & WiC     & Accuracy                                \\ \midrule
STSB  & $\frac{\text{Pearson} + \text{Spearman}}{2}$ &     \\  \toprule                        
\end{tabular}
\caption{Metrics used in our experiments for each task. See Section 4 for more details.}
\label{table-metrics}
\end{table*}

\begin{table*}[!htb]
    
    \begin{minipage}{.5\linewidth}
      \centering
\begin{tabular}{r|l}
\toprule
{Parameter}     & {Range}                                                                                               \\ \toprule
\multicolumn{2}{c}{All Tasks, except RTE}                                                                                               \\ \midrule
\multicolumn{2}{c}{P-Tuning v1/v2/AoT}                                                                                                  \\ \midrule
{batch size}    & $16$, $64$                                                                                    \\ 
{learning rate} & \makecell[l]{$1\mathrm{e}{-4}$, $5\mathrm{e}{-4}$, $5\mathrm{e}{-3}$, $1\mathrm{e}{-3}$}      \\ 
{$p$}    & \makecell[l]{$5$, $10$, $20$, $50$, $100$}                                                                                    \\ 
{LoRA $r$}  & \makecell[l]{$2$, $4$, $16$, $32$, $64$}     \\
{Adapters $r$}  & \makecell[l]{$16$, $32$, $64$, $128$, $256$}     \\

{Kron. $r$}  & \makecell[l]{$5$, $10$, $25$, $30$, $50$}     \\
{FC $r$}     & \makecell[l]{$32$, $64$, $128$, $256$, $512$} \\                                                                              \\ \midrule
\multicolumn{2}{c}{Fine-Tuning}                                                                                                         \\ \midrule
{learning rate} & \makecell[l]{$1\mathrm{e}{-5}$, $5\mathrm{e}{-5}$,  $1\mathrm{e}{-4}$,\\ $5\mathrm{e}{-4}$, $5\mathrm{e}{-3}$} \\ \toprule
\multicolumn{2}{c}{RTE}                                                                                                                 \\ \toprule
{batch size}    & $16$, $32$, $64$, $128$                                                                       \\ 
{learning rate} &
  \makecell[l]{$1\mathrm{e}{-5}$, $5\mathrm{e}{-5}$,  $1\mathrm{e}{-4}$, $5\mathrm{e}{-4}$, \\ $5\mathrm{e}{-3}$, $1\mathrm{e}{-3}$, $2\mathrm{e}{-3}$, $1\mathrm{e}{-2}$} \\ \toprule
\end{tabular}
    \end{minipage}%
    \begin{minipage}{.5\linewidth}
      \centering
        
\begin{tabular}{r|l}
\toprule
{Parameter}  & {Range}                                       \\
\toprule
\multicolumn{2}{c}{P-Tuning v1/v2/AoT}                       \\
\midrule
{batch size} & $16$, $32$, $64$                              \\
{learning rate} & \makecell[l]{$5\mathrm{e}{-5}$, $1\mathrm{e}{-4}$, $3\mathrm{e}{-4}$, $5\mathrm{e}{-4}$, \\ $1\mathrm{e}{-3}$, $2\mathrm{e}{-3}$, $5\mathrm{e}{-3}$} \\
{$p$}        & \makecell[l]{$5$, $10$, $20$, $50$, $100$}    \\
{LoRA $r$}  & \makecell[l]{$2$, $4$, $16$, $32$, $64$}     \\
{Adapters $r$}  & \makecell[l]{$16$, $32$, $64$, $128$, $256$}     \\
{Kron. $r$}  & \makecell[l]{$5$, $10$, $25$, $30$, $50$}     \\
{FC $r$}     & \makecell[l]{$32$, $64$, $128$, $256$, $512$} \\
\midrule
\multicolumn{2}{c}{Fine-Tuning}                              \\
\midrule
{learning rate} & \makecell[l]{$1\mathrm{e}{-5}$, $5\mathrm{e}{-5}$,  $1\mathrm{e}{-4}$,\\ $5\mathrm{e}{-4}$, $5\mathrm{e}{-3}$} \\ \toprule                                     
\end{tabular}
    \end{minipage} 
    \caption{Hyperparameter ranges used in experiments with GLUE and SuperGLUE benchmarking datasets for RoBERTa (left) and DeBERTa (right) models. $p$ is the prompt length used for P-Tuning v1/v2, and $r$ is the rank of weight factorization used for AoT P-Tuning (See Section 3.3). For GLUE experiments, each hyperparameter set was evaluated with different seed values. See Section 4 for more details.}
    \label{table-ranges}
\end{table*}

\begin{table*}[ht!]
\small
\centering
\begin{tabular}{r|cccc|c}
\toprule
\multicolumn{6}{c}{RoBERTa-Base}                                                                                                  \\
\toprule
Model                     & STS-B               & SST-2               & RTE                 & QQP                 &               \\
\toprule
Fine-Tuning               & 90.6 ± 0.3          & 95.0 ± 0.2          & 81.2 ± 0.7          & 89.6 ± 0.2          &               \\
\midrule
Adapters & \textbf{90.7 ± 0.2} & {94.4 ± 0.3} & \underline{80.5 ± 2.0}  & \textbf{89.2 ± 0.1} \\
LoRA & {90.1 ± 0.3} & 94.3 ± 0.5 & \underline{80.5 ± 1.8} & 86.3 ± 0.3 \\
BitFit & \underline{90.3 ± 0.1} & \underline{94.5 ± 0.5} & \textbf{80.9 ± 1.4} & 85.5 ± 0.6 \\
\midrule
P-Tuning v1               & 86.9 ± 0.9          & 94.0 ± 0.3          & 60.3 ± 2.4          & 82.2 ± 1.5                        \\
P-Tuning v2               & 89.2 ± 0.3          & \textbf{94.6 ± 0.2} & \underline{80.5 ± 3.4} & 86.4 ± 3.3          &               \\
\midrule
Kron. AoT P-Tuning (ours) & 89.7 ± 0.2          & 94.0 ± 0.2          & 77.6 ± 1.4          & \underline{88.2 ± 0.1}          &               \\
FC AoT P-Tuning (ours)    & {90.0 ± 0.2} & {94.4 ± 0.3}          & {78.0 ± 1.3}          & {87.9 ± 0.2} &               \\
\toprule
                          & QNLI                & MRPC                & MNLI                & CoLA                & Macro         \\
\toprule
Fine-Tuning               & 92.4 ± 0.1          & 90.8 ± 0.5          & 87.0 ± 0.3          & 63.8 ± 1.4          & 86.3          \\
\midrule
Adapters & \textbf{92.4 ± 0.2} & \textbf{91.1 ± 1.1} & \textbf{86.8 ± 0.1} & \textbf{63.0 ± 1.3} & \textbf{86.0} \\
LoRA & 91.6 ± 0.3 & \underline{90.6 ± 0.7} & 84.7 ± 0.3 &  60.3 ± 1.0 & \underline{84.8} \\
BitFit & 90.9 ± 0.5 & 90.5 ± 1.7 & 85.0 ± 0.1 &  60.4 ± 1.2 & 84.7 \\
\midrule
P-Tuning v1               & 88.3 ± 0.5          & 82.0 ± 1.7          & 80.8 ± 0.6          & 45.8 ± 27.1         & 77.5          \\
P-Tuning v2               & \underline{91.9 ± 1.6} & 89.1 ± 1.1          & 85.3 ± 0.2          & \underline{60.7 ± 2.6} & {84.7} \\
\midrule
Kron. AoT P-Tuning (ours) & 90.7 ± 0.4          & 89.5 ± 1.1          & 84.6 ± 0.1          & 59.3 ± 1.2          & 84.2          \\
FC AoT P-Tuning (ours)            & 91.3 ± 0.4           & {90.3 ± 0.3}  & \underline{85.4 ± 0.1}  & 60.3 ± 2.2           & {84.7}        \\
\toprule
\multicolumn{6}{c}{RoBERTa-Large}  \\
\toprule
Model                     & STS-B               & SST-2               & RTE                 & QQP                 &               \\
\toprule
Fine-Tuning               & 91.9 ± 0.2          & 96.1 ± 0.4          & 88.1 ± 1.5          & 90.3 ± 0.2          &               \\
\midrule
Adapters & \textbf{92.1 ± 0.2} & \underline{96.3 ± 0.4} &  \textbf{90.0 ± 0.1} &  \textbf{94.2 ± 0.1} \\
LoRA & 91.4 ± 0.3 & 95.9 ± 0.2 & 87.2 ± 18.7 &  \underline{93.7 ± 23.7} \\
BitFit & \underline{91.8 ± 0.2} & 96.2 ± 0.4 & 87.7 ± 0.8 & 87.2 ± 0.6 \\
\midrule
P-Tuning v1               & 75.5 ± 6.3          & 94.4 ± 0.4          & 62.8 ± 2.3          & 76.9 ± 2.5          &               \\
P-Tuning v2               & 91.0 ± 0.4          & 96.1 ± 0.3          & 87.4 ± 1.5          & 86.6 ± 0.6          &               \\
\midrule
Kron. AoT P-Tuning (ours) & 91.1 ± 0.8          & 96.2 ± 0.2          & 84.8 ± 1.3          & {89.4 ± 0.1}          &               \\
FC AoT P-Tuning (ours)            & {91.7 ± 0.4}  & \textbf{96.7 ± 0.1}  & \underline{88.4 ± 0.9}  & 88.7 ± 0.2  &                      \\
\toprule
                          & QNLI                & MRPC                & MNLI                & CoLA                & Macro         \\
                          \toprule
Fine-Tuning               & 94.3 ± 0.2          & 91.6 ± 0.6          & 89.9 ± 0.2          & 68.1 ± 1.9          & 88.8          \\
\midrule
Adapters &  91.3 ± 0.4 &  90.1 ± 0.2 &  67.2 ± 1.3 & \textbf{87.7 ± 18.5} & \underline{88.6} \\
LoRA & 91.0 ± 7.2 & 88.9 ± 24.0  & 66.3 ± 1.9 &  \underline{87.4 ± 1.6} & 87.7 \\
BitFit & \underline{94.1 ± 0.4} & 91.0 ± 1.0 &  \underline{89.4 ± 0.1} &  69.8 ± 3.1 & 88.4 \\
\midrule
P-Tuning v1               & 79.1 ± 2.4          & 79.0 ± 1.1          & 75.9 ± 18.3         & 24.7 ± 17.6         & 71.0          \\
P-Tuning v2               & 94.0 ± 1.1          & \underline{91.2 ± 0.9}          & \underline{89.4 ± 0.7}          & 66.9 ± 1.5          & 87.8          \\
\midrule
Kron. AoT P-Tuning (ours) & \textbf{94.2 ± 0.1} & 89.7 ± 0.9          & 89.3 ± 0.1          & 65.5 ± 1.9          & 87.5          \\
FC AoT P-Tuning (ours)            & \underline{94.1 ± 0.2}           & \textbf{91.6 ± 0.8}  & \textbf{89.6 ± 0.1}  & {69.2 ± 0.9}  & \textbf{88.8}                \\
\toprule
\end{tabular}
\caption{Results on the GLUE Dev set. Each result is median and std across several seeds, and the Macro column is a mean score across all tasks. We bolded the best results and underlined the second best results. Fine-tuning is omitted from comparison with other methods and was not bolded for visibility. See Section 4 for details.}
\label{table-glue}
\end{table*}

\begin{table*}[]
\centering
\begin{tabular}{c|cccc|ccccc}
\toprule
 & \thead{RTE} & \thead{MNLI, \\ QQP} & \thead{QNLI} & \thead{Other\\Tasks} & \thead{WiC} & \thead{CB, \\ COPA, \\ WSC} & \thead{MultiRC} & \thead{Other\\Tasks} \\ \toprule
Epochs & $200$ & $5$ & $10$ & $100$ & $500$ & $500$ & $10$ & $100$ \\
Patience & $20$ & $2$ & $2$ & $10$ & $20$ & $100$ & $4$ & $10$ \\ \toprule
\end{tabular}
\caption{The number of maximum epochs used for each GLUE and SuperGLUE Task. Once the Dev score stopped increasing for "patience" steps, training was halted. See Section 4 for more details.}
\label{table-epochs}
\end{table*}

\begin{figure*}[h!]
  \centering

    \medskip
    \begin{subfigure}[t]{.24\linewidth}
    \centering\includegraphics[width=\linewidth]{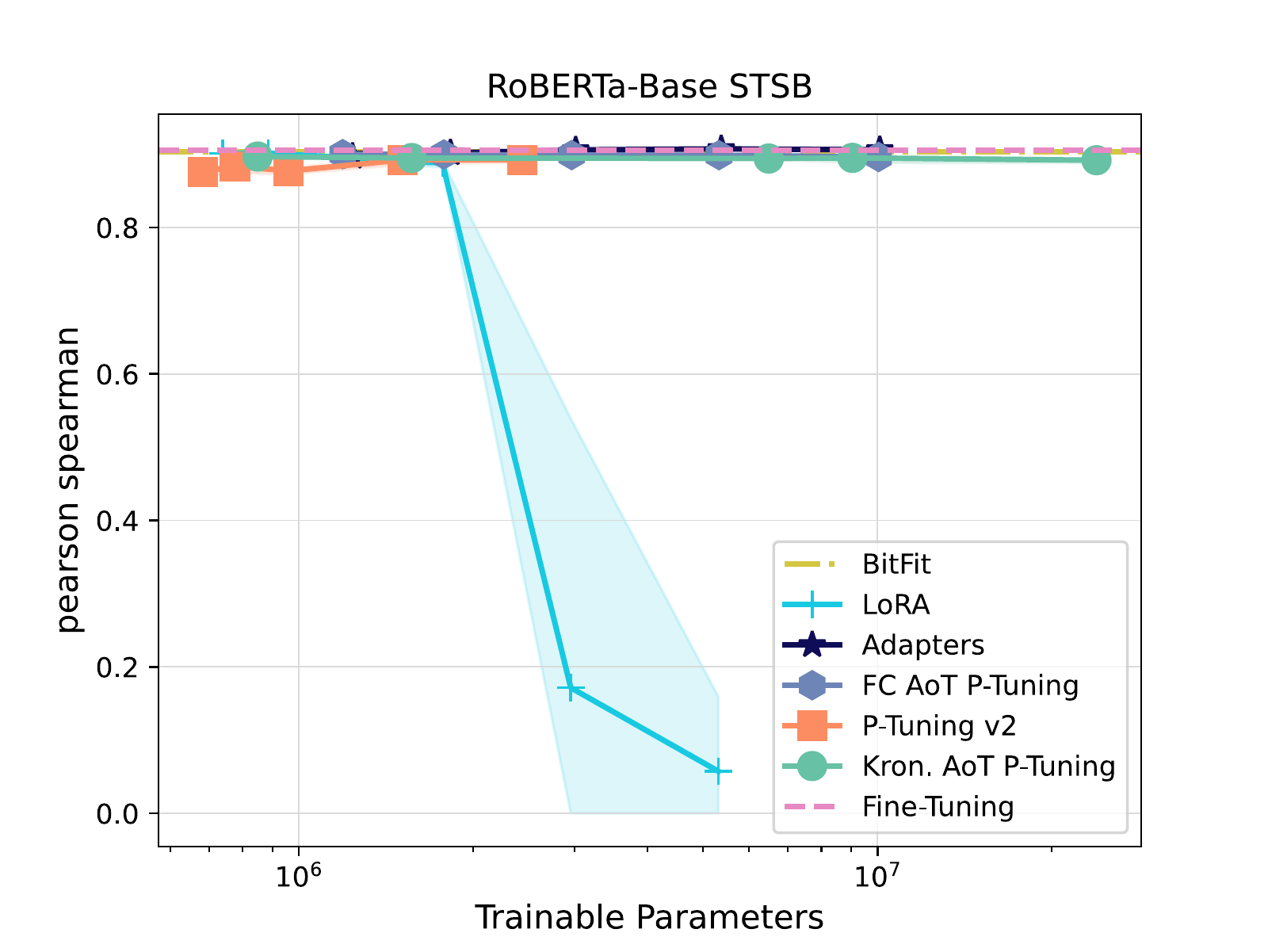}
    \caption{}
  \end{subfigure}
    \begin{subfigure}[t]{.24\linewidth}
    \centering\includegraphics[width=\linewidth]{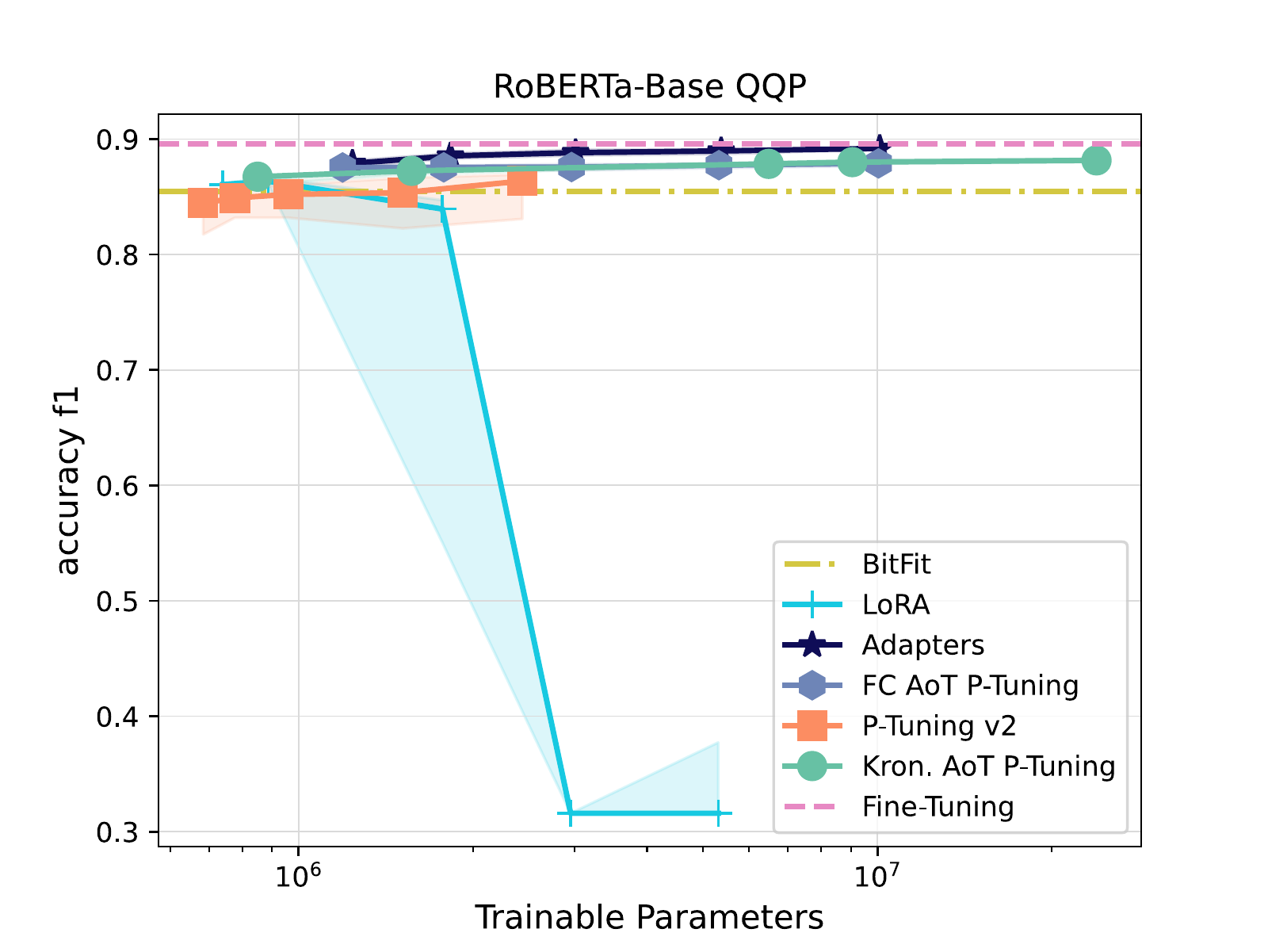}
    \caption{}
  \end{subfigure}
    \begin{subfigure}[t]{.24\linewidth}
    \centering\includegraphics[width=\linewidth]{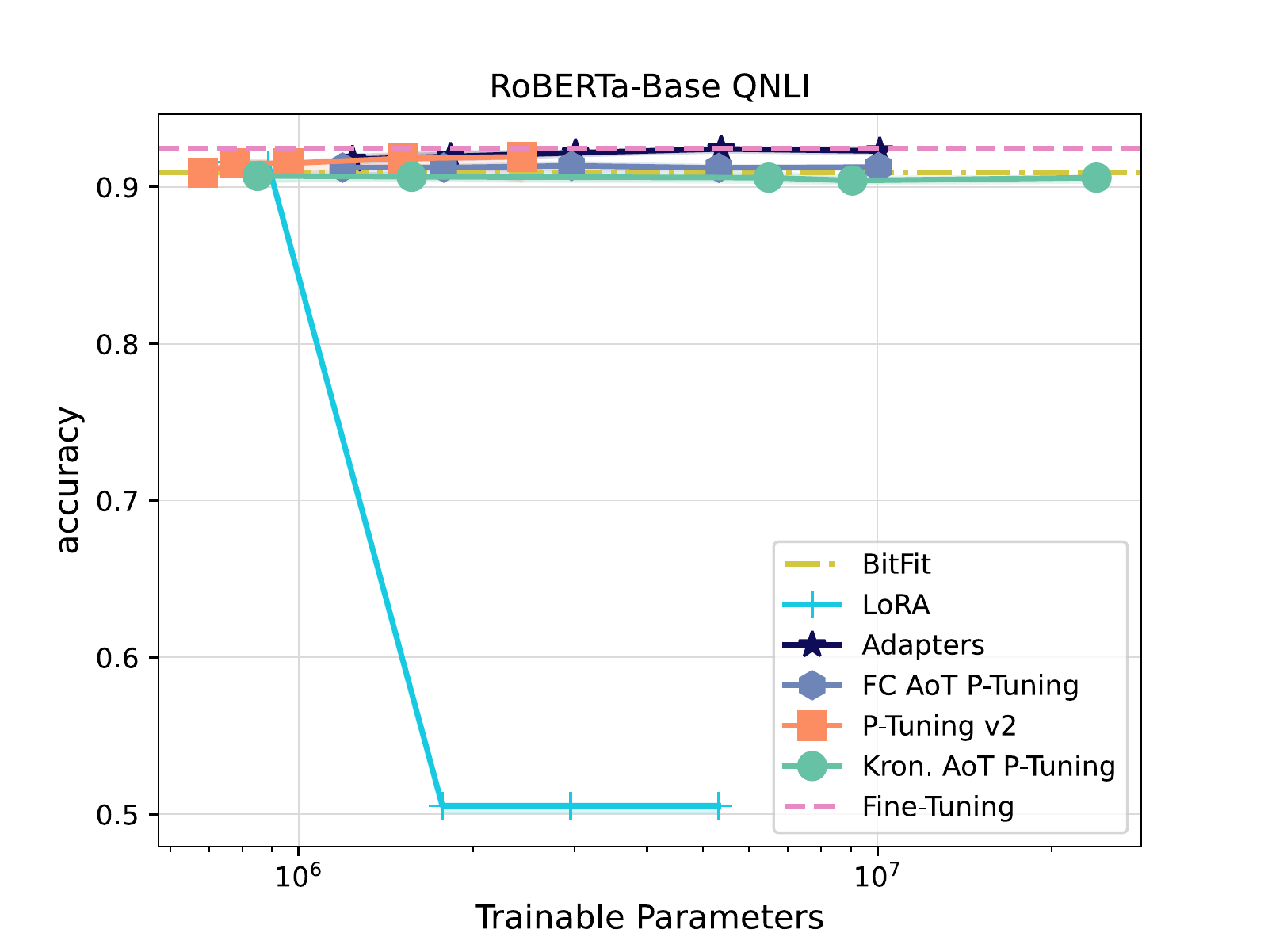}
    \caption{}
  \end{subfigure}
      \begin{subfigure}[t]{.24\linewidth}
    \centering\includegraphics[width=\linewidth]{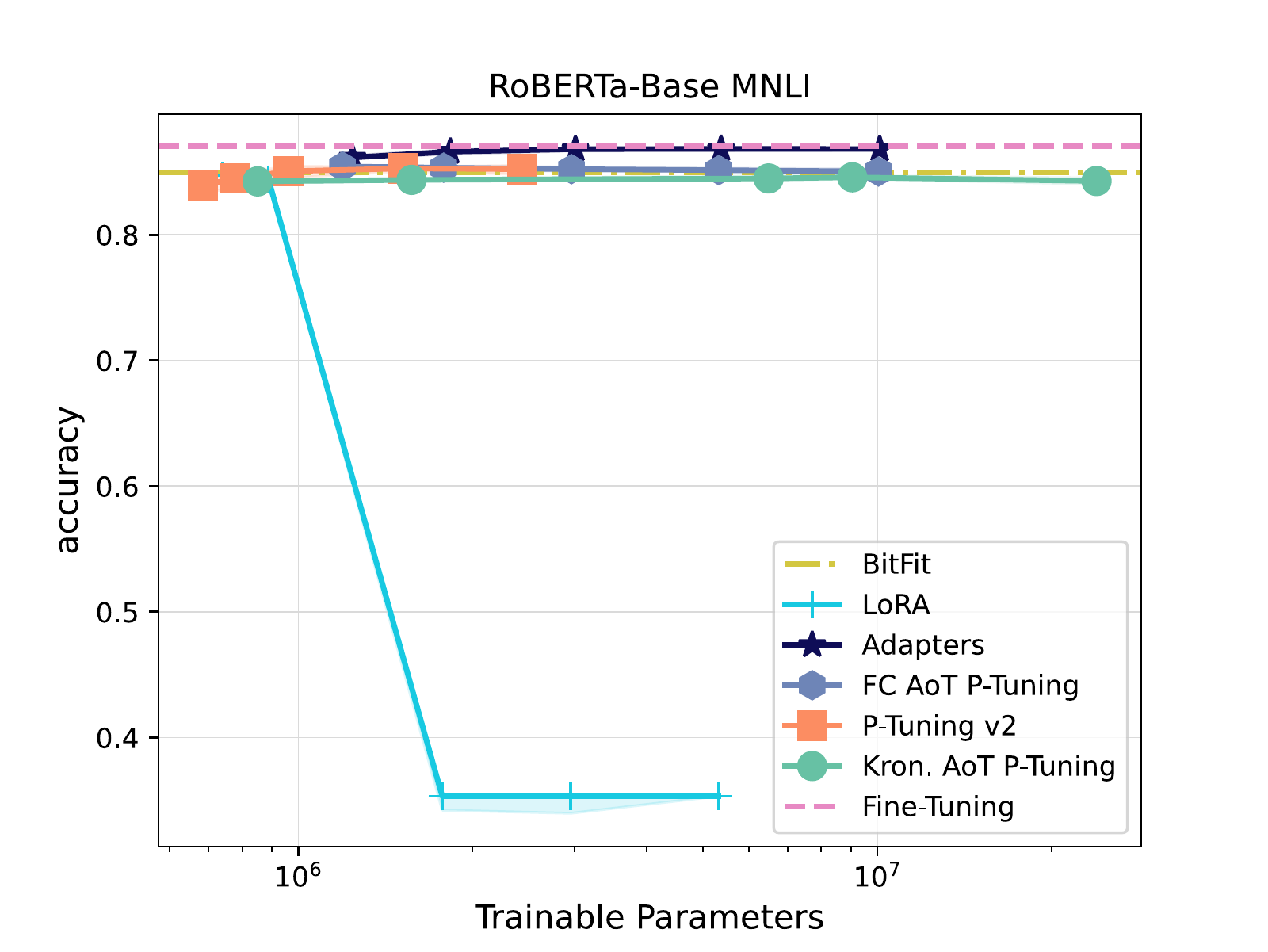}
    \caption{}
  \end{subfigure}
      \begin{subfigure}[t]{.24\linewidth}
    \centering\includegraphics[width=\linewidth]{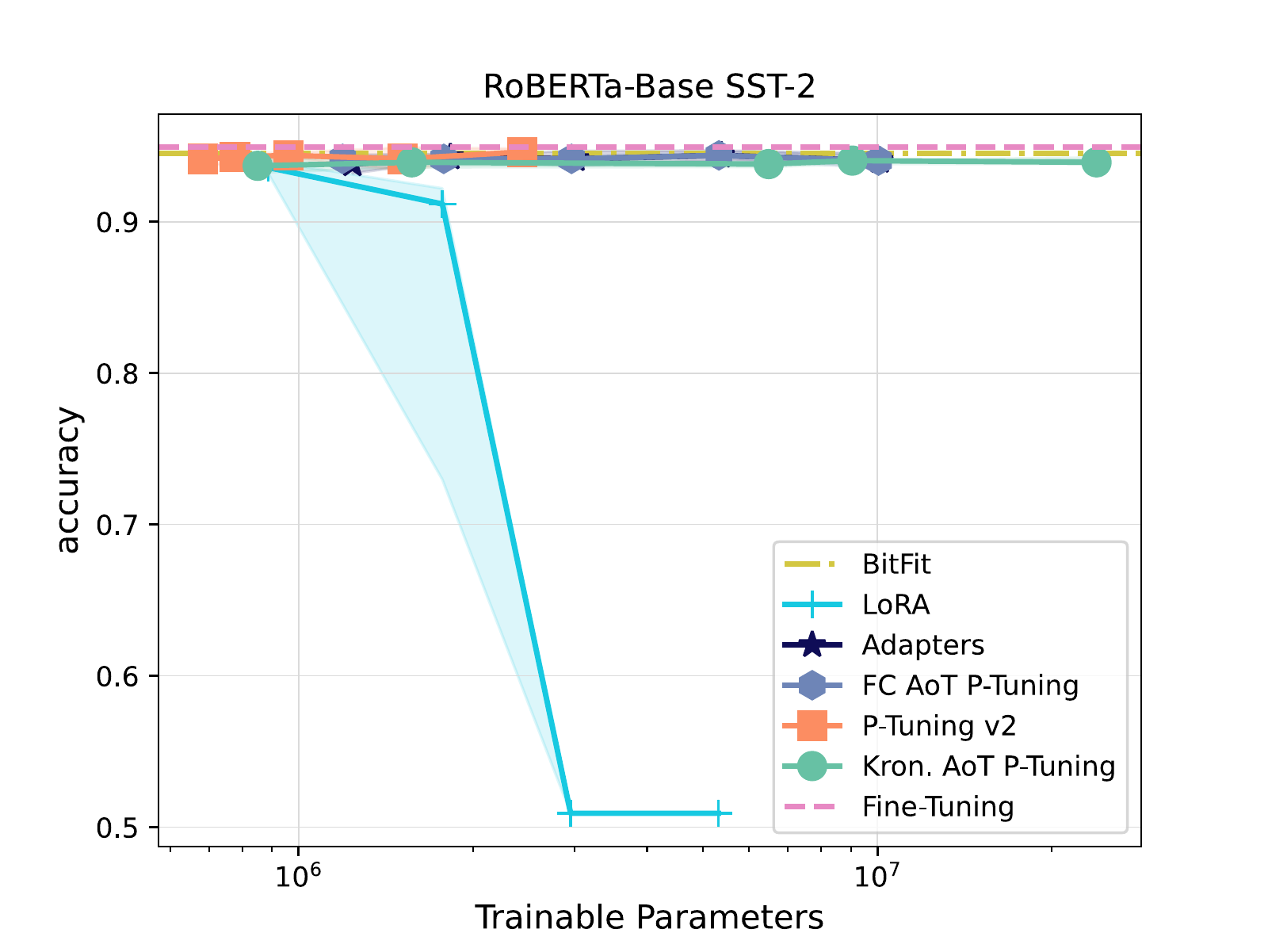}
    \caption{}
  \end{subfigure}
      \begin{subfigure}[t]{.24\linewidth}
    \centering\includegraphics[width=\linewidth]{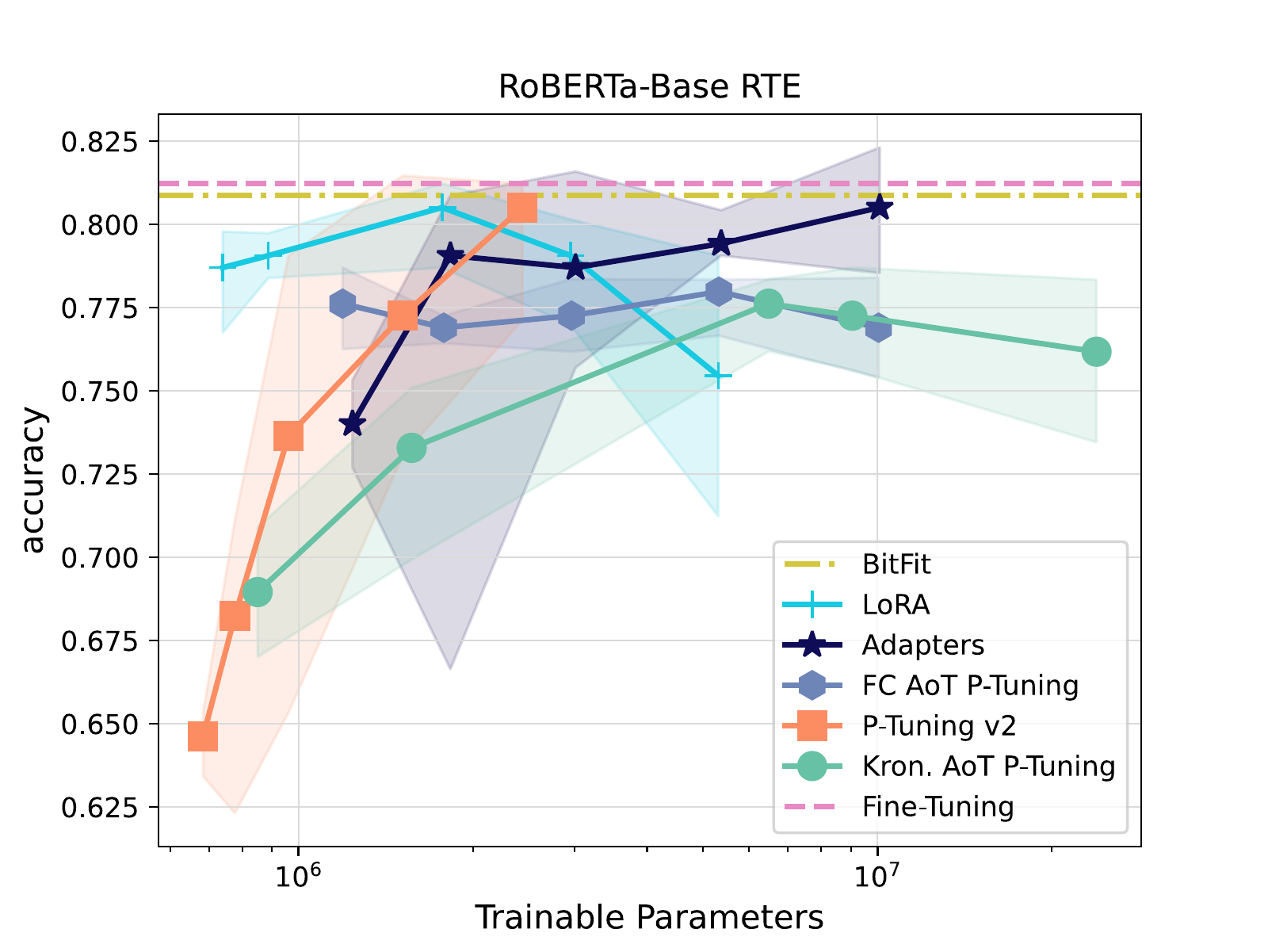}
    \caption{}
  \end{subfigure}
      \begin{subfigure}[t]{.24\linewidth}
    \centering\includegraphics[width=\linewidth]{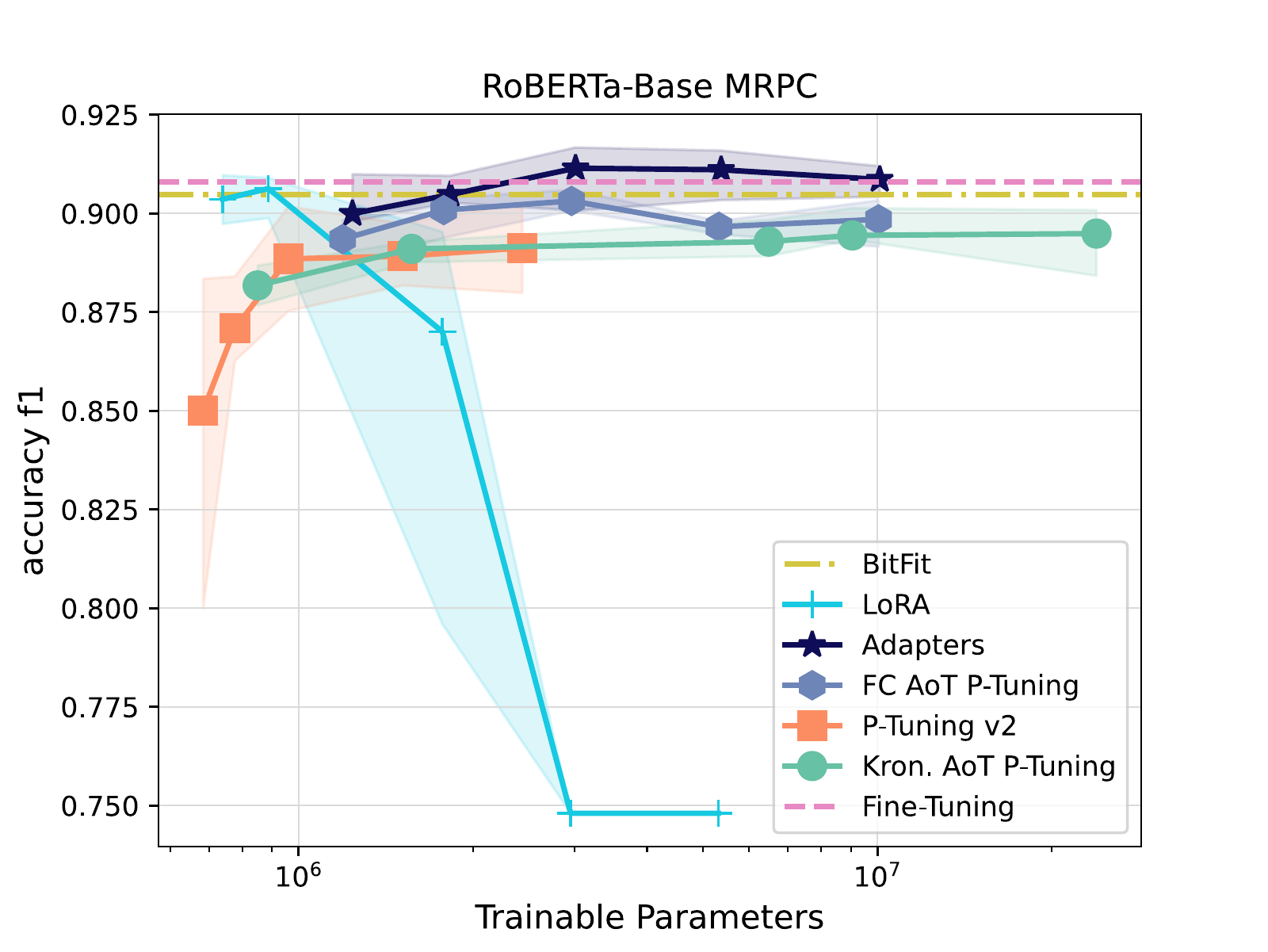}
    \caption{}
  \end{subfigure}
      \begin{subfigure}[t]{.24\linewidth}
    \centering\includegraphics[width=\linewidth]{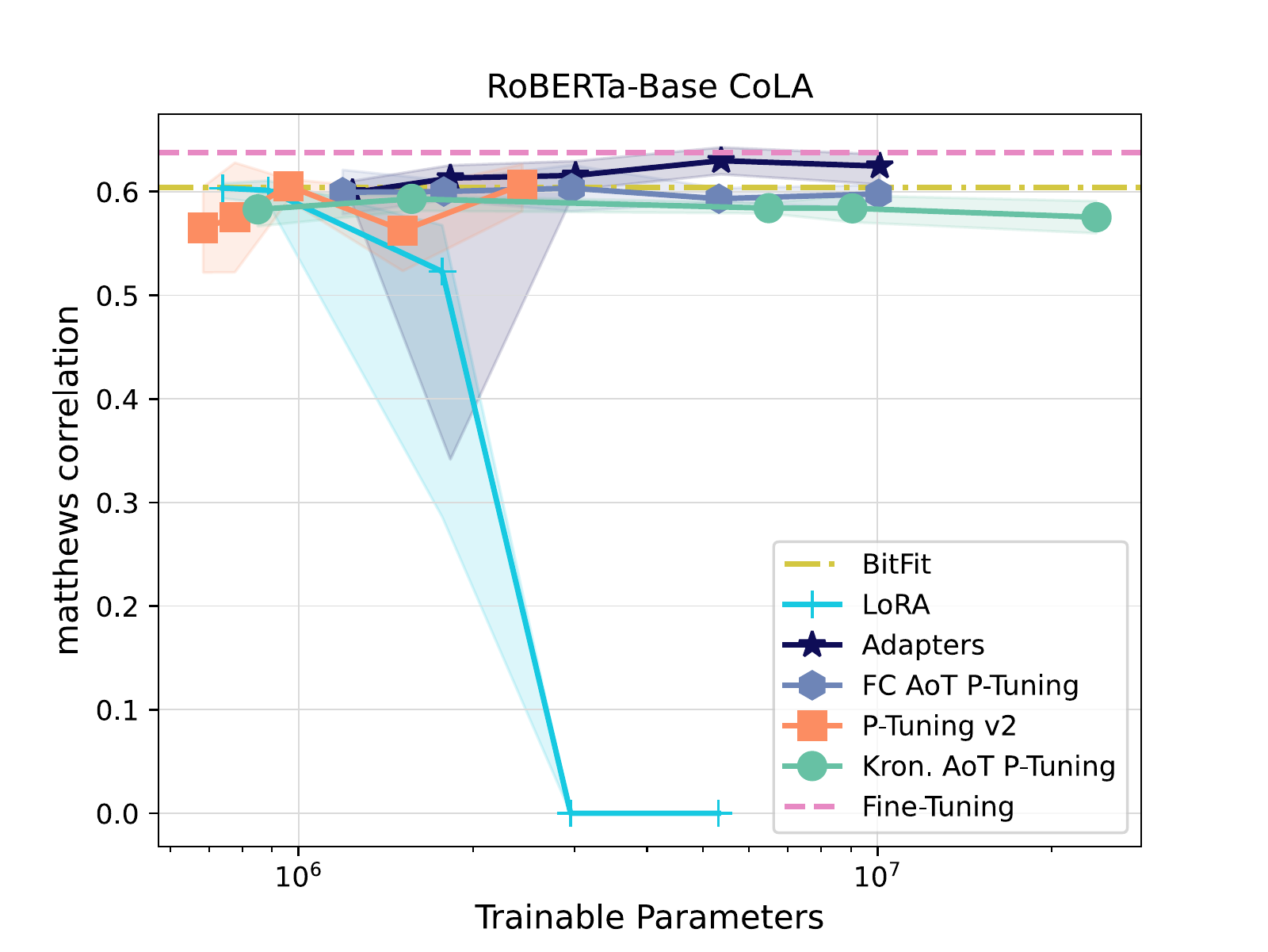}
    \caption{}
  \end{subfigure}
   \medskip
    \begin{subfigure}[t]{.24\linewidth}
    \centering\includegraphics[width=\linewidth]{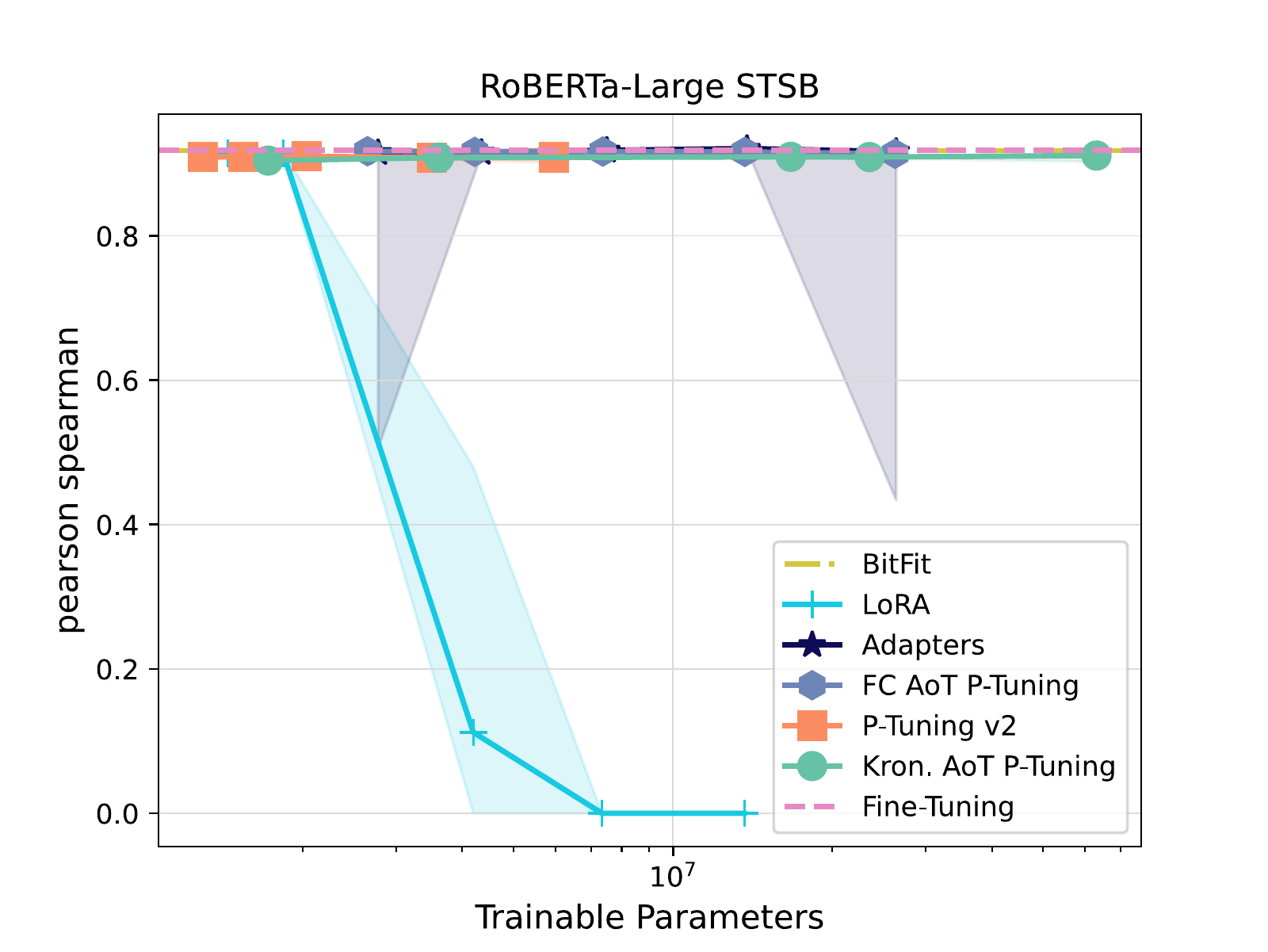}
    \caption{}
  \end{subfigure}
    \begin{subfigure}[t]{.24\linewidth}
    \centering\includegraphics[width=\linewidth]{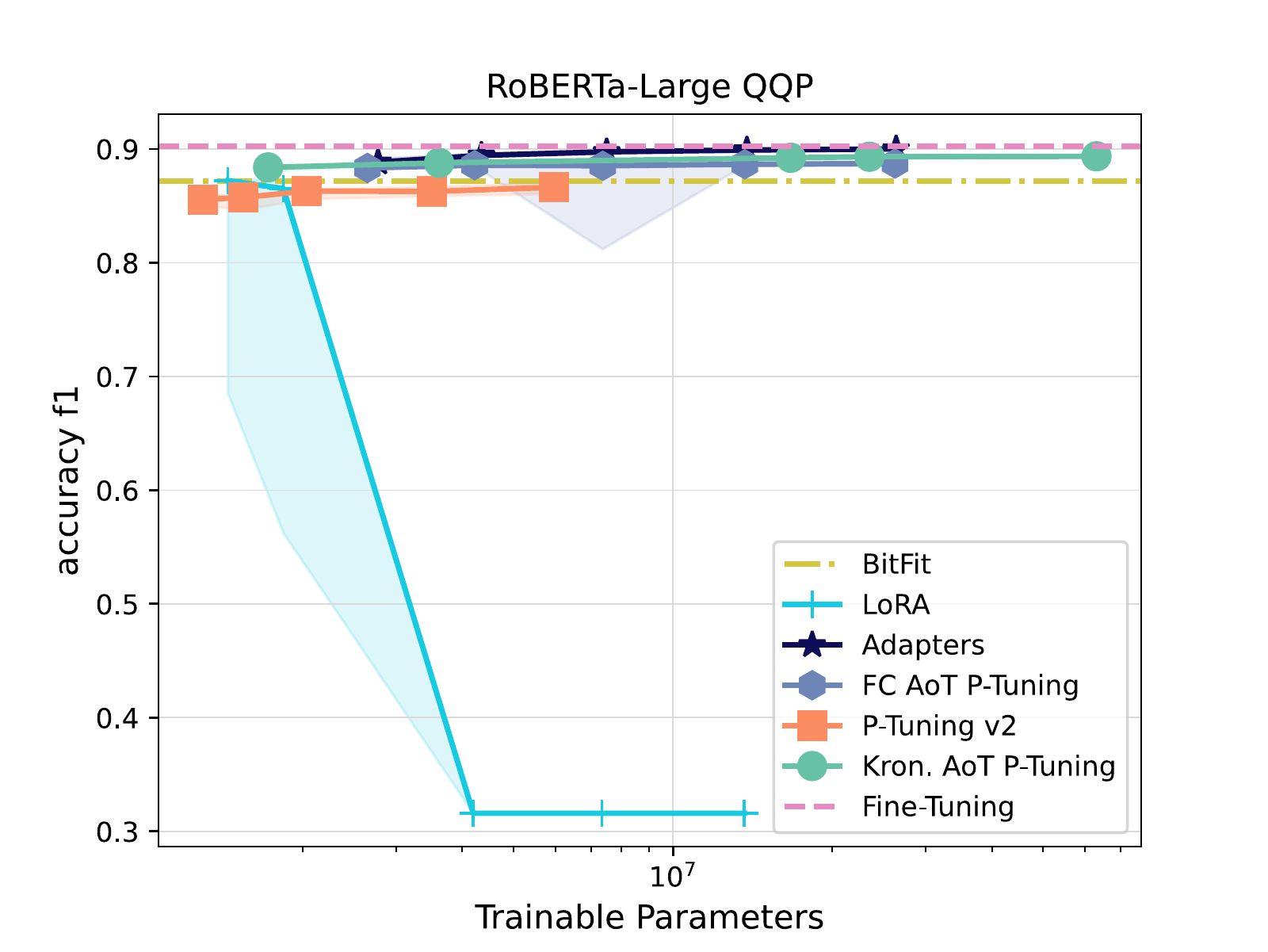}
    \caption{}
  \end{subfigure}
    \begin{subfigure}[t]{.24\linewidth}
    \centering\includegraphics[width=\linewidth]{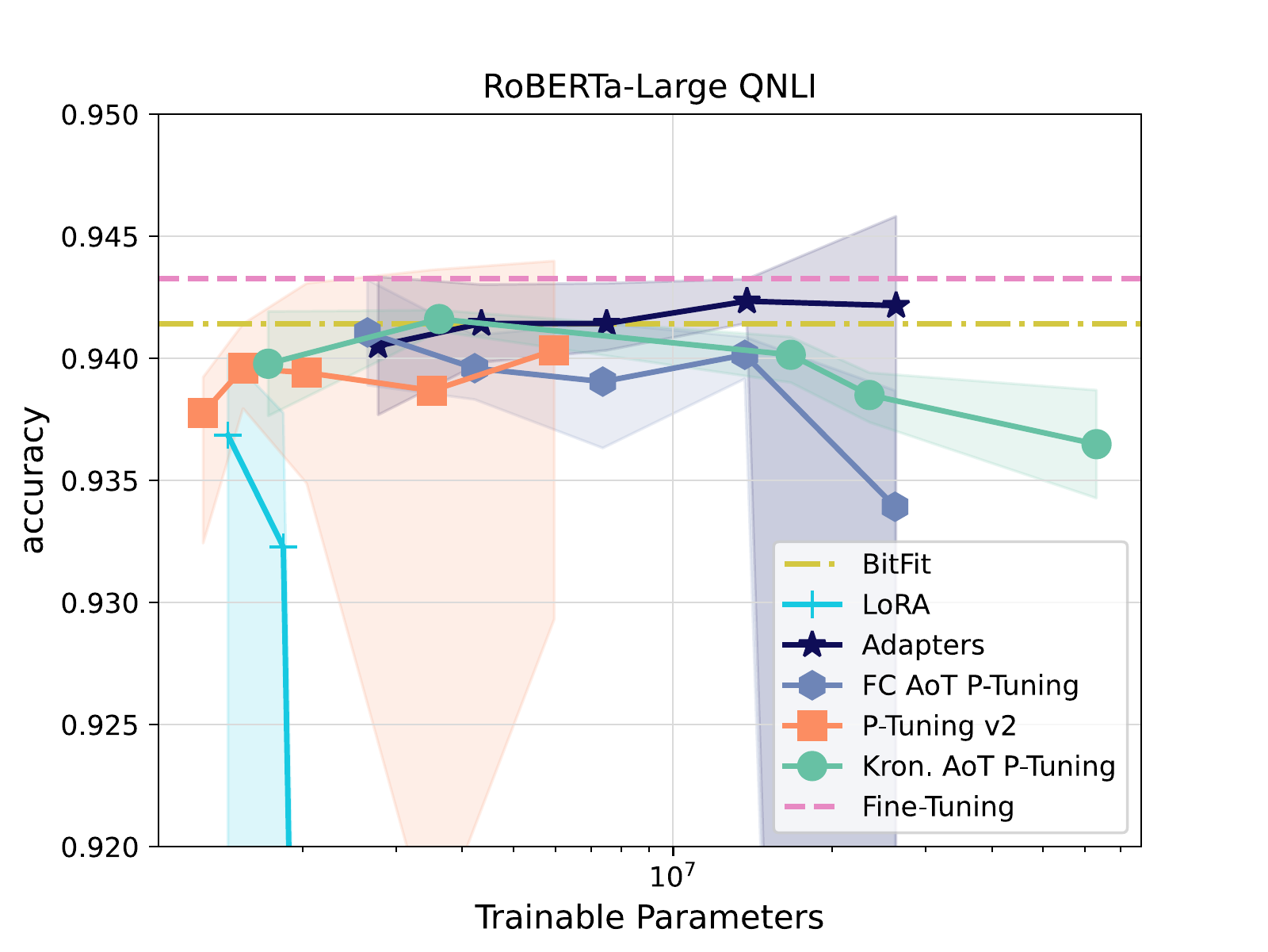}
    \caption{}
  \end{subfigure}
      \begin{subfigure}[t]{.24\linewidth}
    \centering\includegraphics[width=\linewidth]{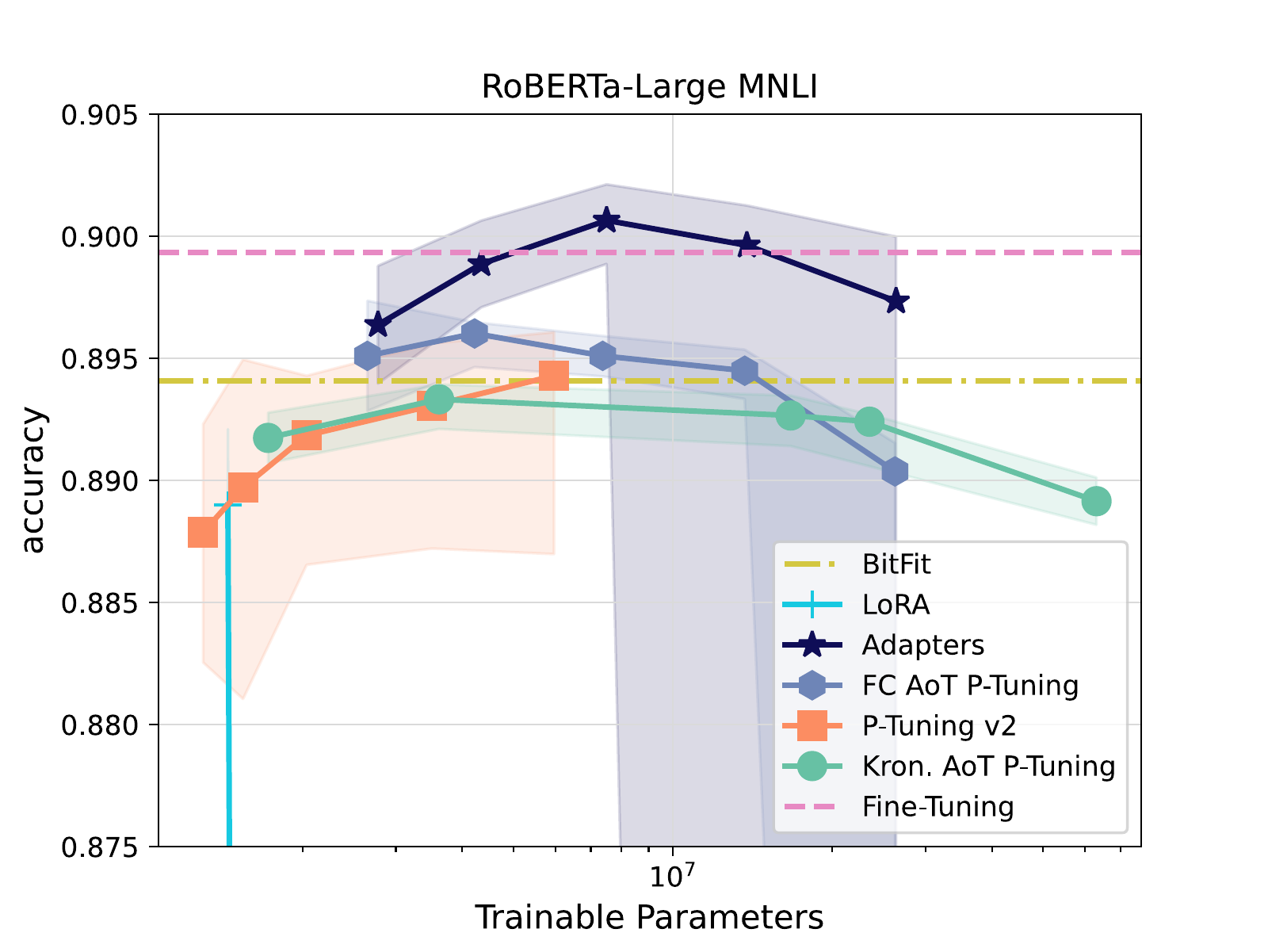}
    \caption{}
  \end{subfigure}
      \begin{subfigure}[t]{.24\linewidth}
    \centering\includegraphics[width=\linewidth]{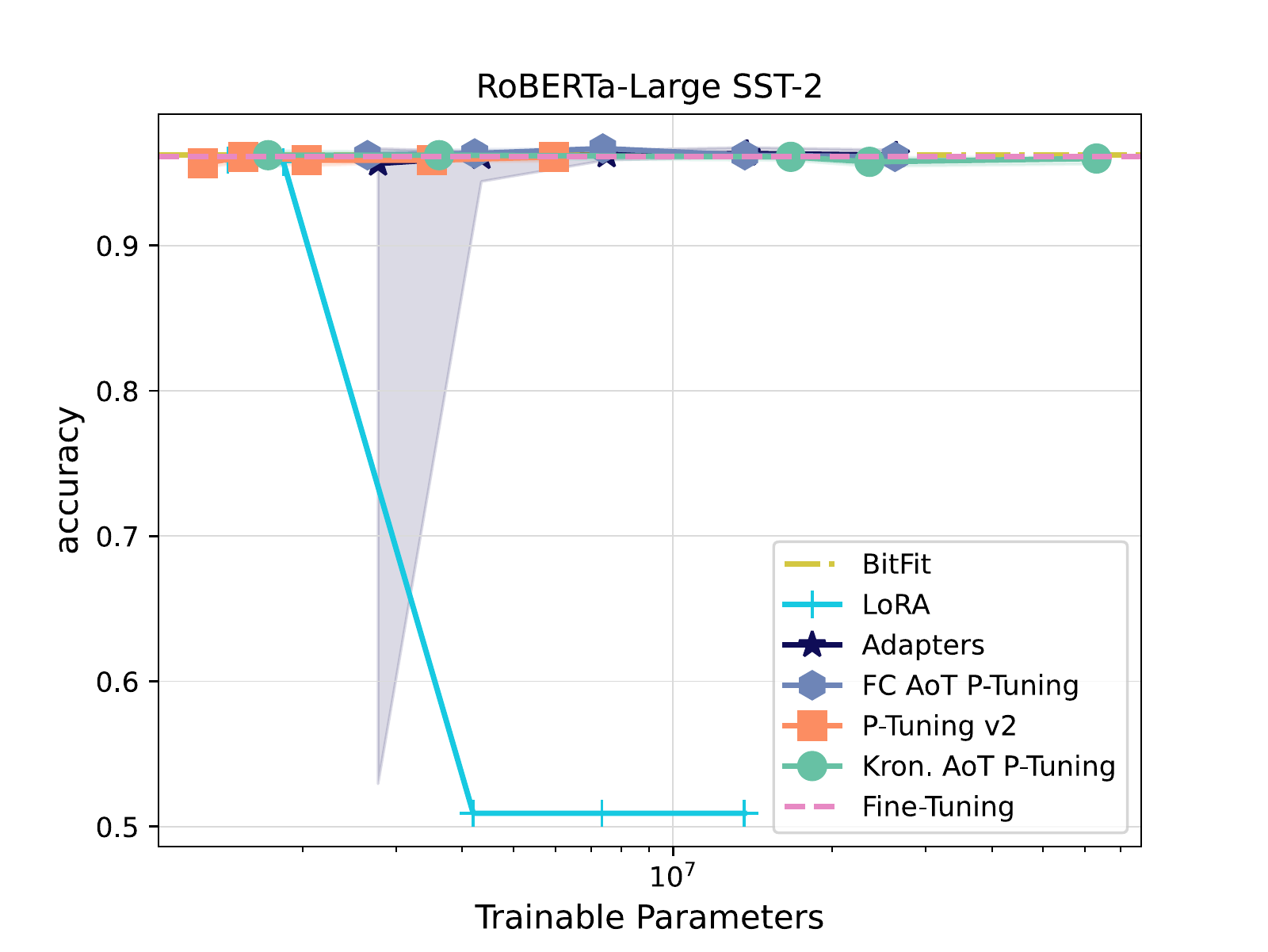}
    \caption{}
  \end{subfigure}
      \begin{subfigure}[t]{.24\linewidth}
    \centering\includegraphics[width=\linewidth]{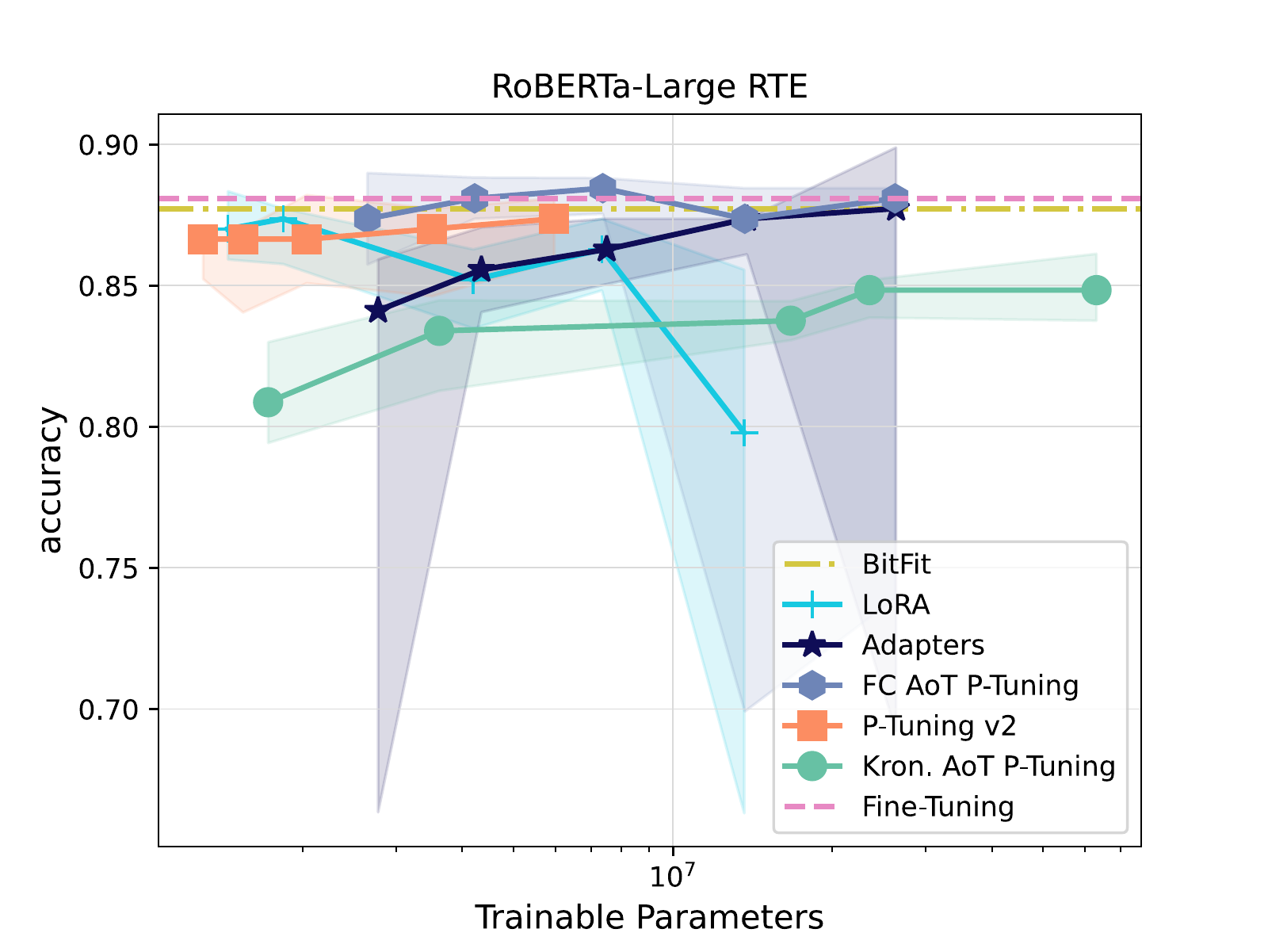}
    \caption{}
  \end{subfigure}
      \begin{subfigure}[t]{.24\linewidth}
    \centering\includegraphics[width=\linewidth]{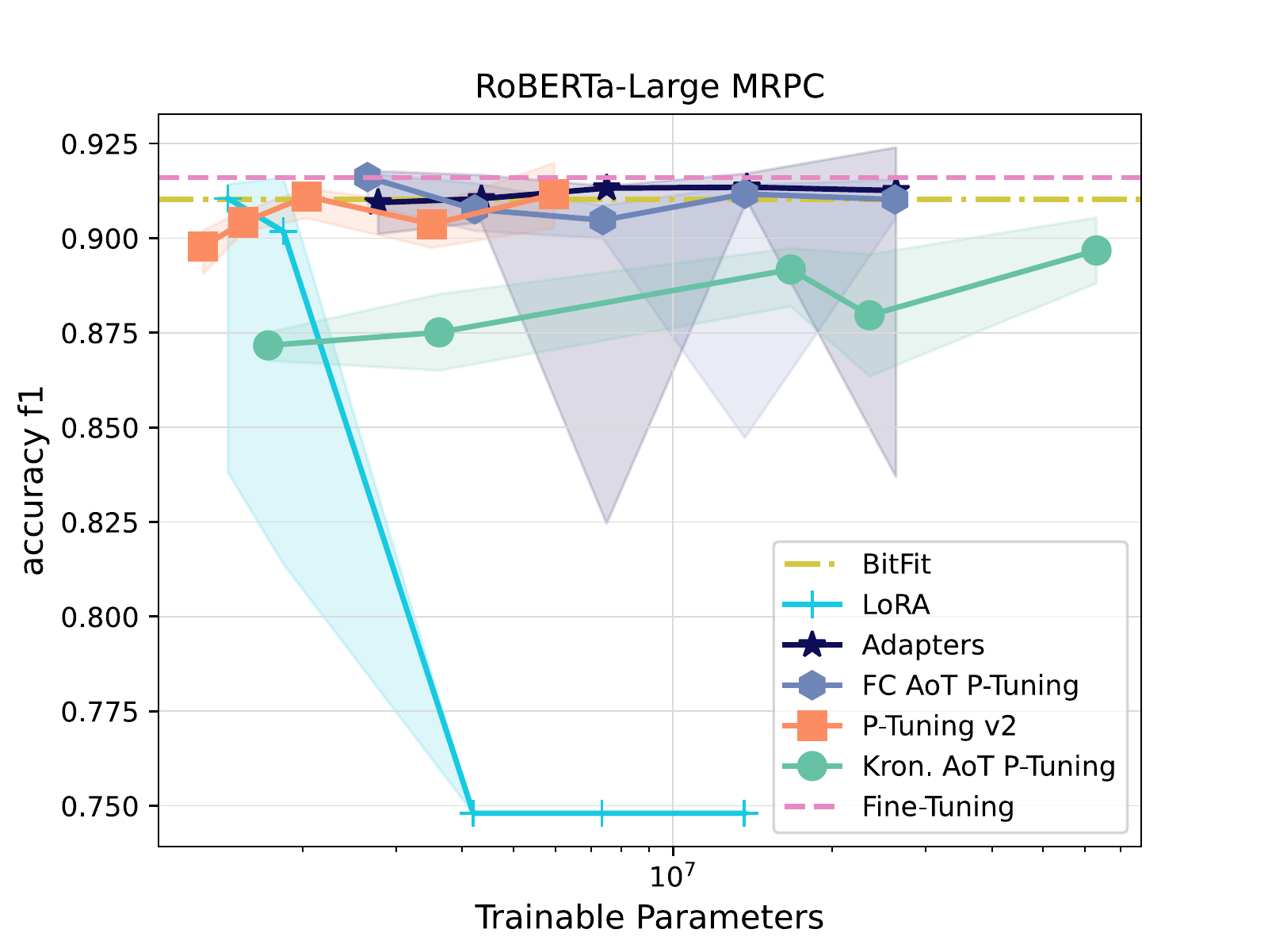}
    \caption{}
  \end{subfigure}
      \begin{subfigure}[t]{.24\linewidth}
    \centering\includegraphics[width=\linewidth]{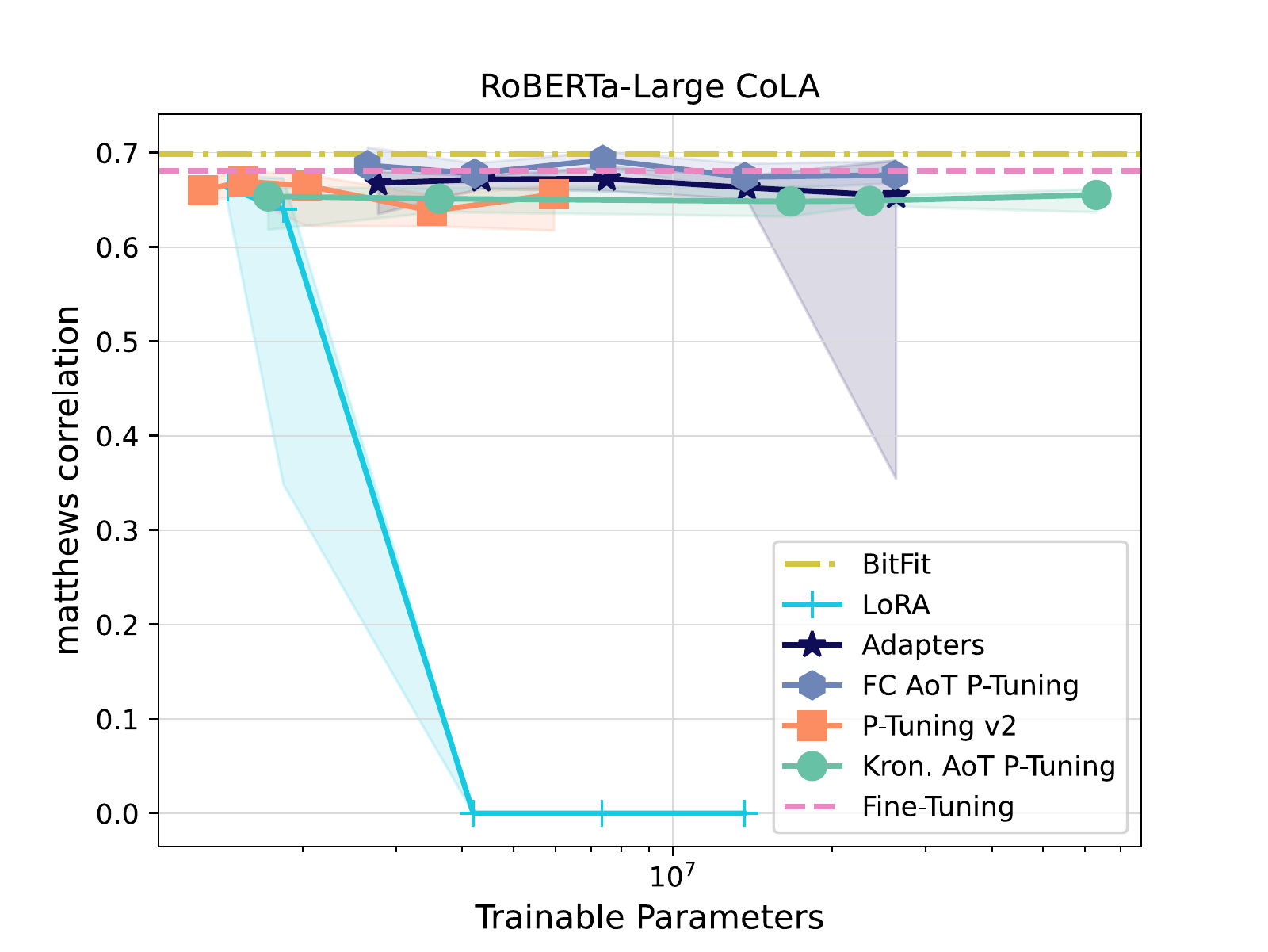}
    \caption{}
  \end{subfigure}
  \caption{Per-task GLUE Benchmarking Dataset results for a different number of trained parameters of P-Tuning v2 and AoT P-Tuning with RoBERTa-Base (a-h) and RoBERTa-Large (i-p). We also provide results of plain fine-tuning for reference. See Section 4 for more details.}
  \label{figure-parameters-glue}
\end{figure*}

\begin{figure*}[h!]
  \centering

  \medskip
    \begin{subfigure}[t]{.24\linewidth}
    \centering\includegraphics[width=\linewidth]{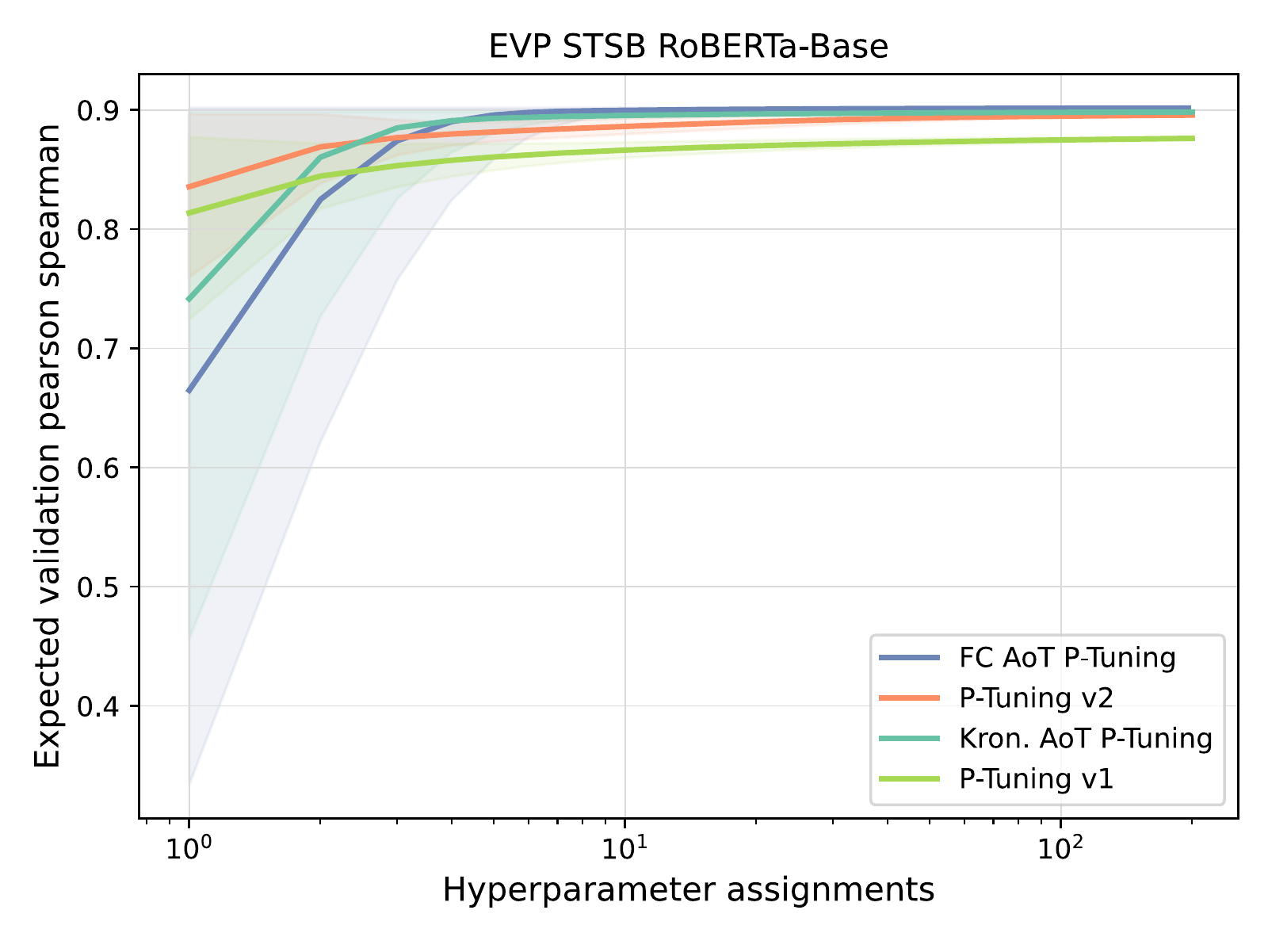}
    \caption{}
  \end{subfigure}
    \begin{subfigure}[t]{.24\linewidth}
    \centering\includegraphics[width=\linewidth]{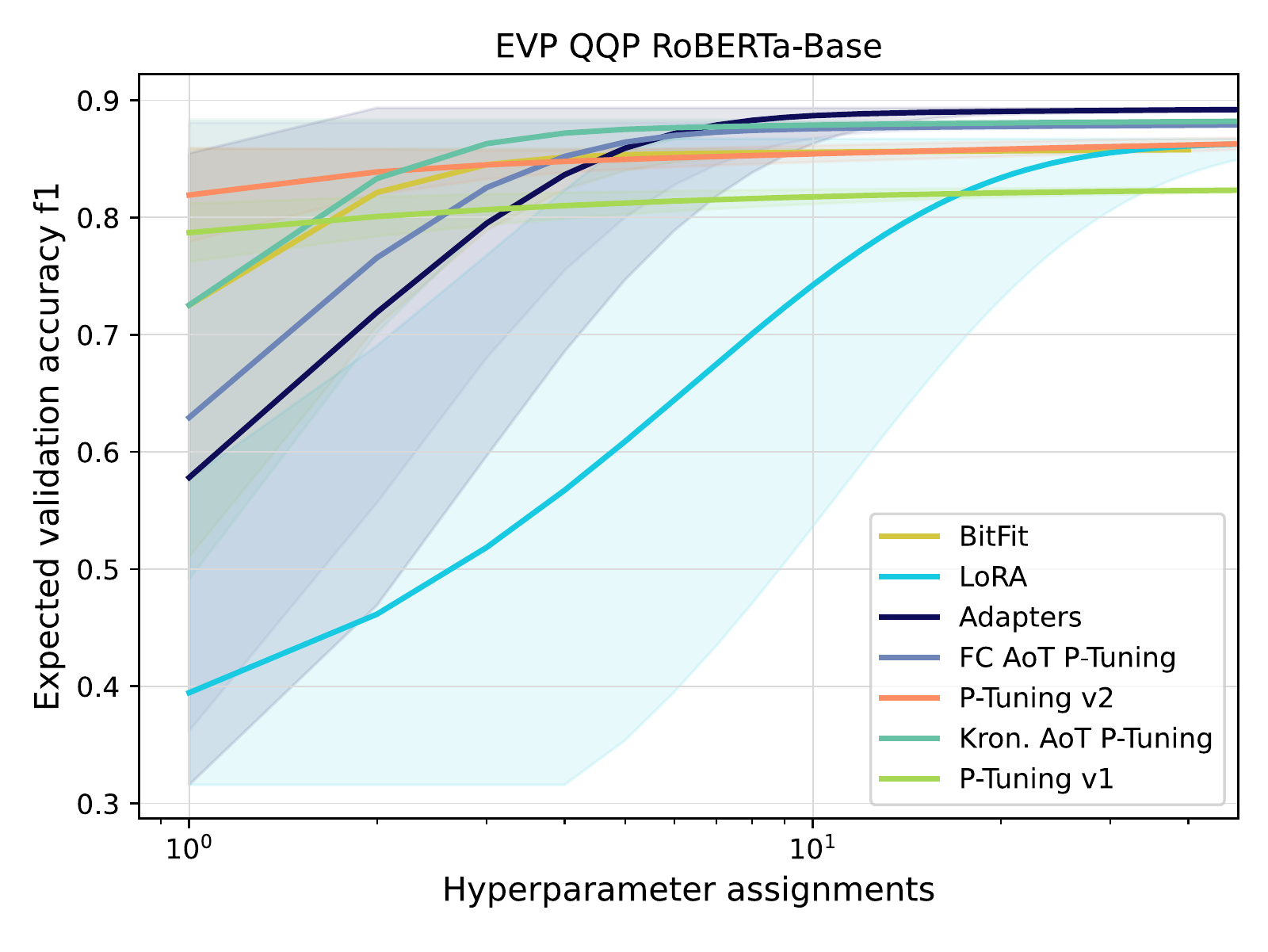}
    \caption{}
  \end{subfigure}
    \begin{subfigure}[t]{.24\linewidth}
    \centering\includegraphics[width=\linewidth]{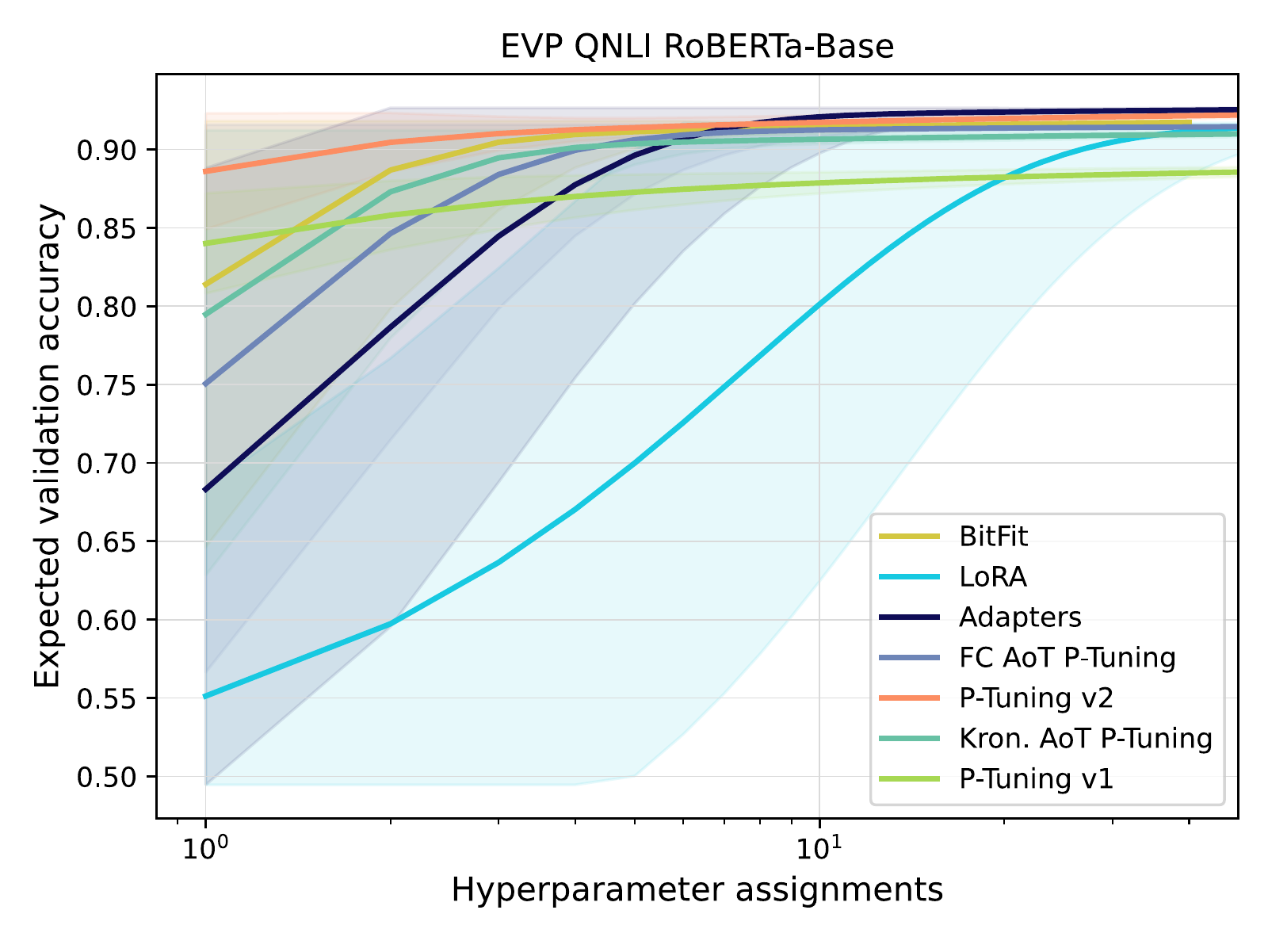}
    \caption{}
  \end{subfigure}
      \begin{subfigure}[t]{.24\linewidth}
    \centering\includegraphics[width=\linewidth]{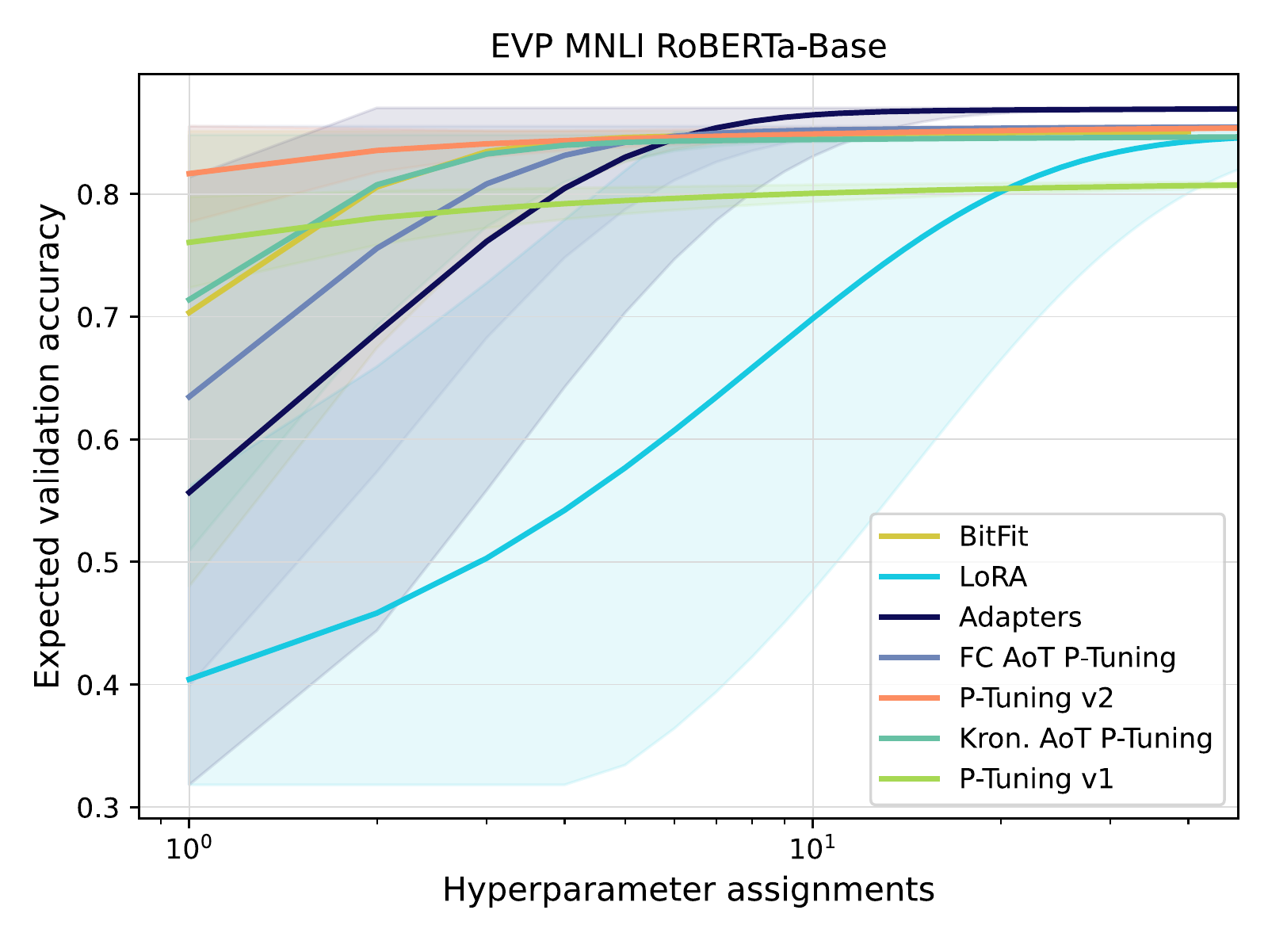}
    \caption{}
  \end{subfigure}
      \begin{subfigure}[t]{.24\linewidth}
    \centering\includegraphics[width=\linewidth]{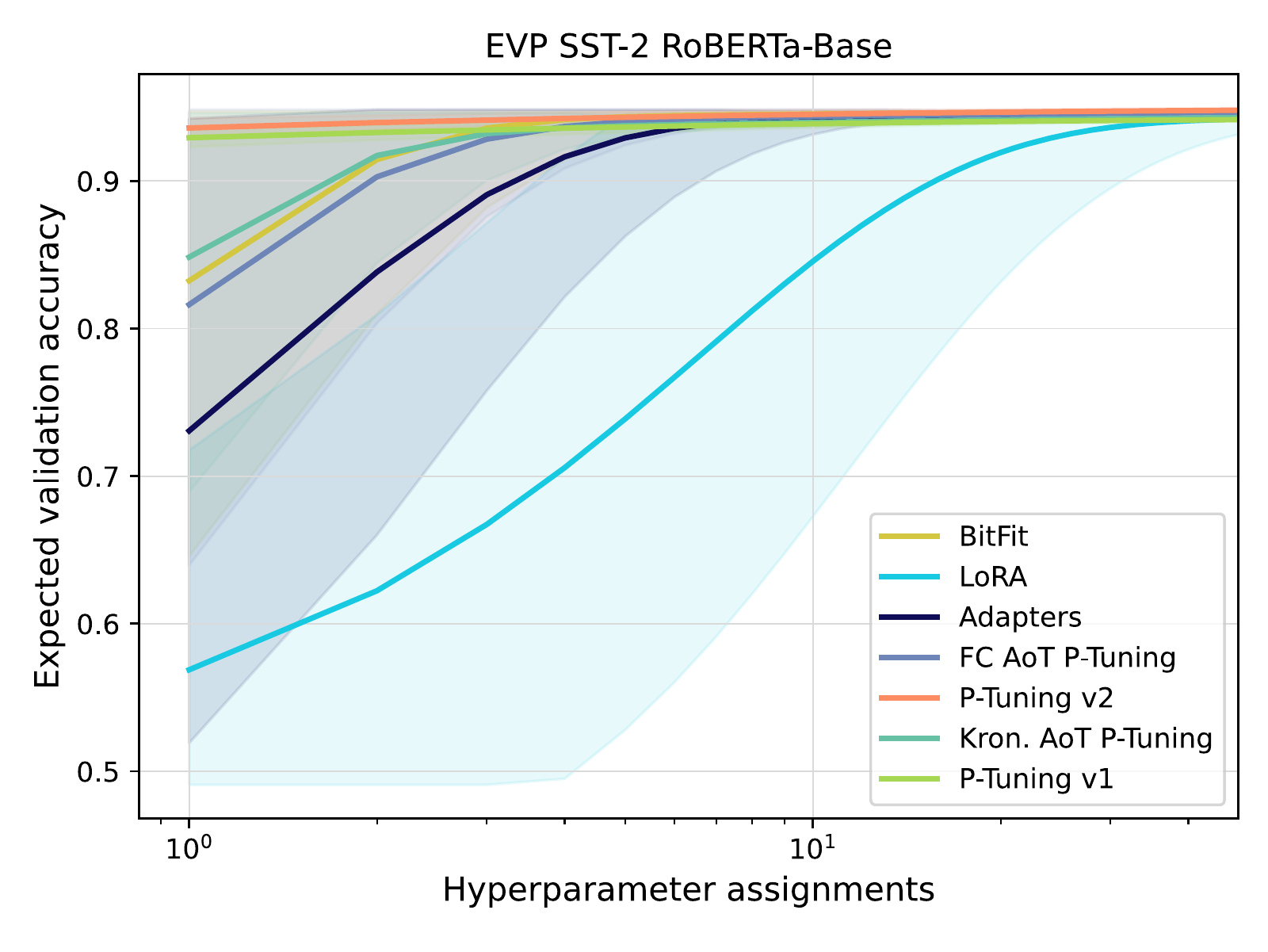}
    \caption{}
  \end{subfigure}
      \begin{subfigure}[t]{.24\linewidth}
    \centering\includegraphics[width=\linewidth]{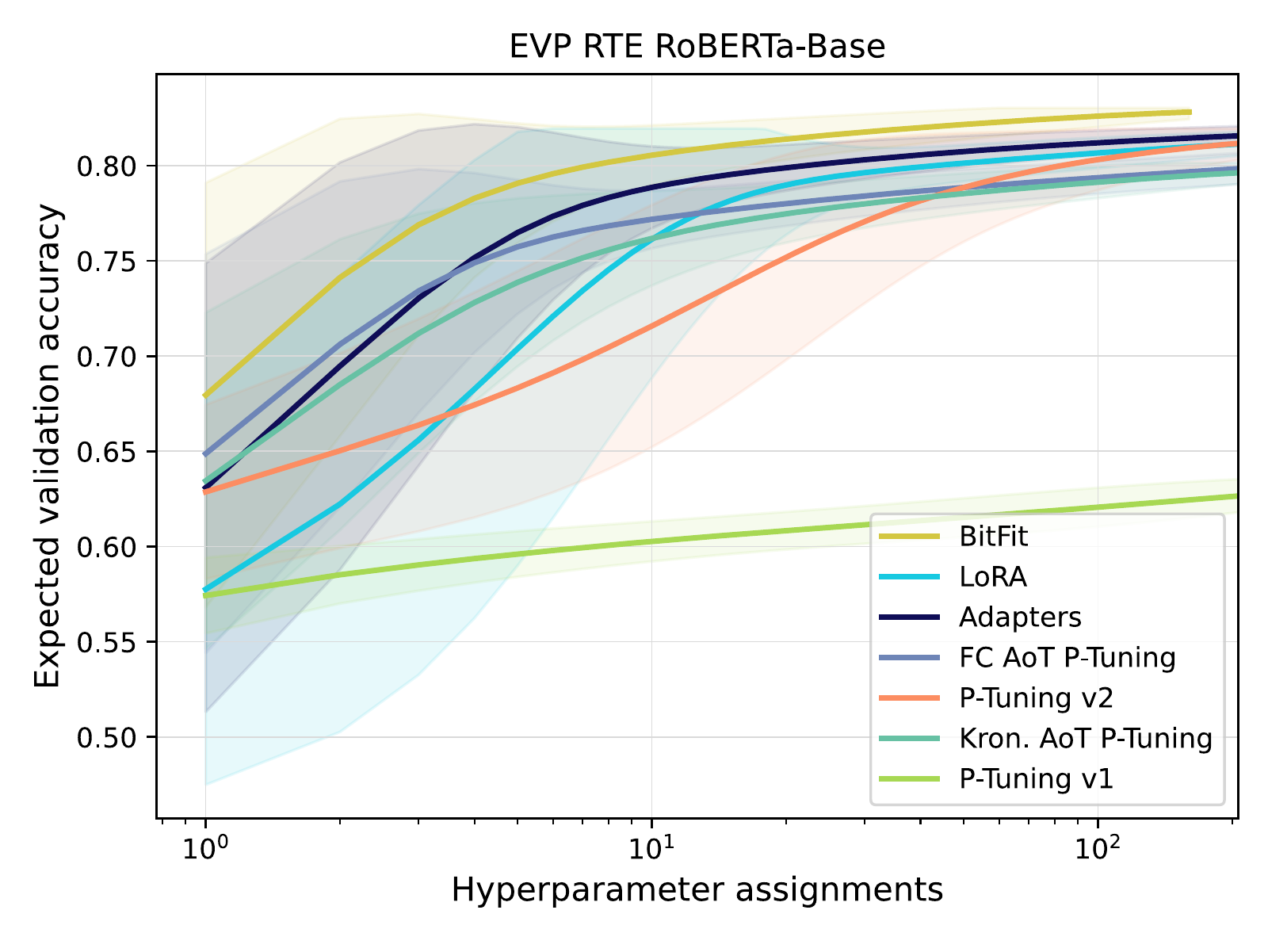}
    \caption{}
  \end{subfigure}
      \begin{subfigure}[t]{.24\linewidth}
    \centering\includegraphics[width=\linewidth]{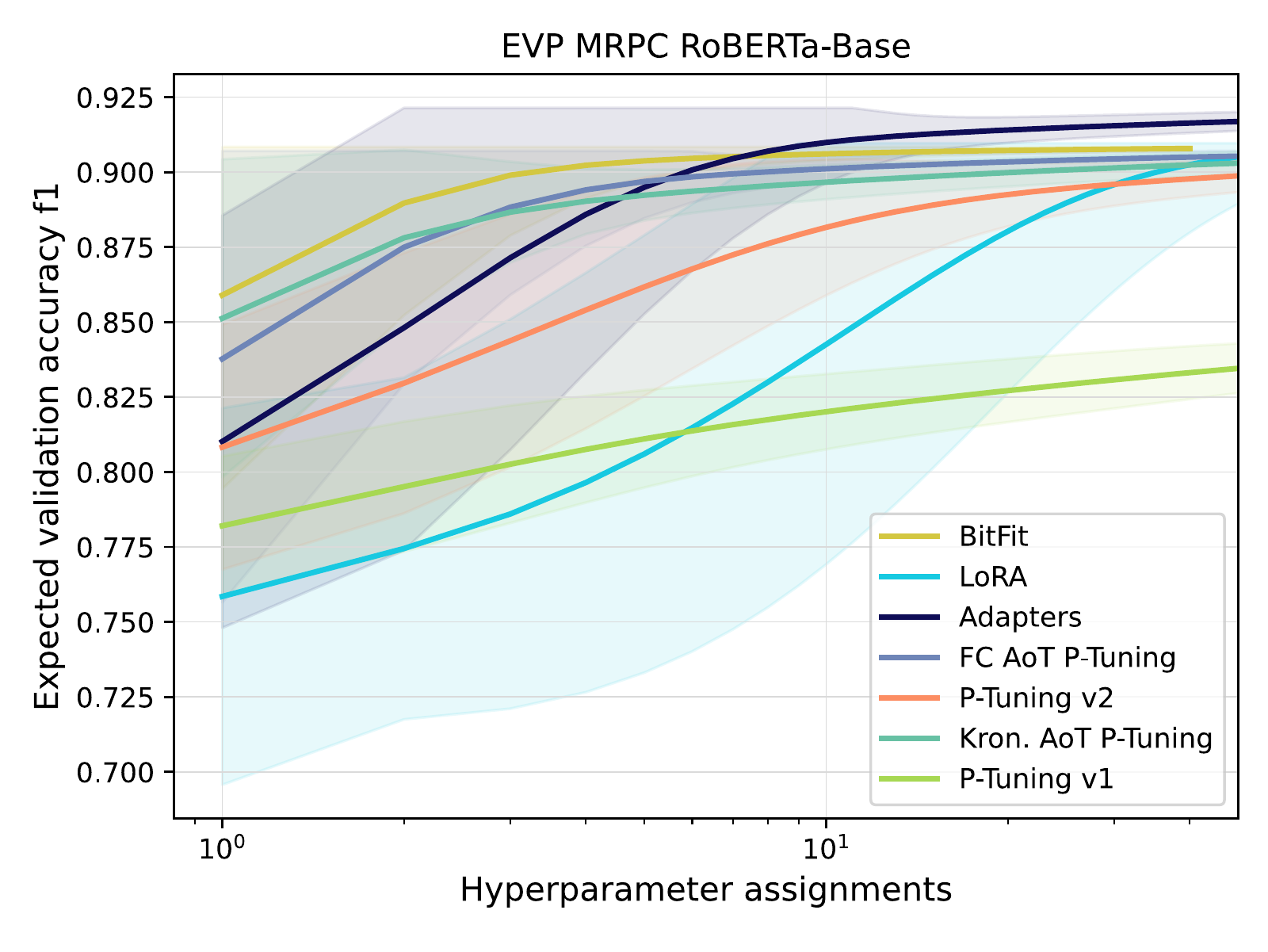}
    \caption{}
  \end{subfigure}
      \begin{subfigure}[t]{.24\linewidth}
    \centering\includegraphics[width=\linewidth]{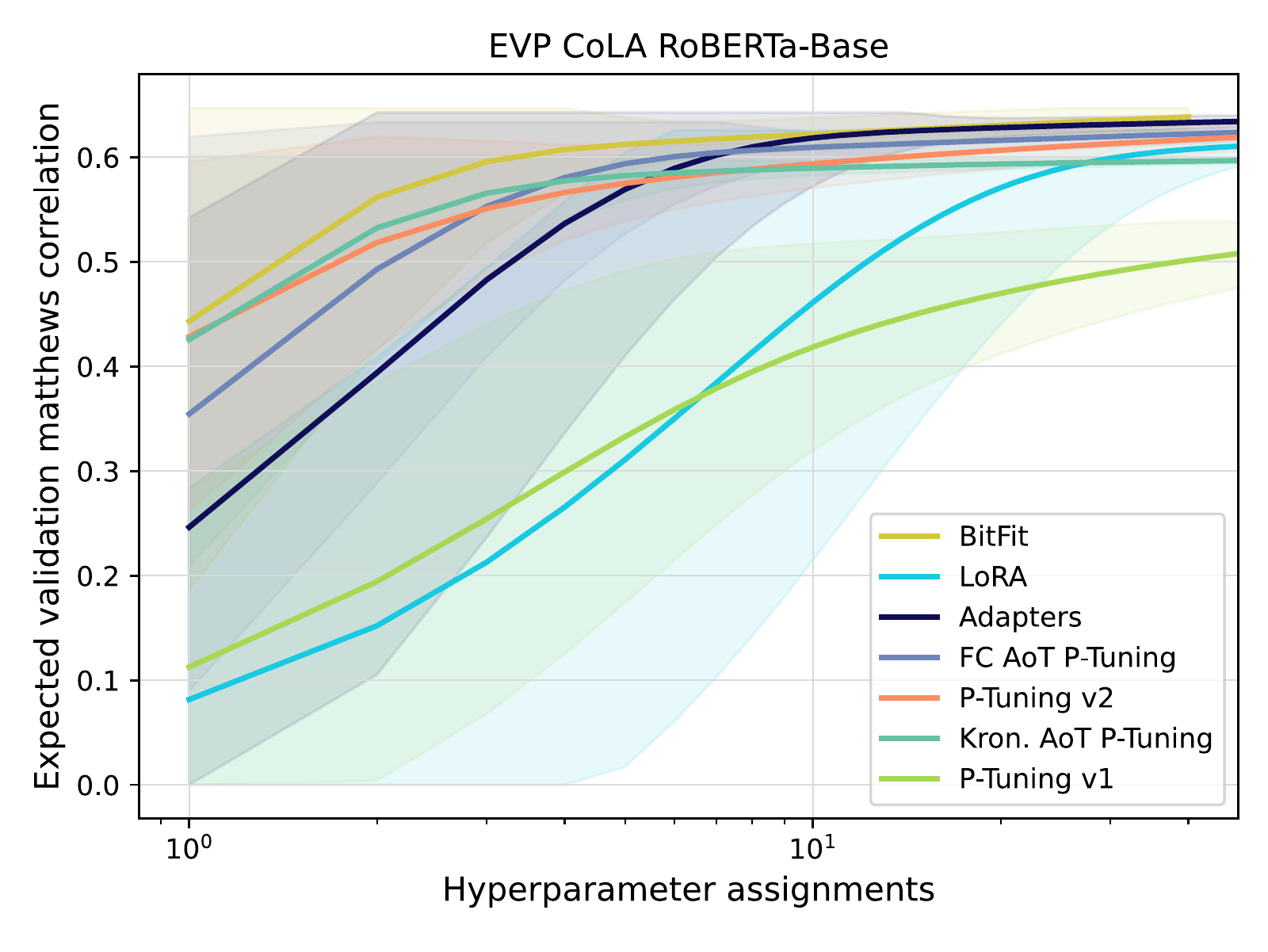}
    \caption{}
  \end{subfigure}
   \medskip
    \begin{subfigure}[t]{.24\linewidth}
    \centering\includegraphics[width=\linewidth]{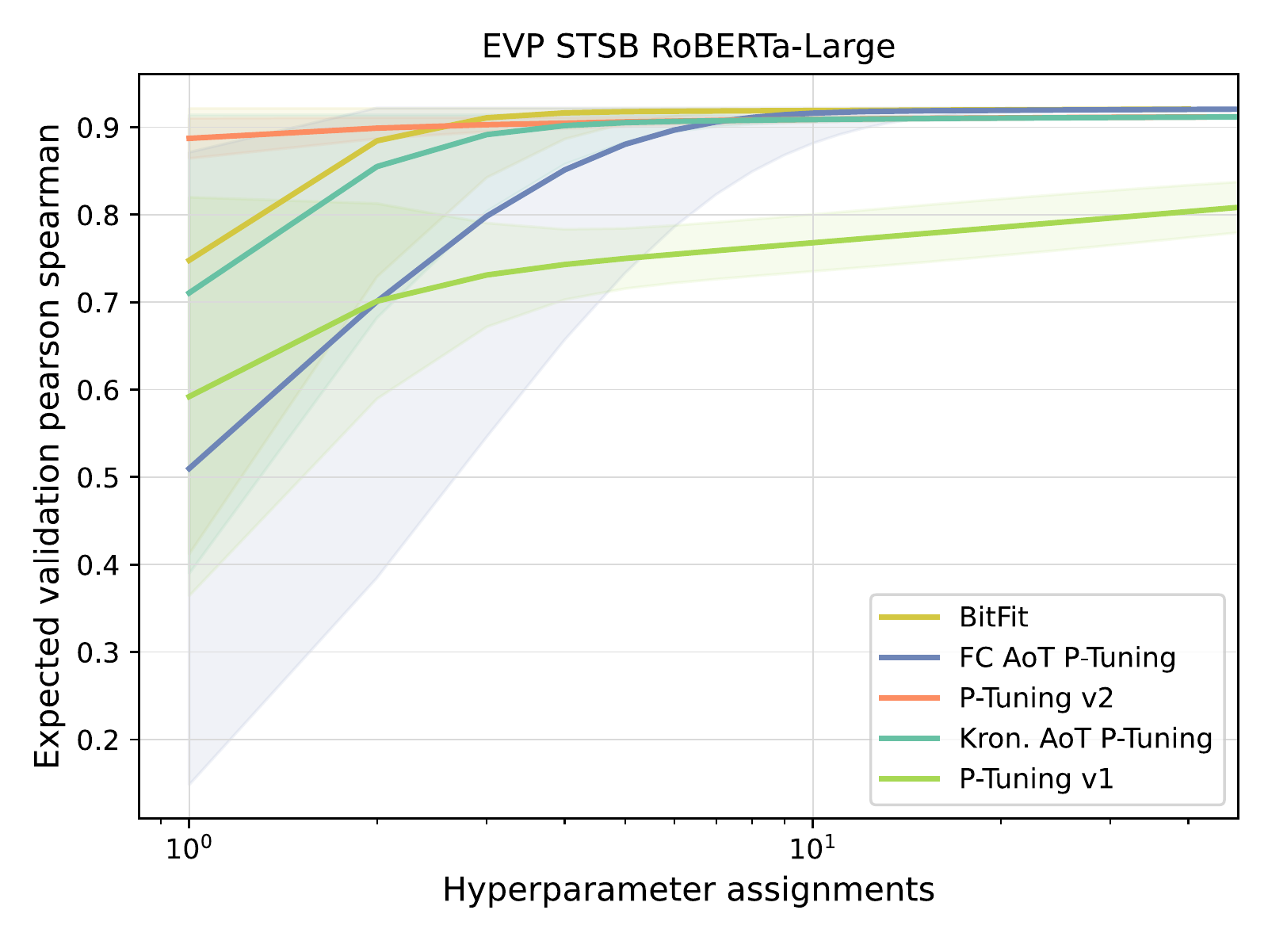}
    \caption{}
  \end{subfigure}
    \begin{subfigure}[t]{.24\linewidth}
    \centering\includegraphics[width=\linewidth]{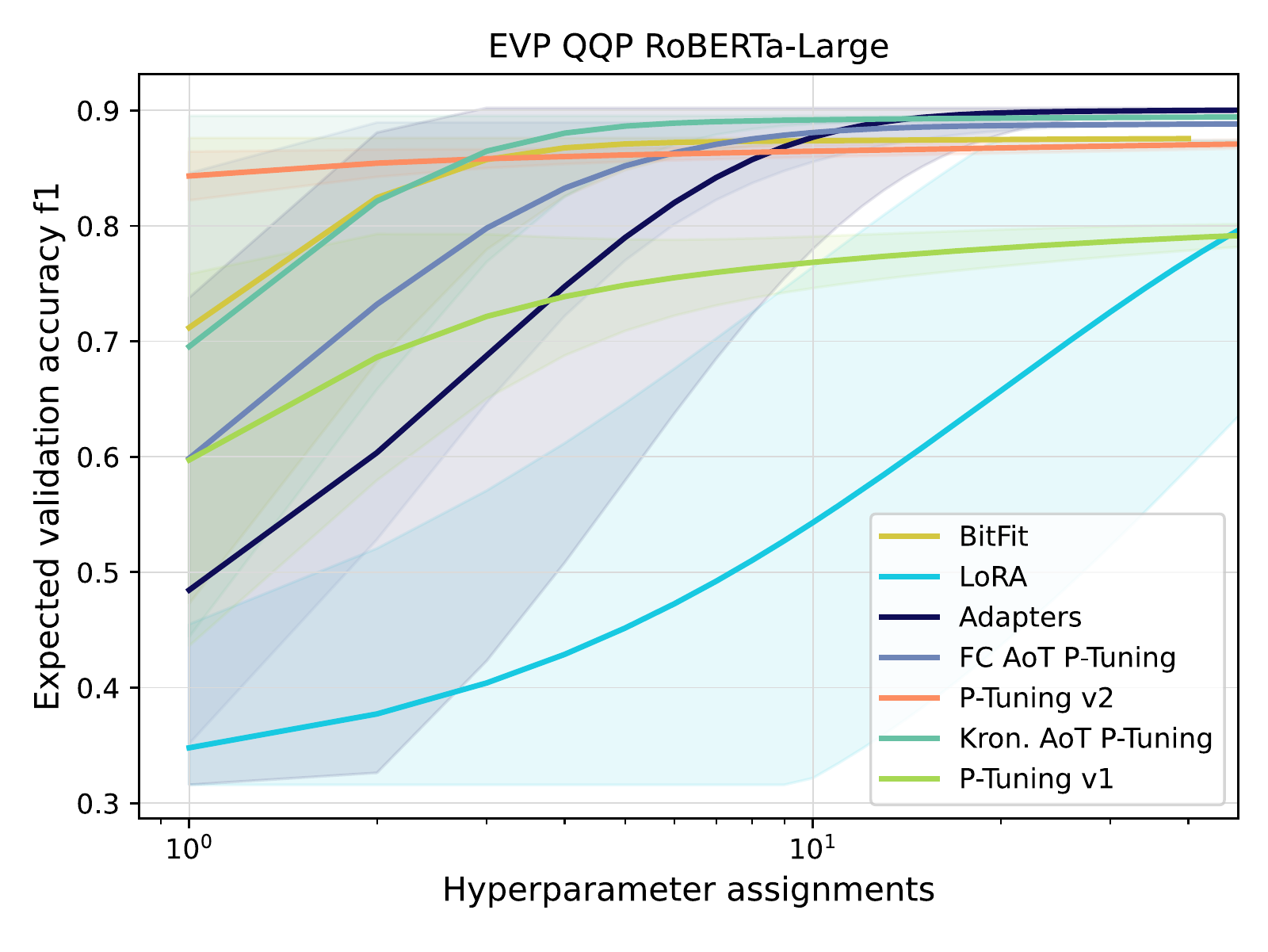}
    \caption{}
  \end{subfigure}
    \begin{subfigure}[t]{.24\linewidth}
    \centering\includegraphics[width=\linewidth]{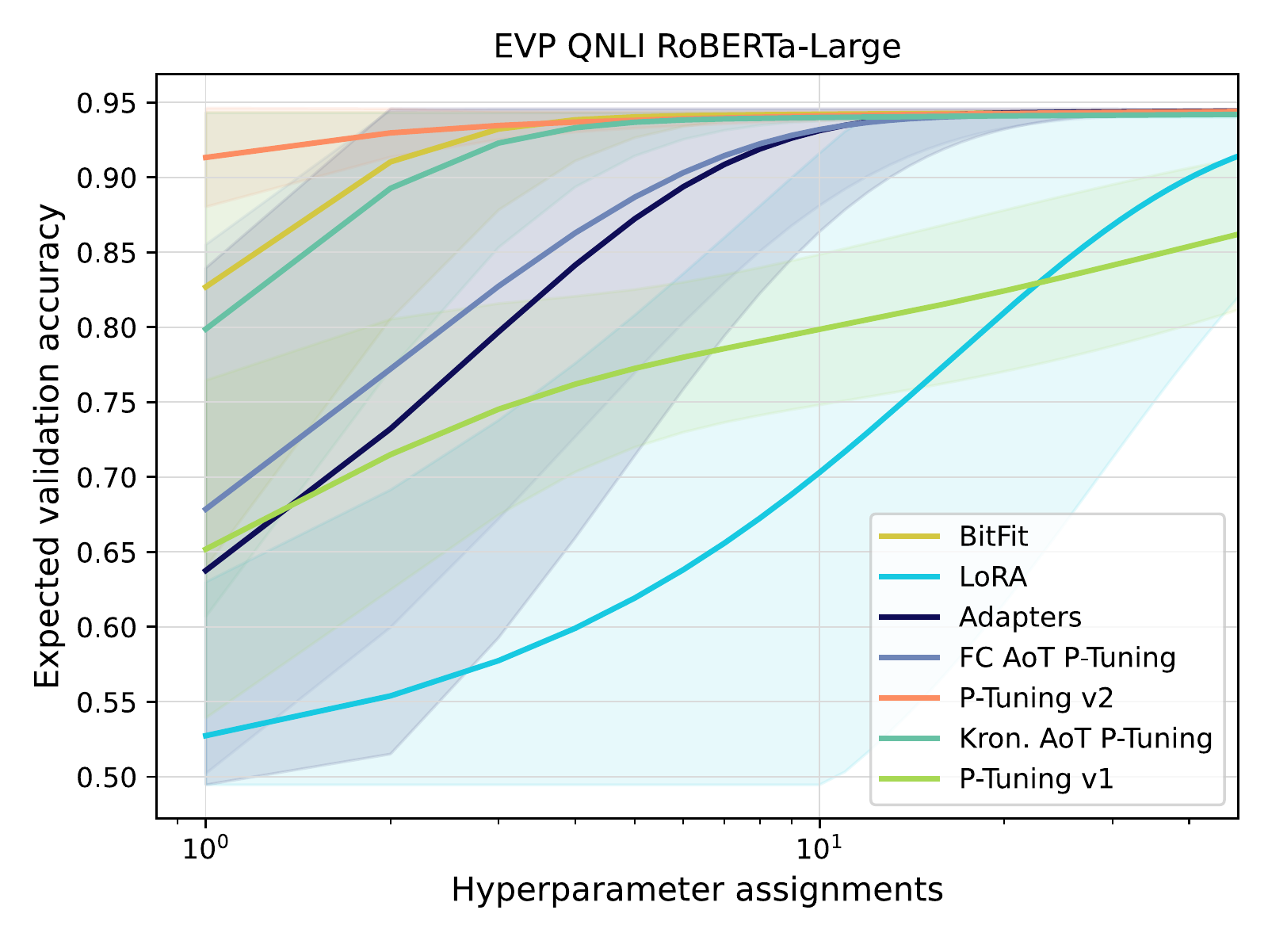}
    \caption{}
  \end{subfigure}
      \begin{subfigure}[t]{.24\linewidth}
    \centering\includegraphics[width=\linewidth]{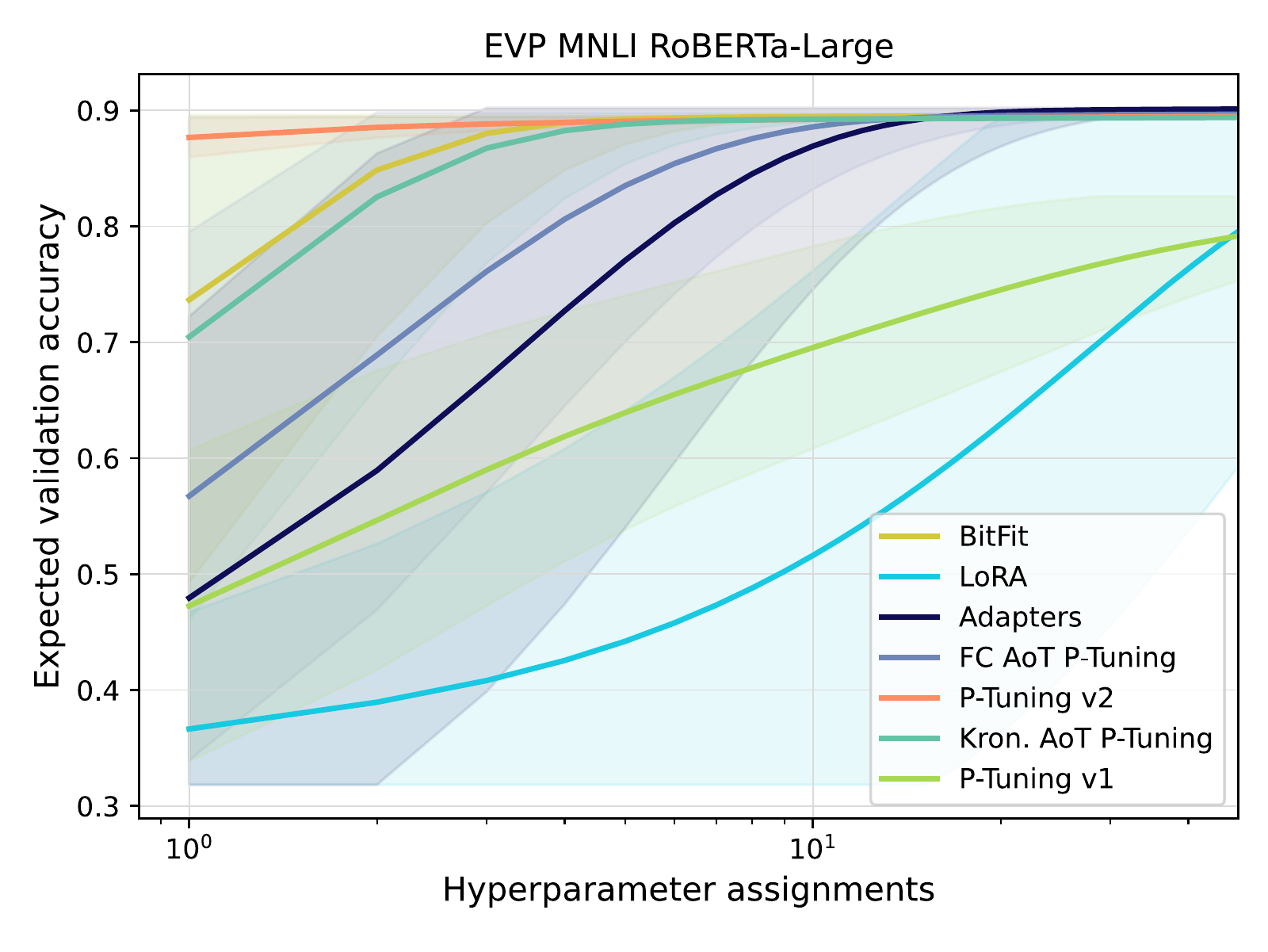}
    \caption{}
  \end{subfigure}
      \begin{subfigure}[t]{.24\linewidth}
    \centering\includegraphics[width=\linewidth]{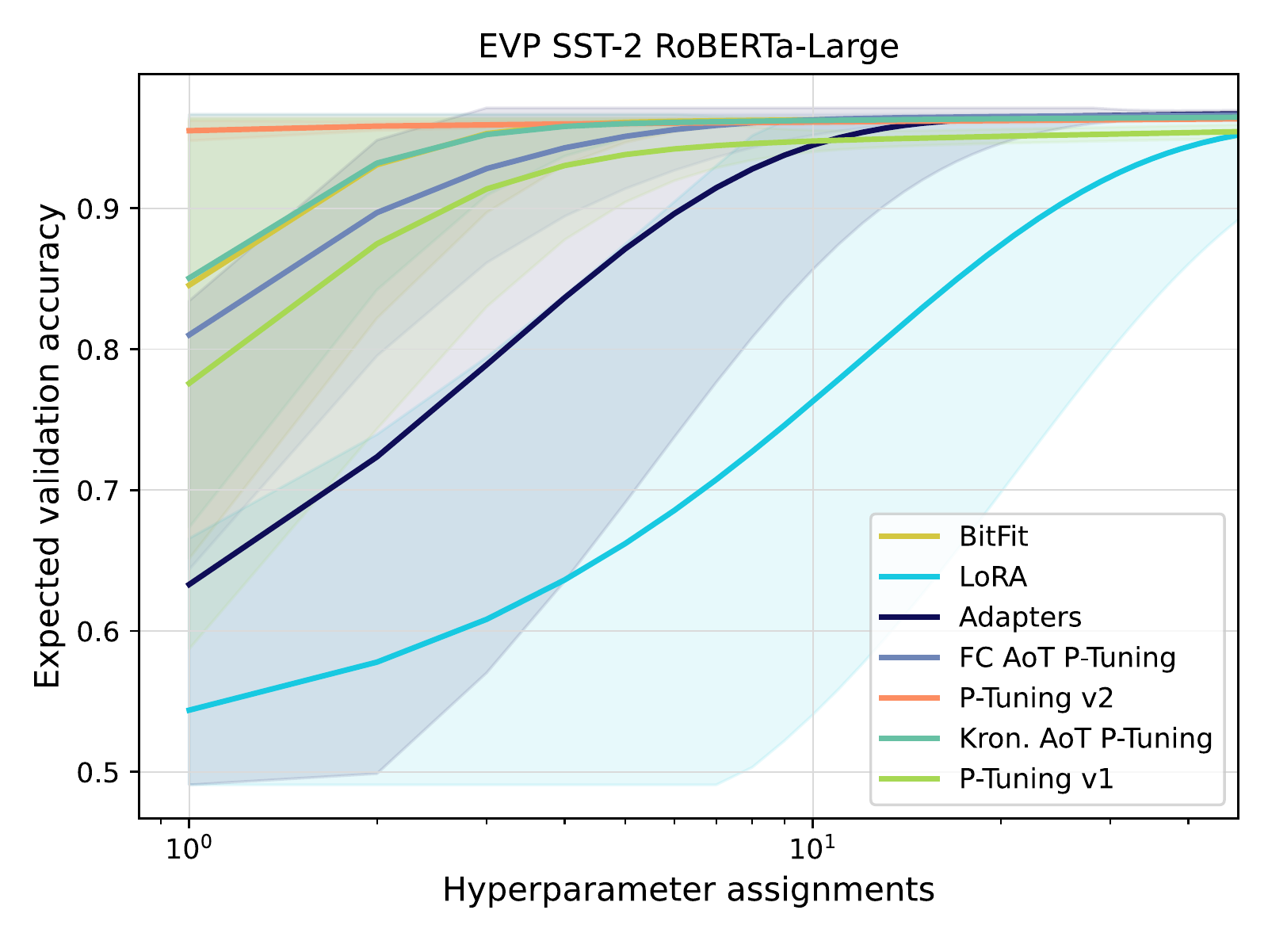}
    \caption{}
  \end{subfigure}
      \begin{subfigure}[t]{.24\linewidth}
    \centering\includegraphics[width=\linewidth]{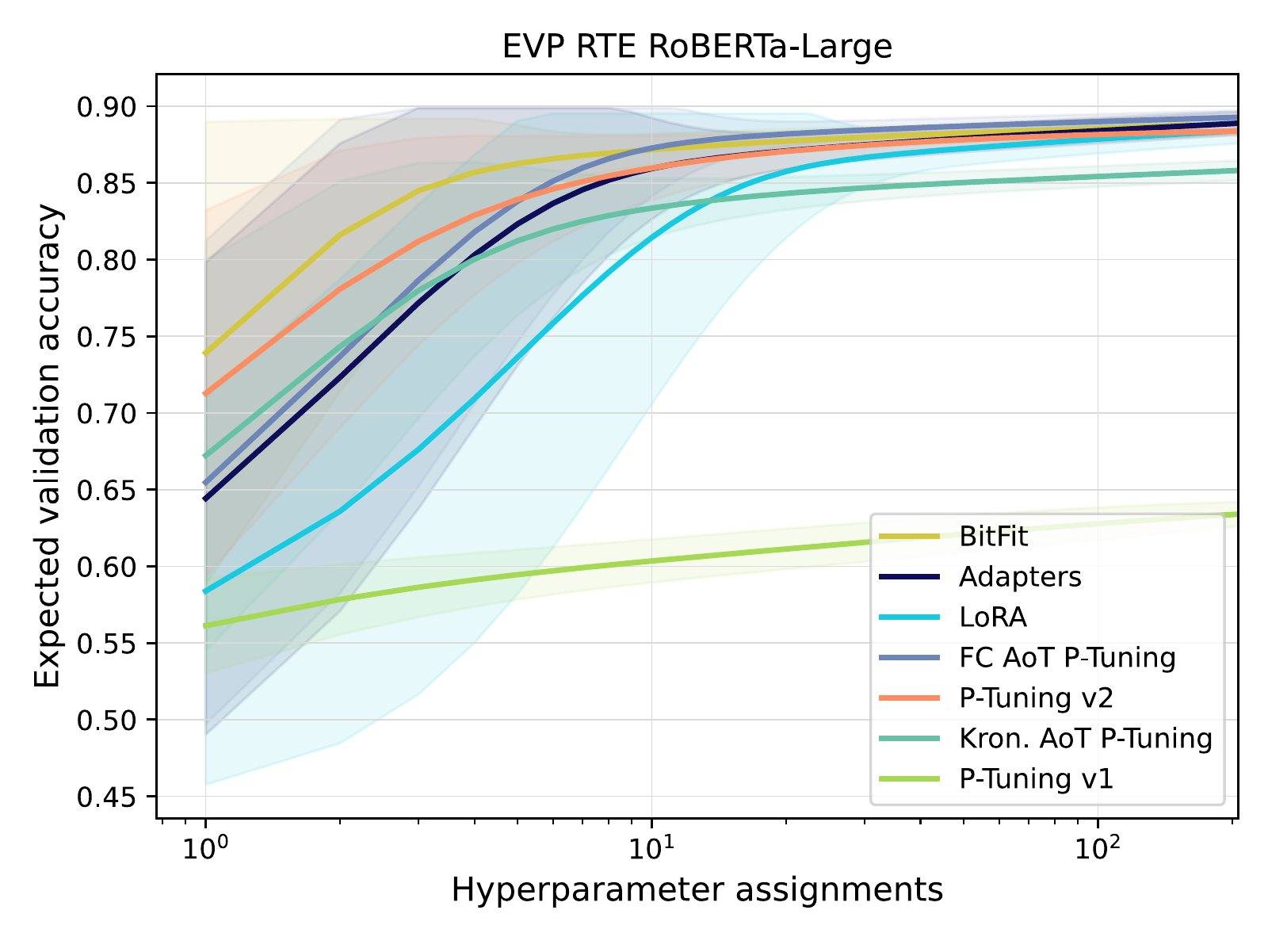}
    \caption{}
  \end{subfigure}
      \begin{subfigure}[t]{.24\linewidth}
    \centering\includegraphics[width=\linewidth]{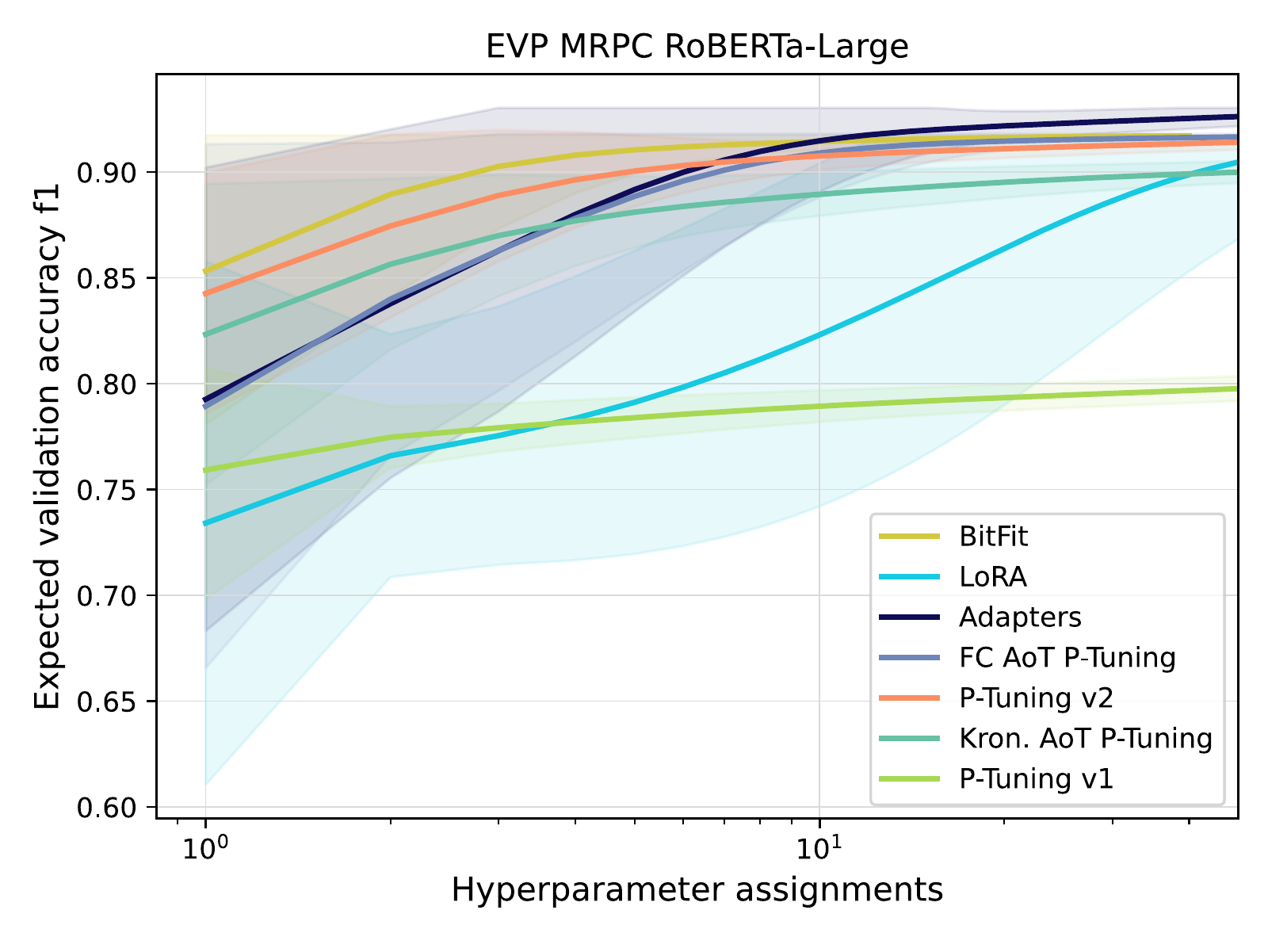}
    \caption{}
  \end{subfigure}
      \begin{subfigure}[t]{.24\linewidth}
    \centering\includegraphics[width=\linewidth]{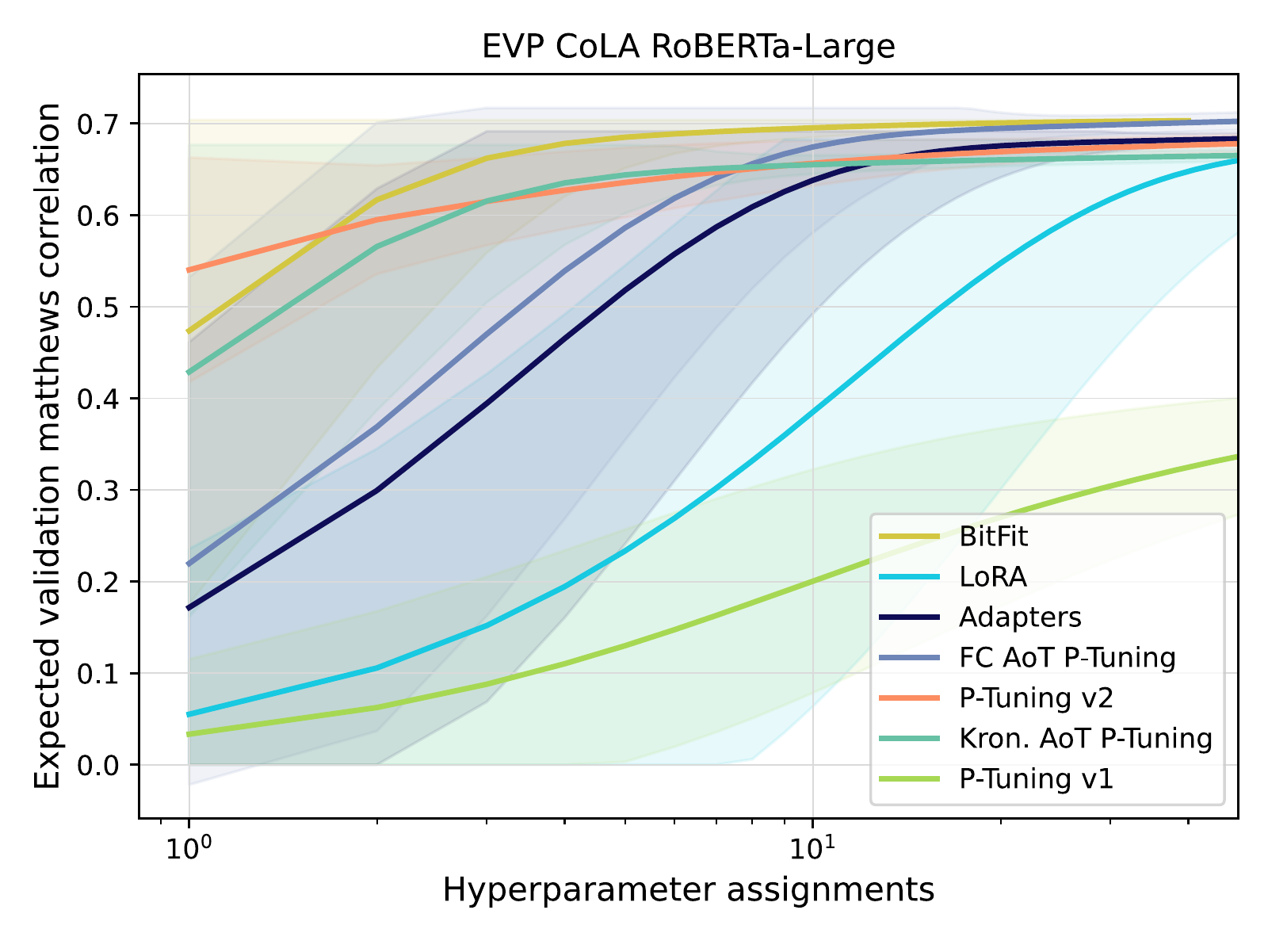}
    \caption{}
  \end{subfigure}
  \caption{Expected Validation Performance of trained models with GLUE Benchmarking Datasets for RoBERTa-Base (a-h) and RoBERTa-Large (i-p). See Section 4.2 for more details.}
  \label{figure-evp-glue}
\end{figure*}

\begin{figure*}[h!]
  \centering

   \medskip
    \begin{subfigure}[t]{.24\linewidth}
    \centering\includegraphics[width=\linewidth]{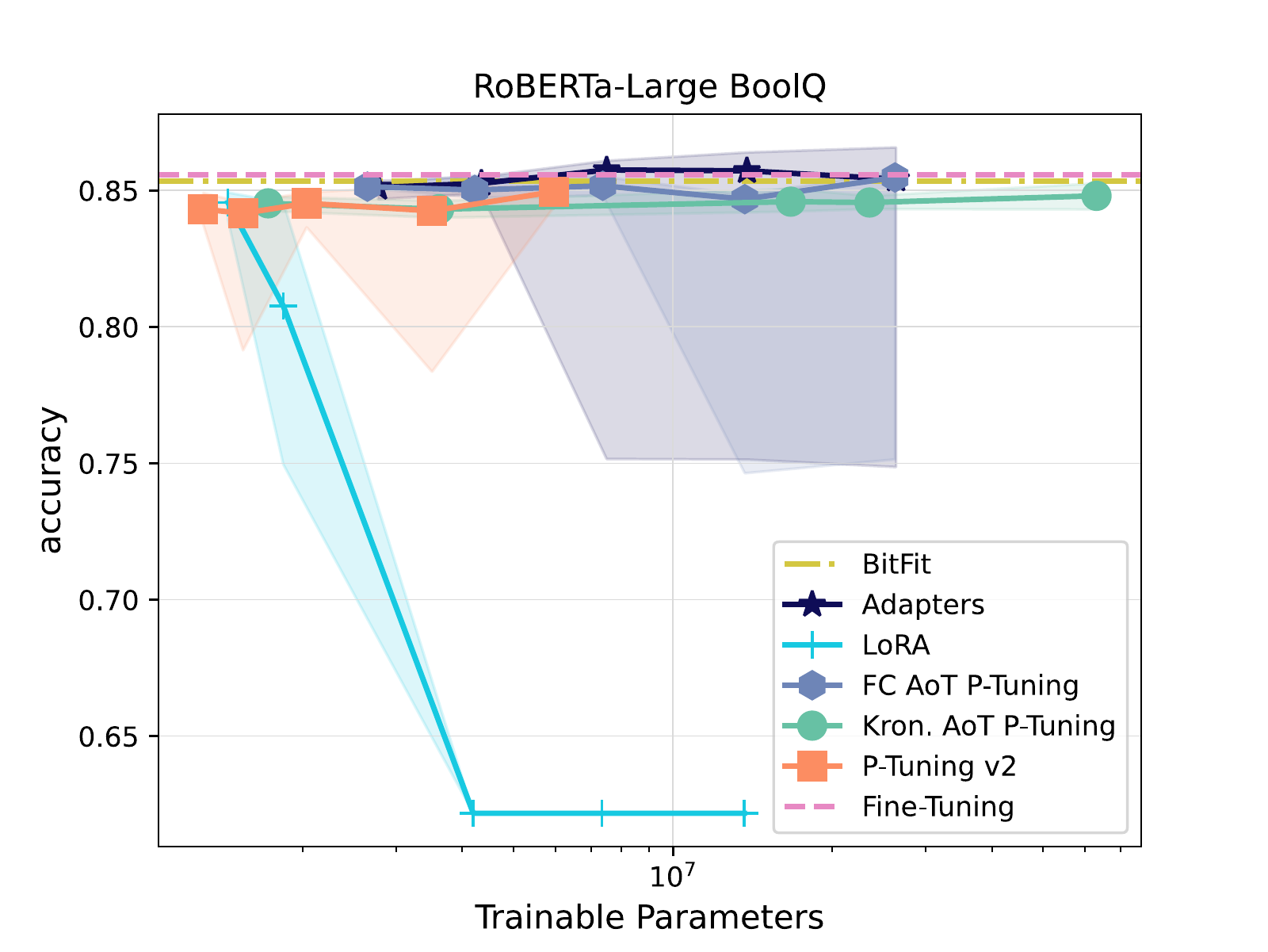}
    \caption{}
  \end{subfigure}
    \begin{subfigure}[t]{.24\linewidth}
    \centering\includegraphics[width=\linewidth]{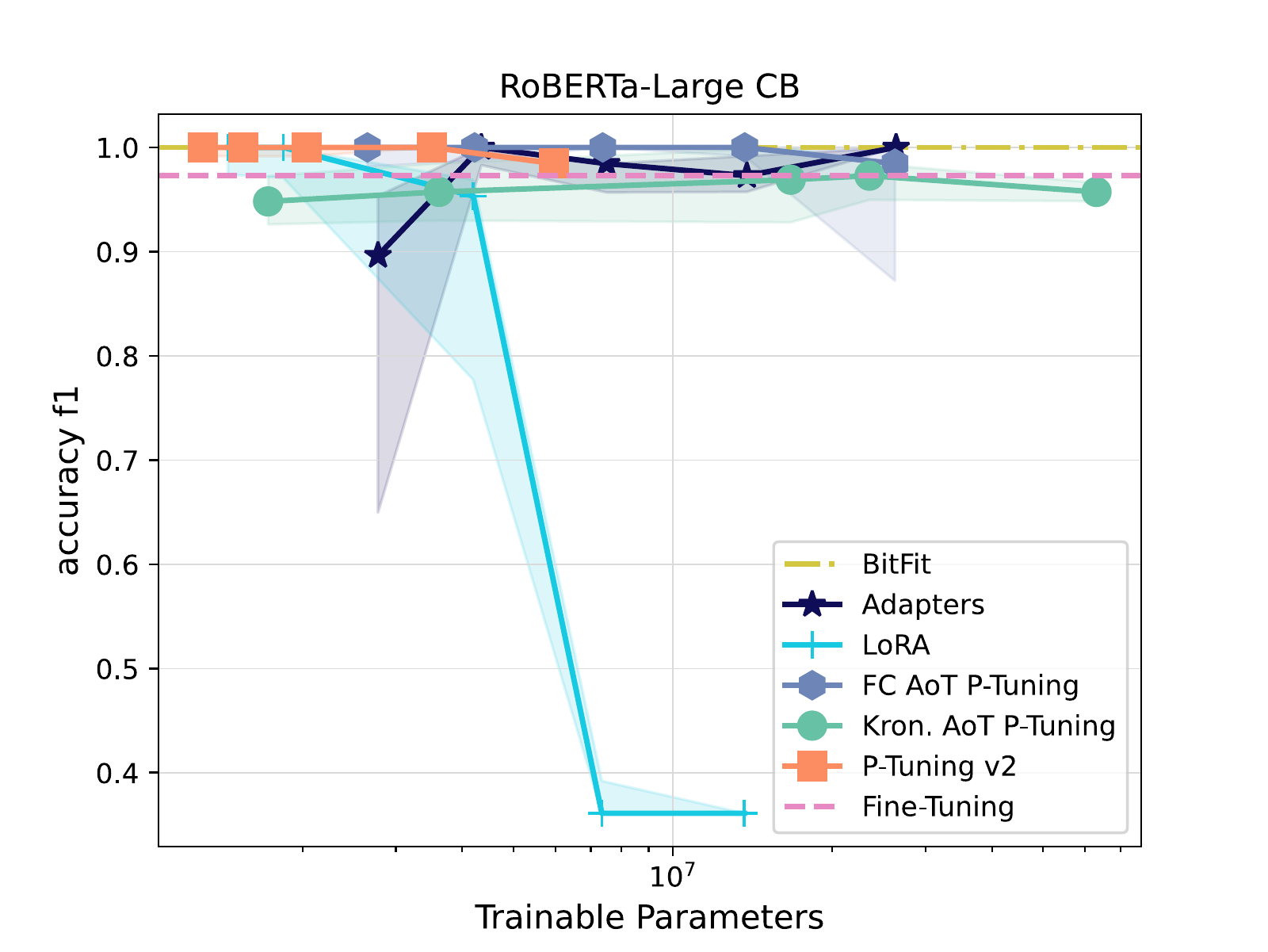}
    \caption{}
  \end{subfigure}
    \begin{subfigure}[t]{.24\linewidth}
    \centering\includegraphics[width=\linewidth]{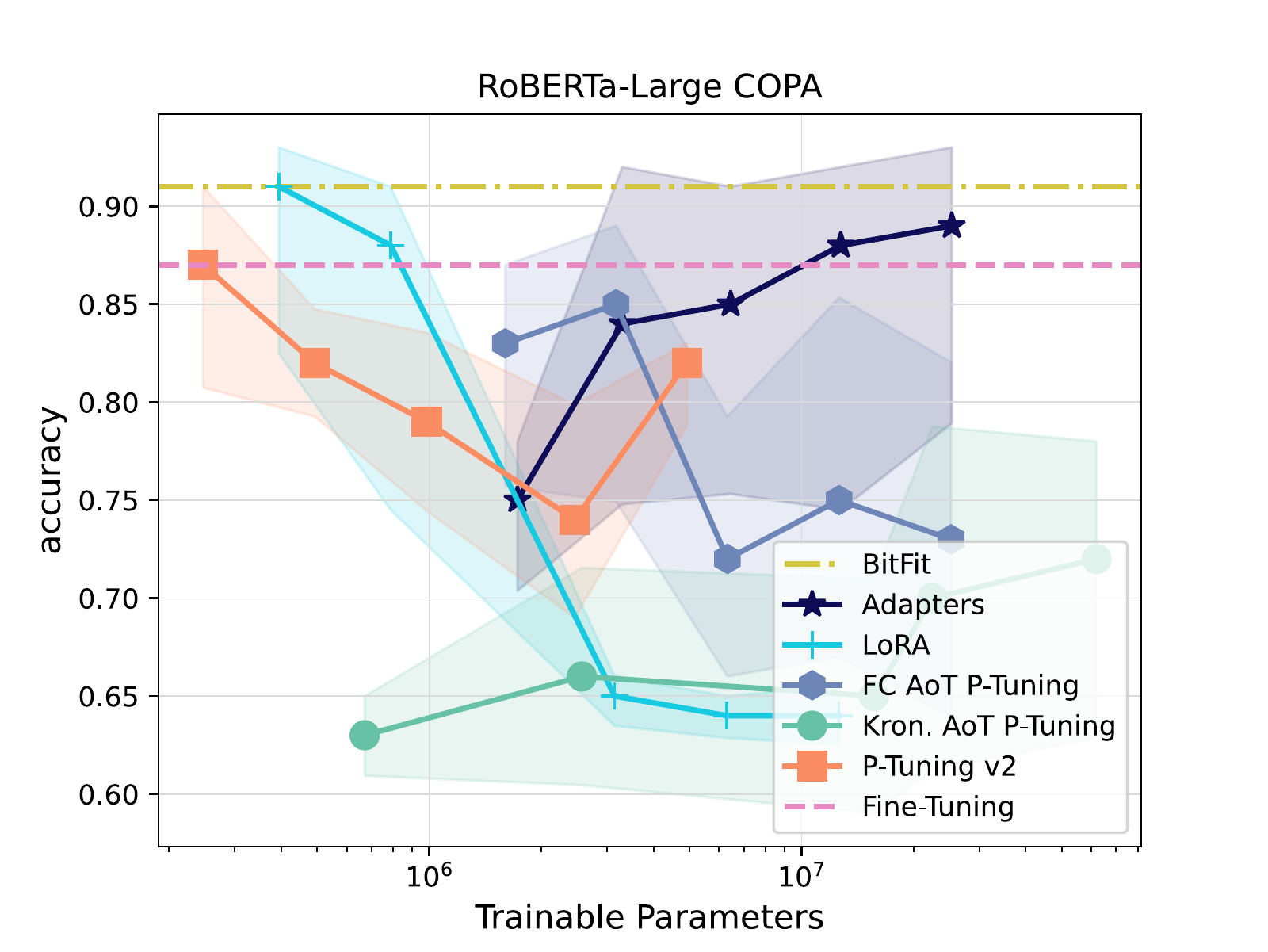}
    \caption{}
  \end{subfigure}
      \begin{subfigure}[t]{.24\linewidth}
    \centering\includegraphics[width=\linewidth]{images/roberta-large_rte_params.pdf}
    \caption{}
  \end{subfigure}
      \begin{subfigure}[t]{.26\linewidth}
    \centering\includegraphics[width=\linewidth]{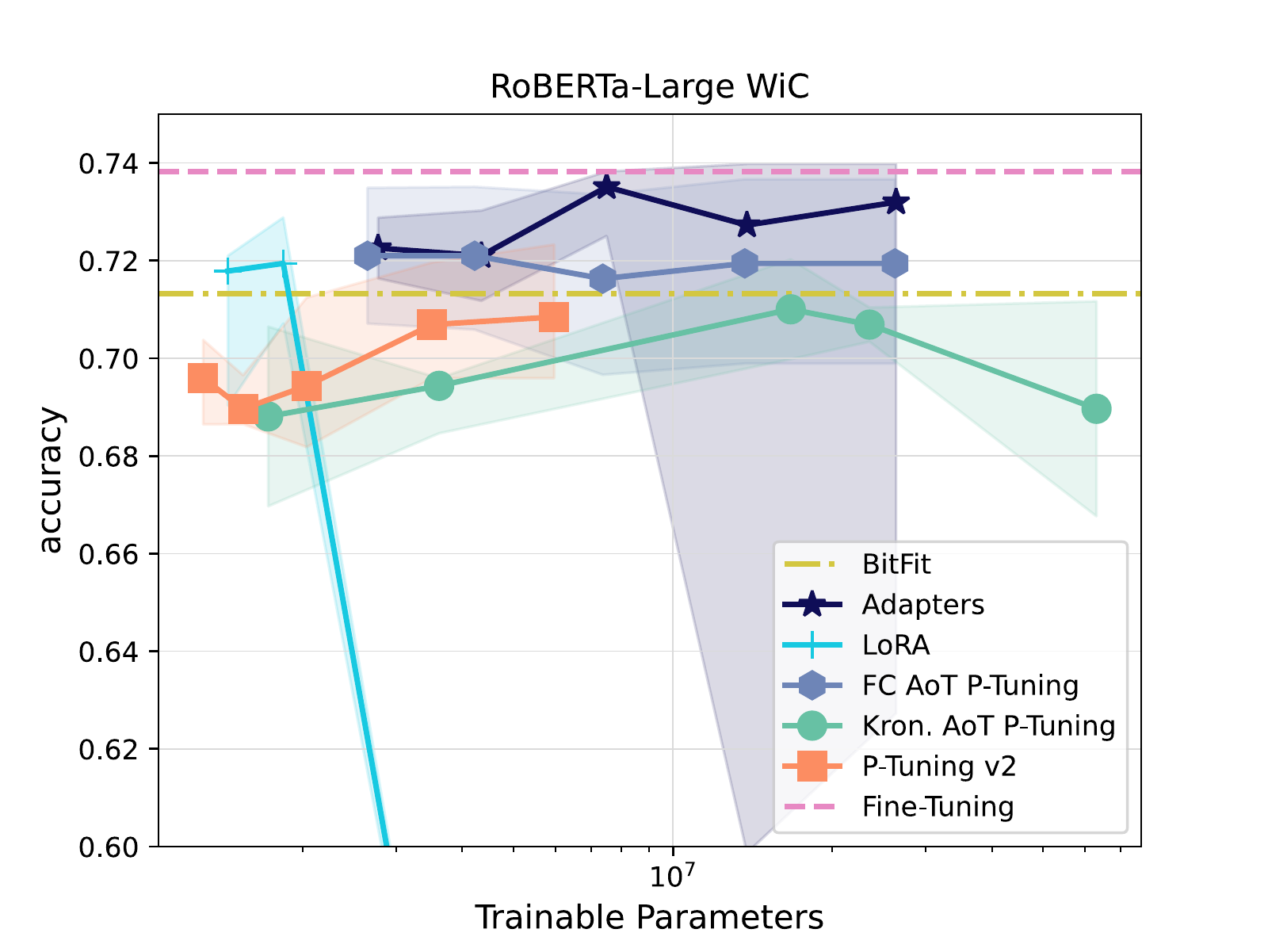}
    \caption{}
  \end{subfigure}
      \begin{subfigure}[t]{.26\linewidth}
    \centering\includegraphics[width=\linewidth]{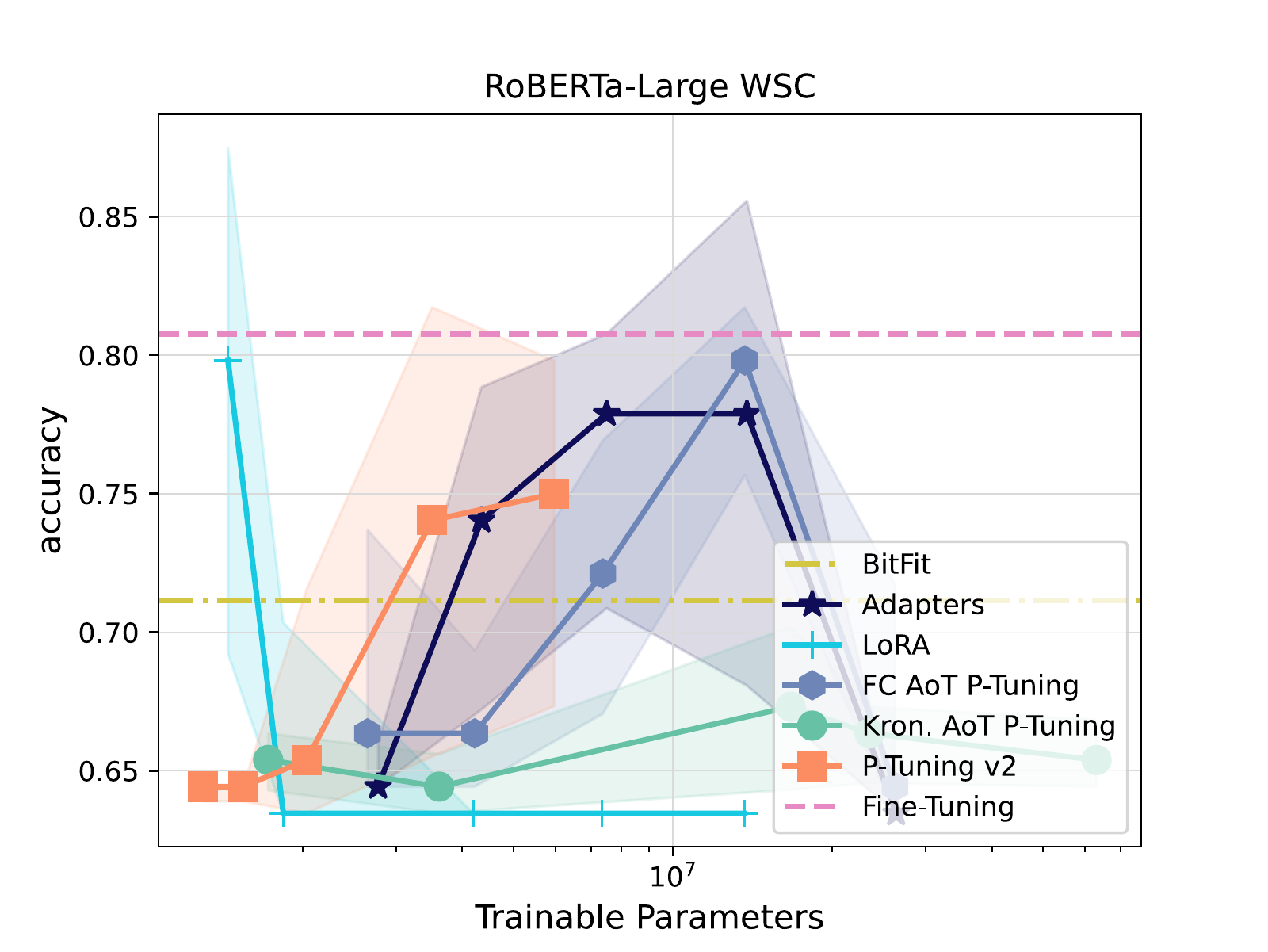}
    \caption{}
  \end{subfigure}
      \begin{subfigure}[t]{.26\linewidth}
    \centering\includegraphics[width=\linewidth]{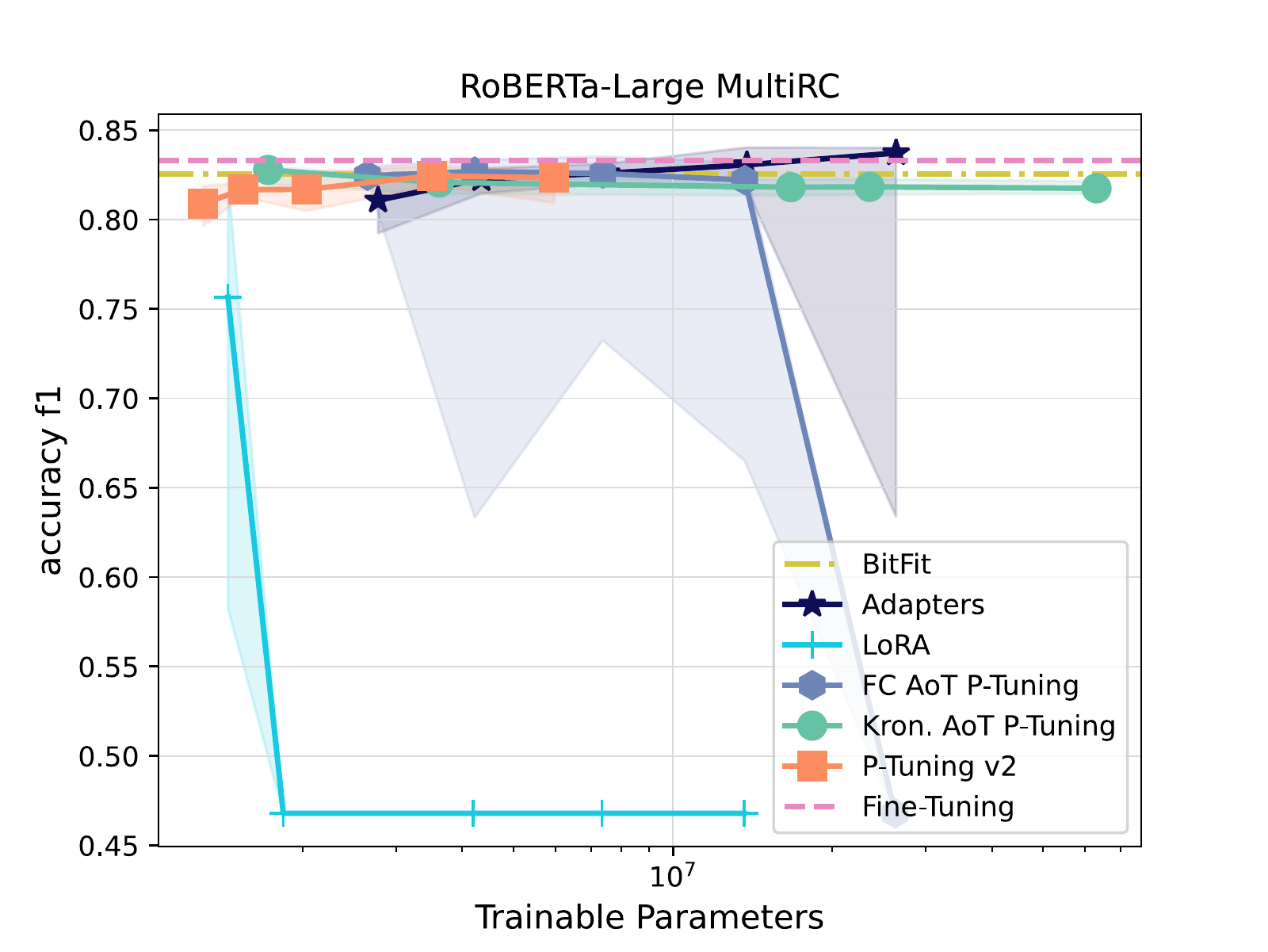}
    \caption{}
  \end{subfigure}
    \begin{subfigure}[t]{.24\linewidth}
    \centering\includegraphics[width=\linewidth]{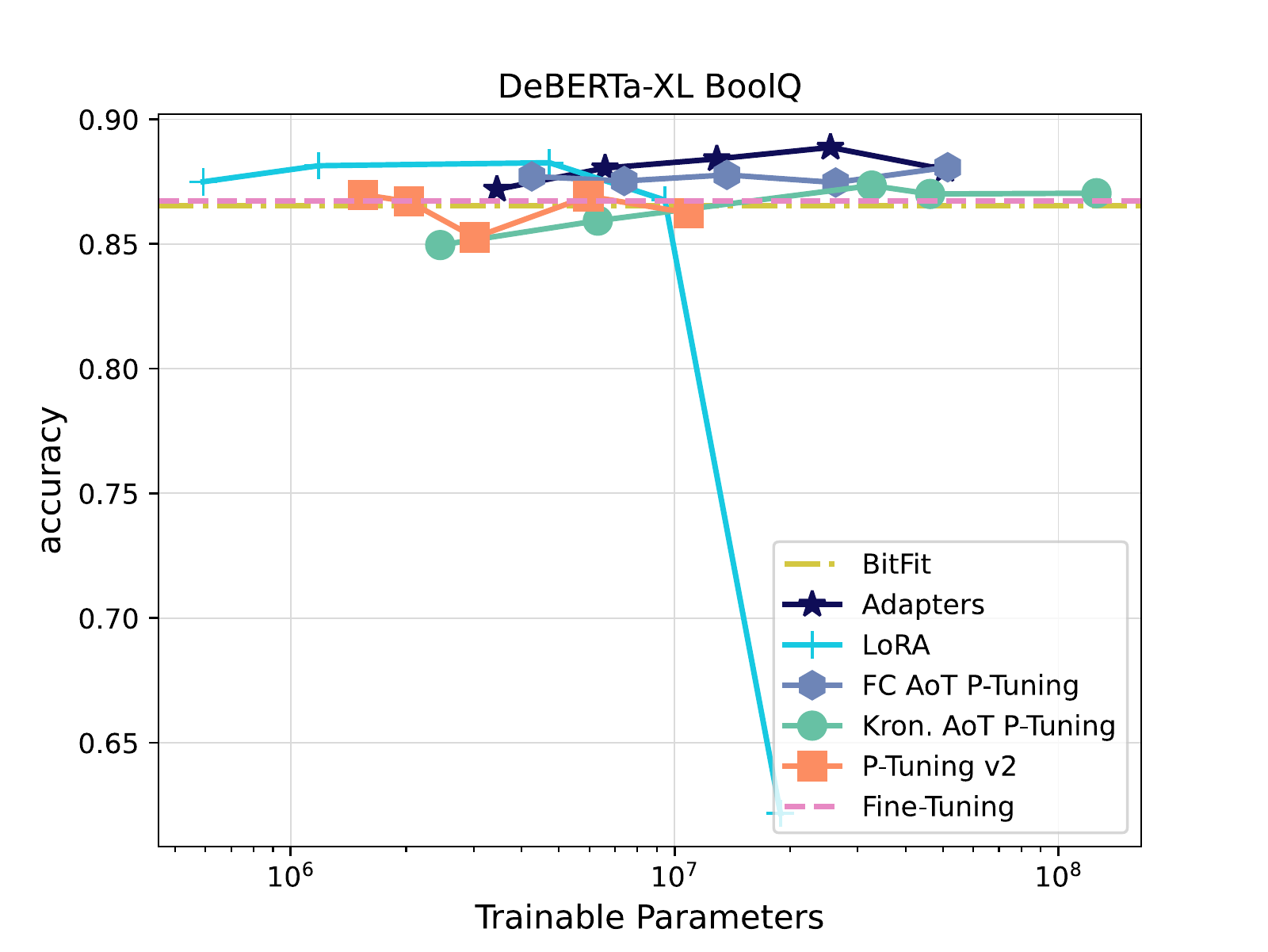}
    \caption{}
  \end{subfigure}
    \begin{subfigure}[t]{.24\linewidth}
    \centering\includegraphics[width=\linewidth]{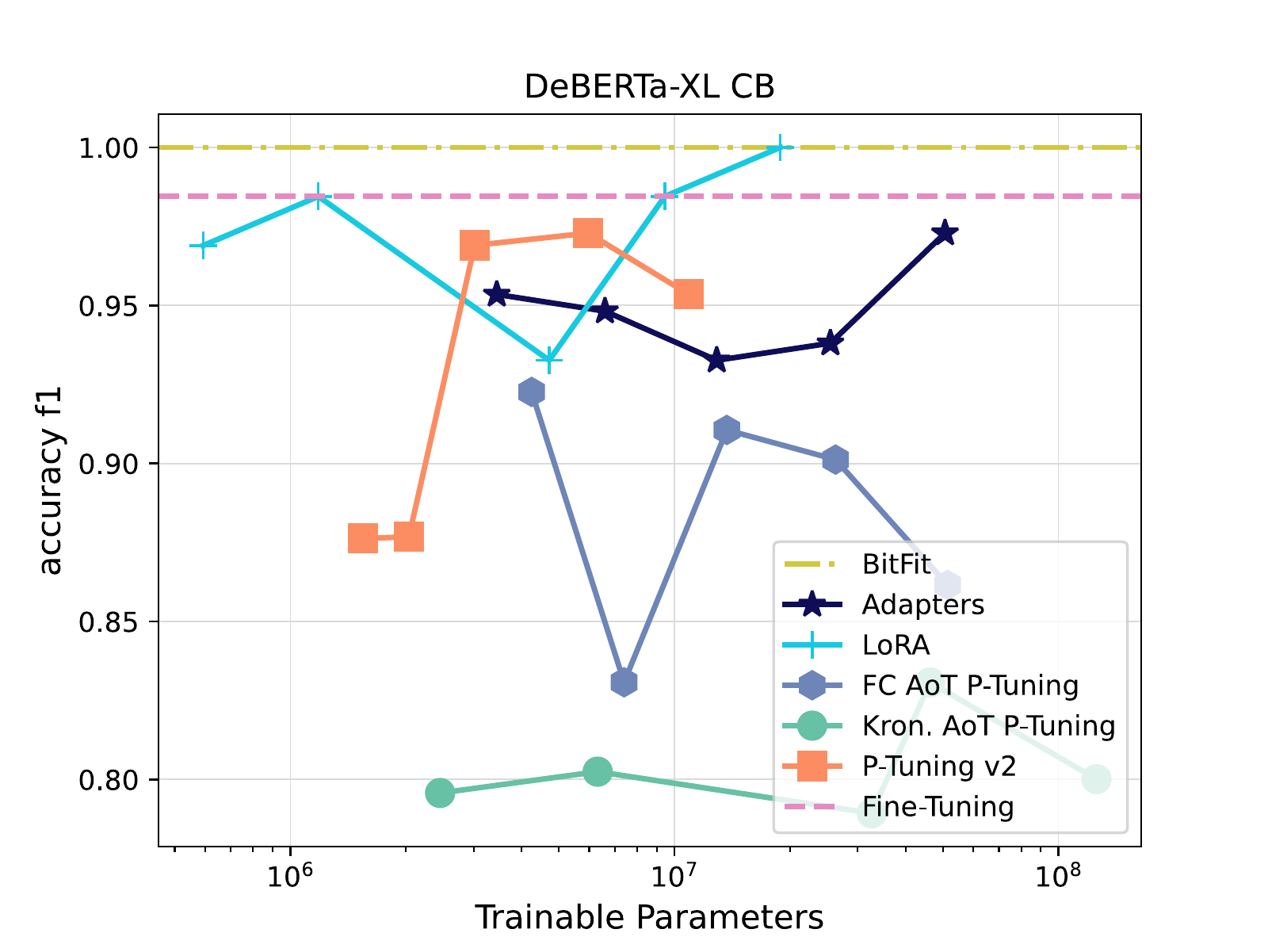}
    \caption{}
  \end{subfigure}
    \begin{subfigure}[t]{.24\linewidth}
    \centering\includegraphics[width=\linewidth]{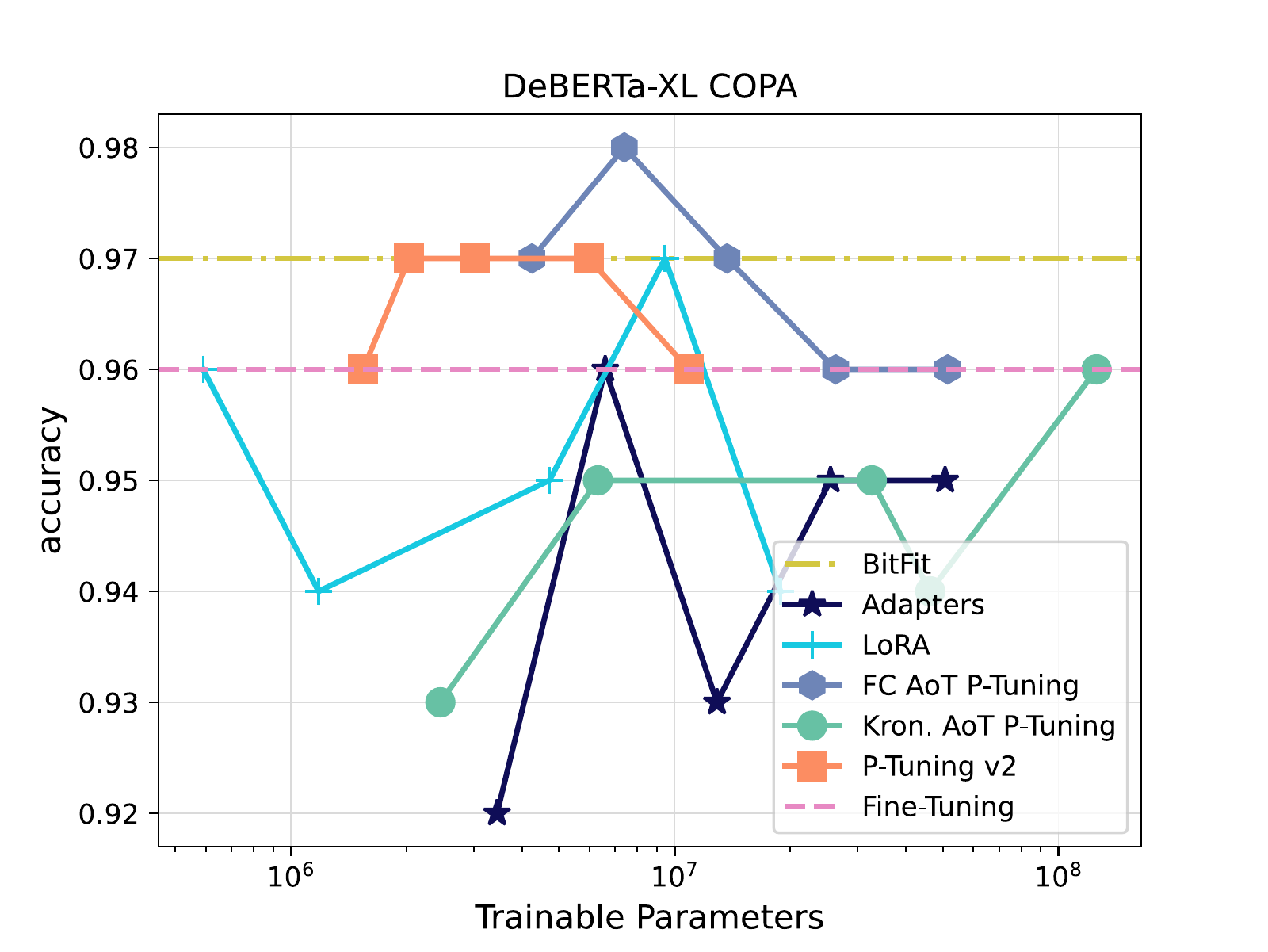}
    \caption{}
  \end{subfigure}
      \begin{subfigure}[t]{.24\linewidth}
    \centering\includegraphics[width=\linewidth]{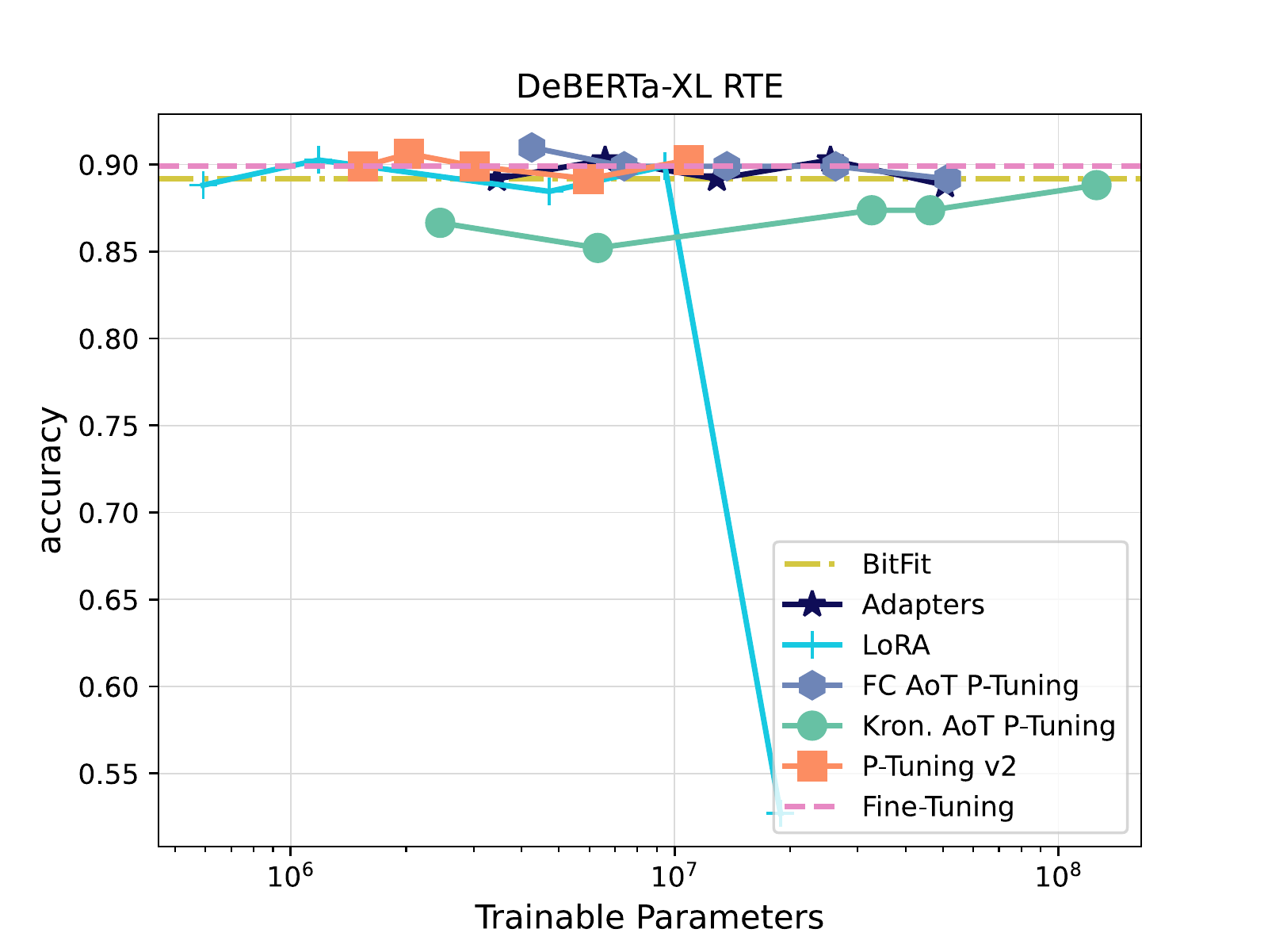}
    \caption{}
  \end{subfigure}
      \begin{subfigure}[t]{.26\linewidth}
    \centering\includegraphics[width=\linewidth]{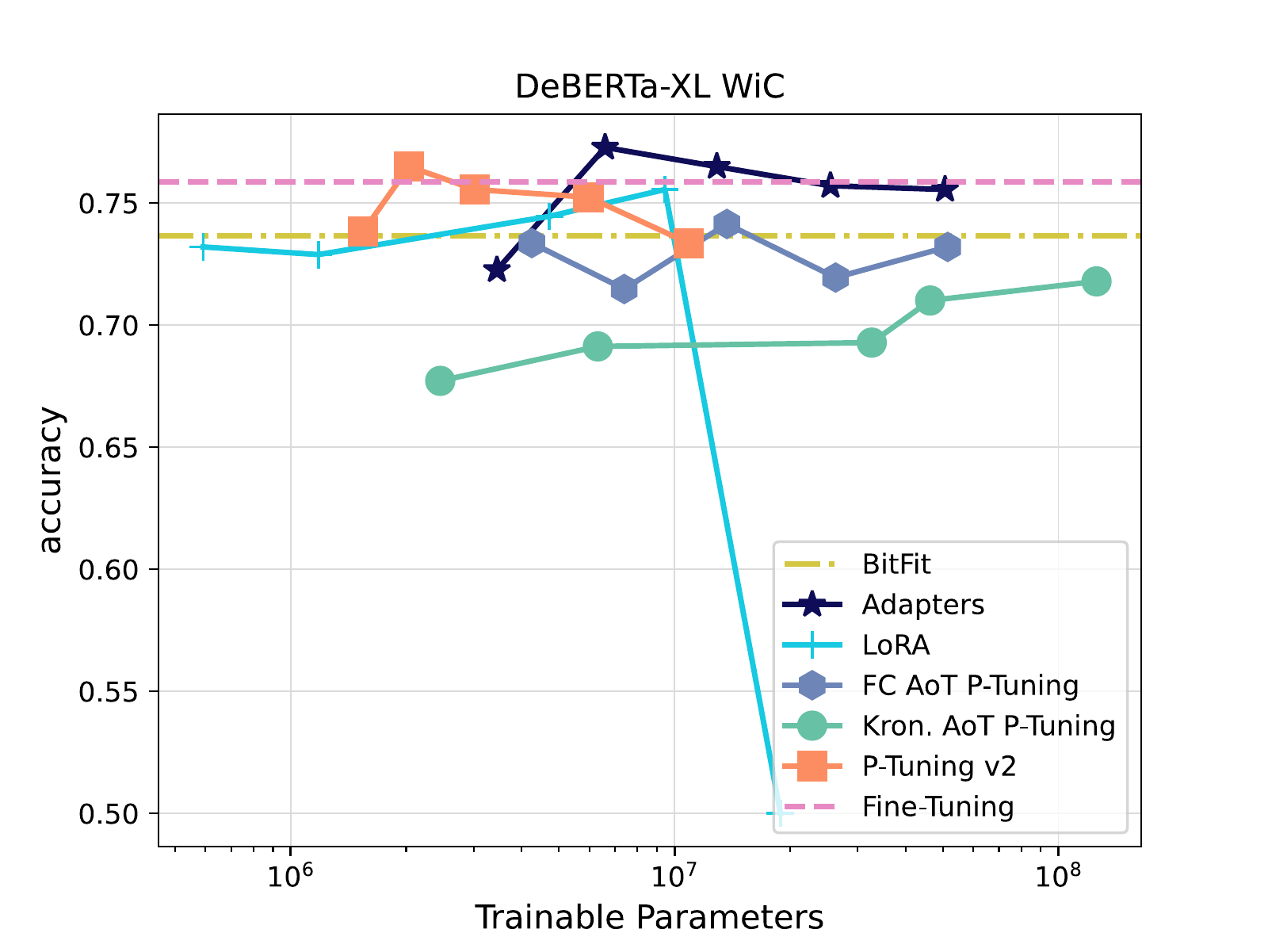}
    \caption{}
  \end{subfigure}
      \begin{subfigure}[t]{.26\linewidth}
    \centering\includegraphics[width=\linewidth]{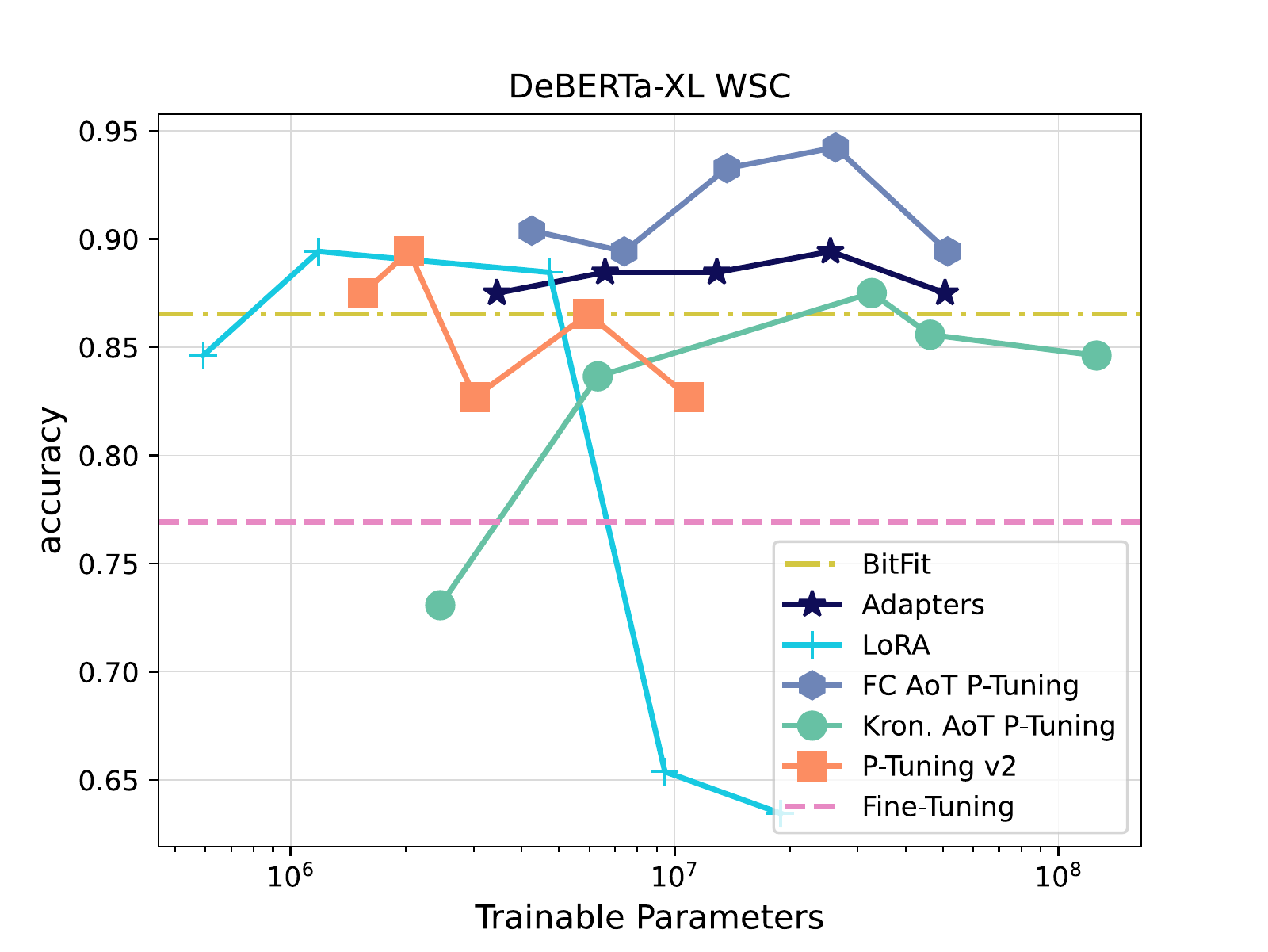}
    \caption{}
  \end{subfigure}
      \begin{subfigure}[t]{.26\linewidth}
    \centering\includegraphics[width=\linewidth]{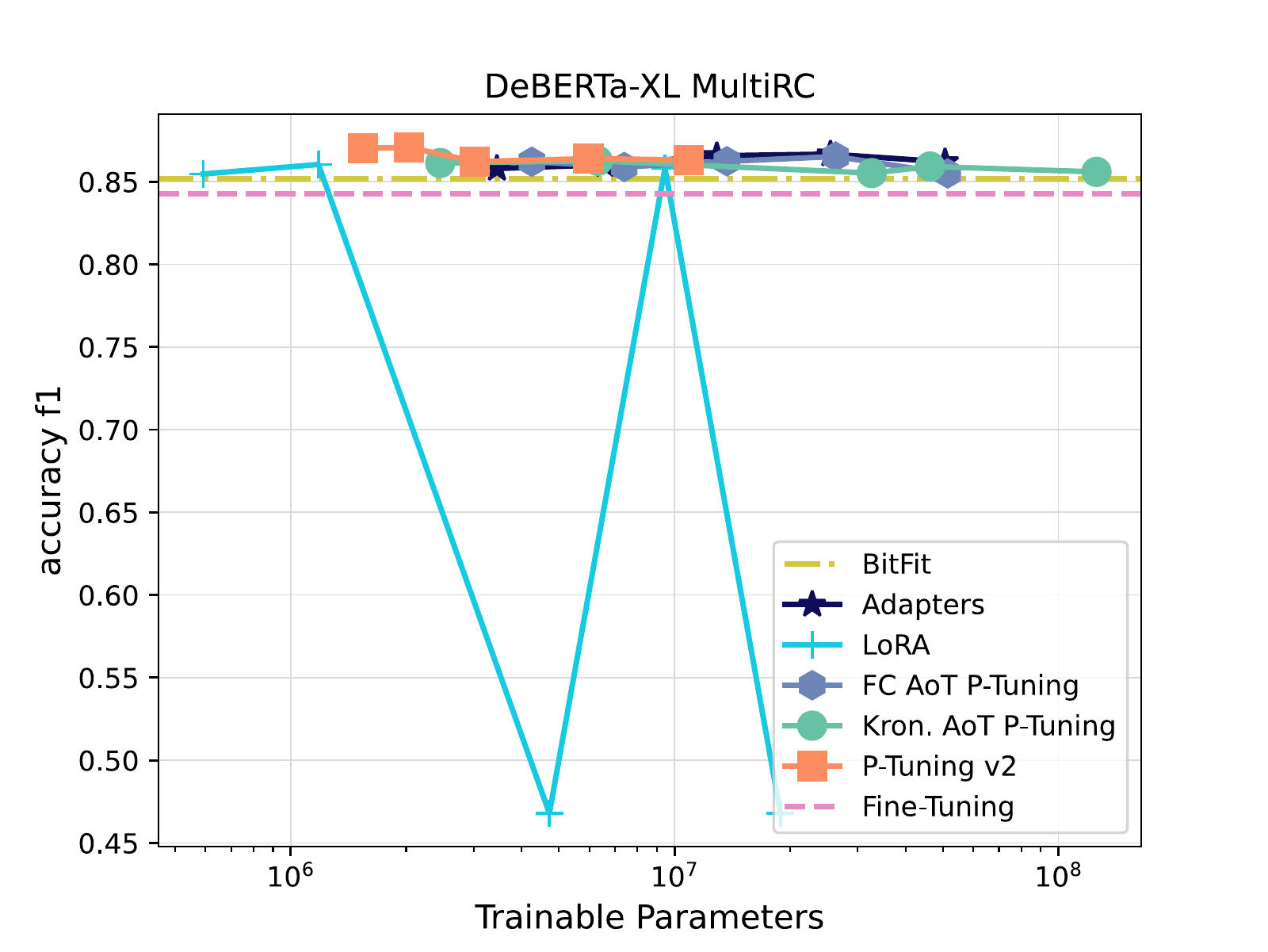}
    \caption{}
  \end{subfigure}
  \caption{Per-task SuperGLUE Benchmarking Dataset results for a different number of trained parameters of P-Tuning v2 and AoT P-Tuning with RoBERTa-Large (a-g) and RoBERTa-Large (h-n). We also provide results of plain fine-tuning for reference. See Section 4.2 for more details.}
  \label{figure-parameters-superglue}
\end{figure*}

\begin{figure*}[h!]
  \centering

  \medskip
    \begin{subfigure}[t]{.24\linewidth}
    \centering\includegraphics[width=\linewidth]{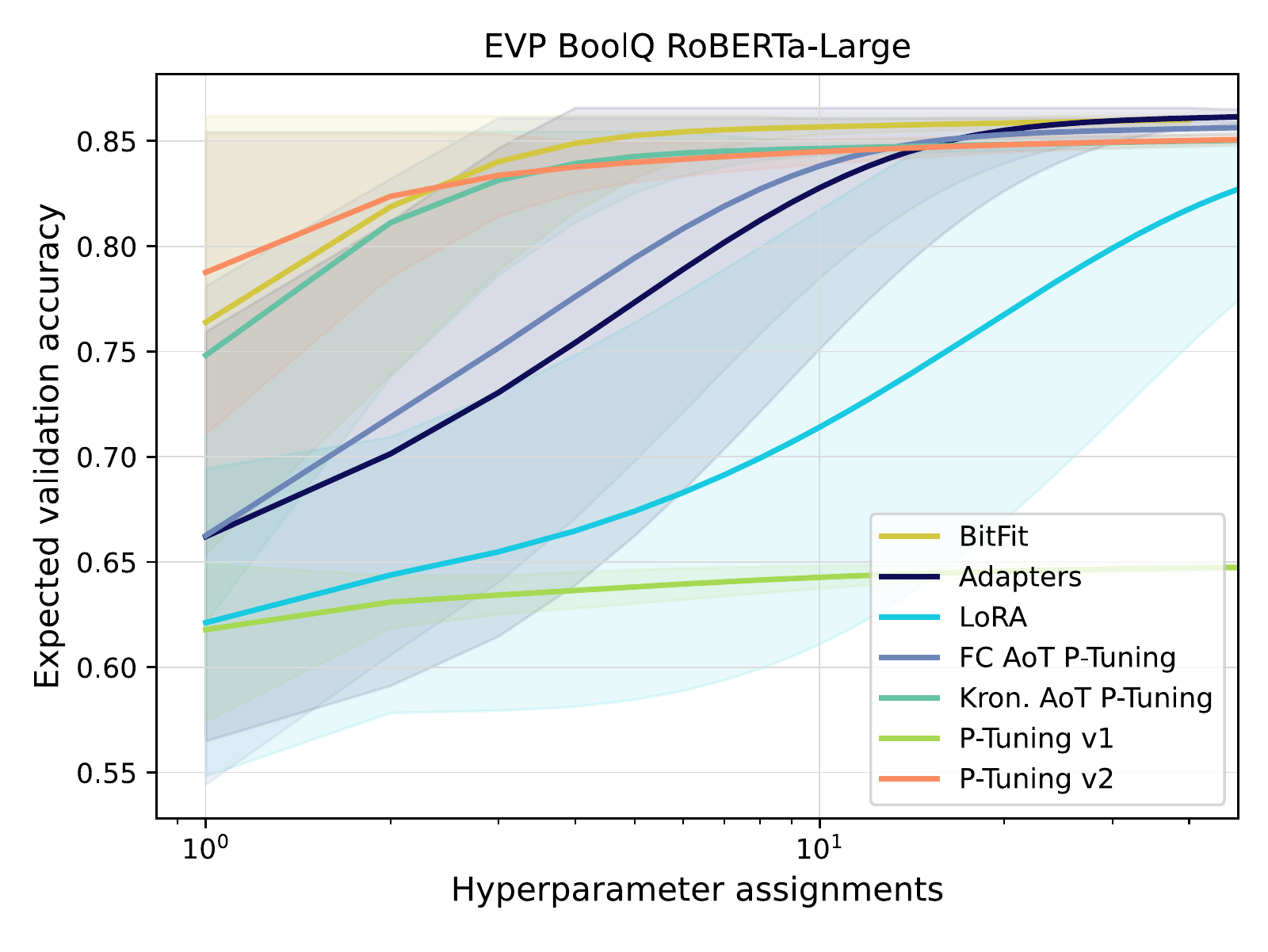}
    \caption{}
  \end{subfigure}
    \begin{subfigure}[t]{.24\linewidth}
    \centering\includegraphics[width=\linewidth]{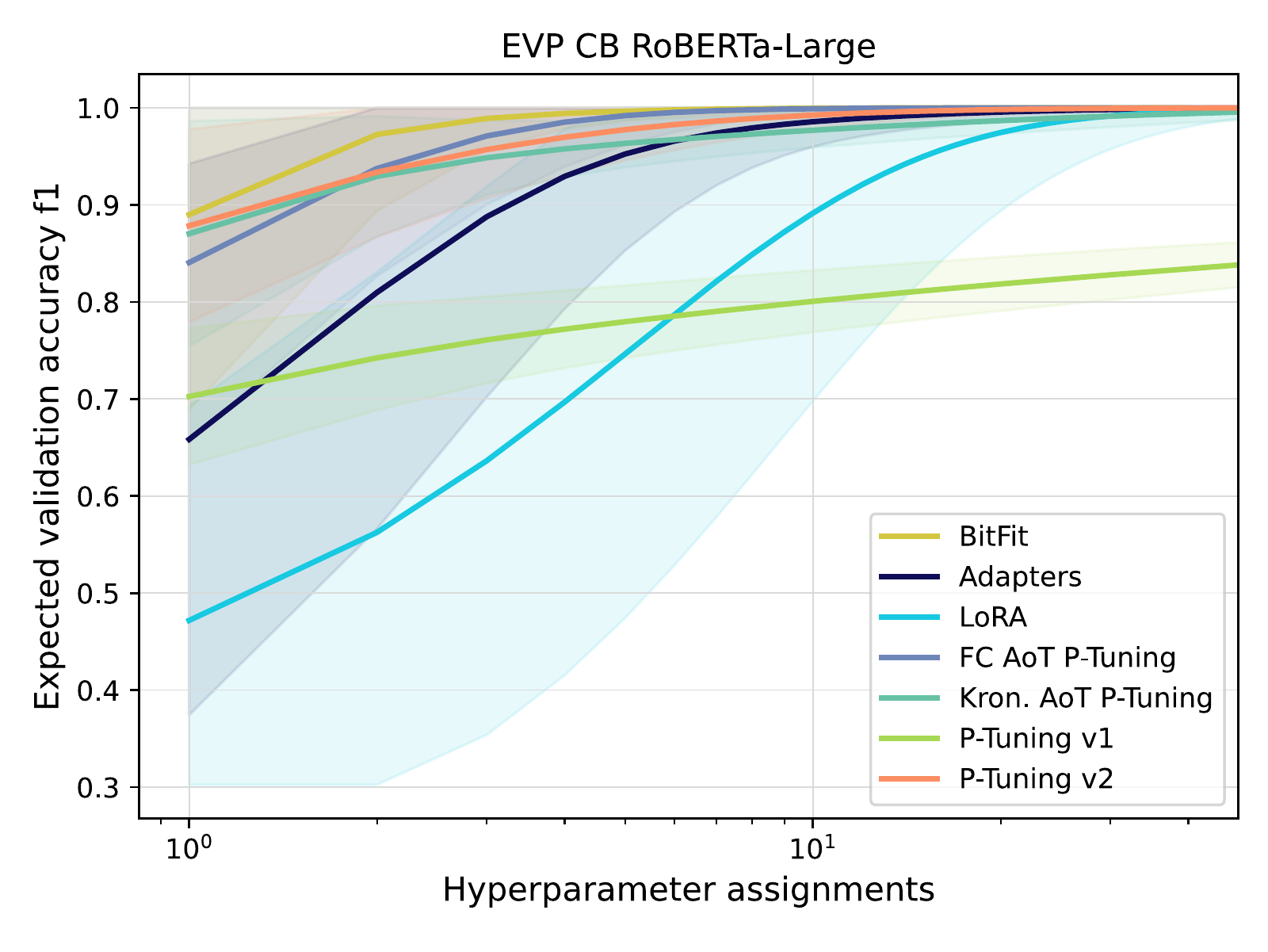}
    \caption{}
  \end{subfigure}
    \begin{subfigure}[t]{.24\linewidth}
    \centering\includegraphics[width=\linewidth]{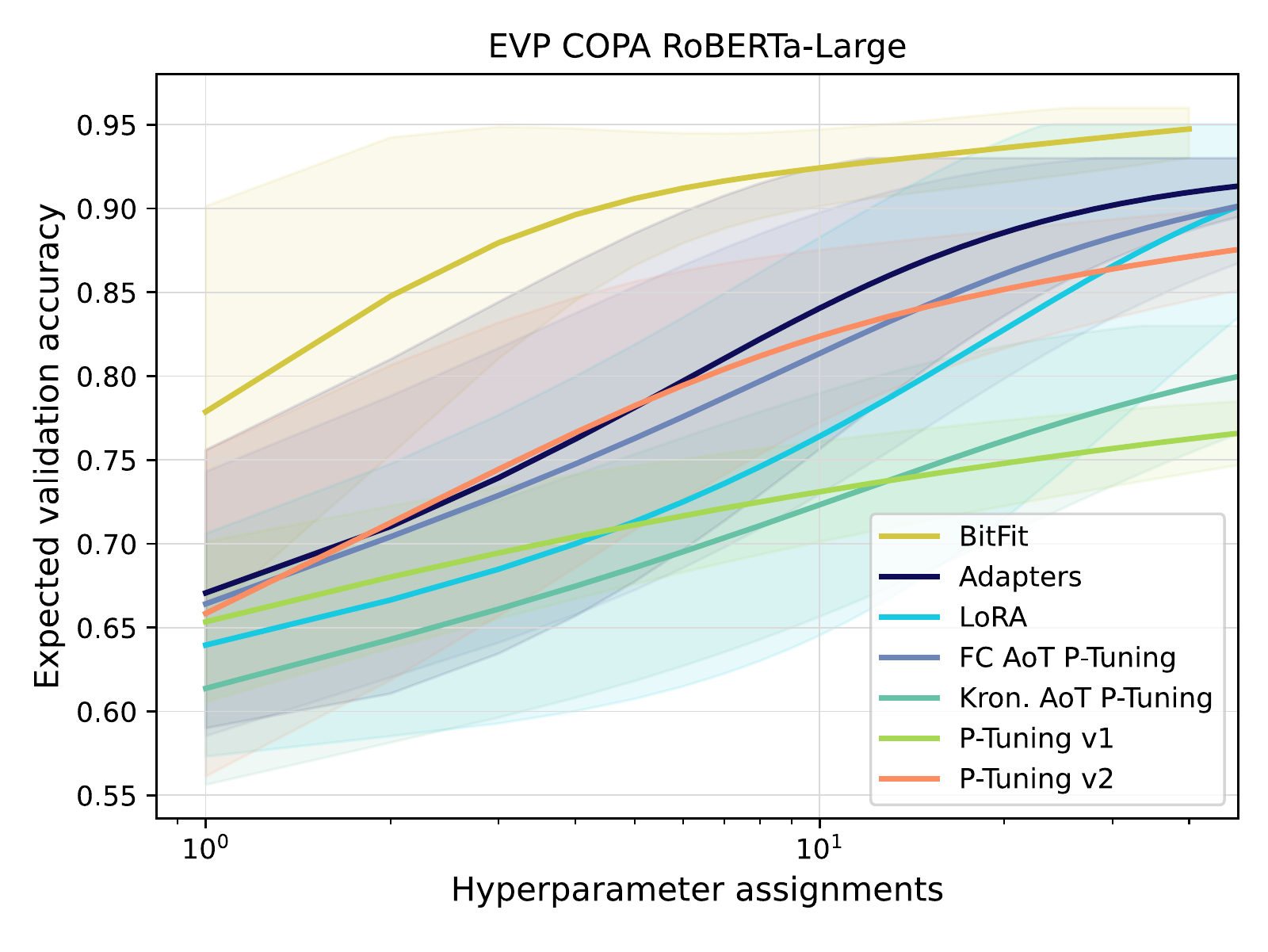}
    \caption{}
  \end{subfigure}
      \begin{subfigure}[t]{.24\linewidth}
    \centering\includegraphics[width=\linewidth]{images/roberta-large_rte_evp.pdf}
    \caption{}
  \end{subfigure}
      \begin{subfigure}[t]{.26\linewidth}
    \centering\includegraphics[width=\linewidth]{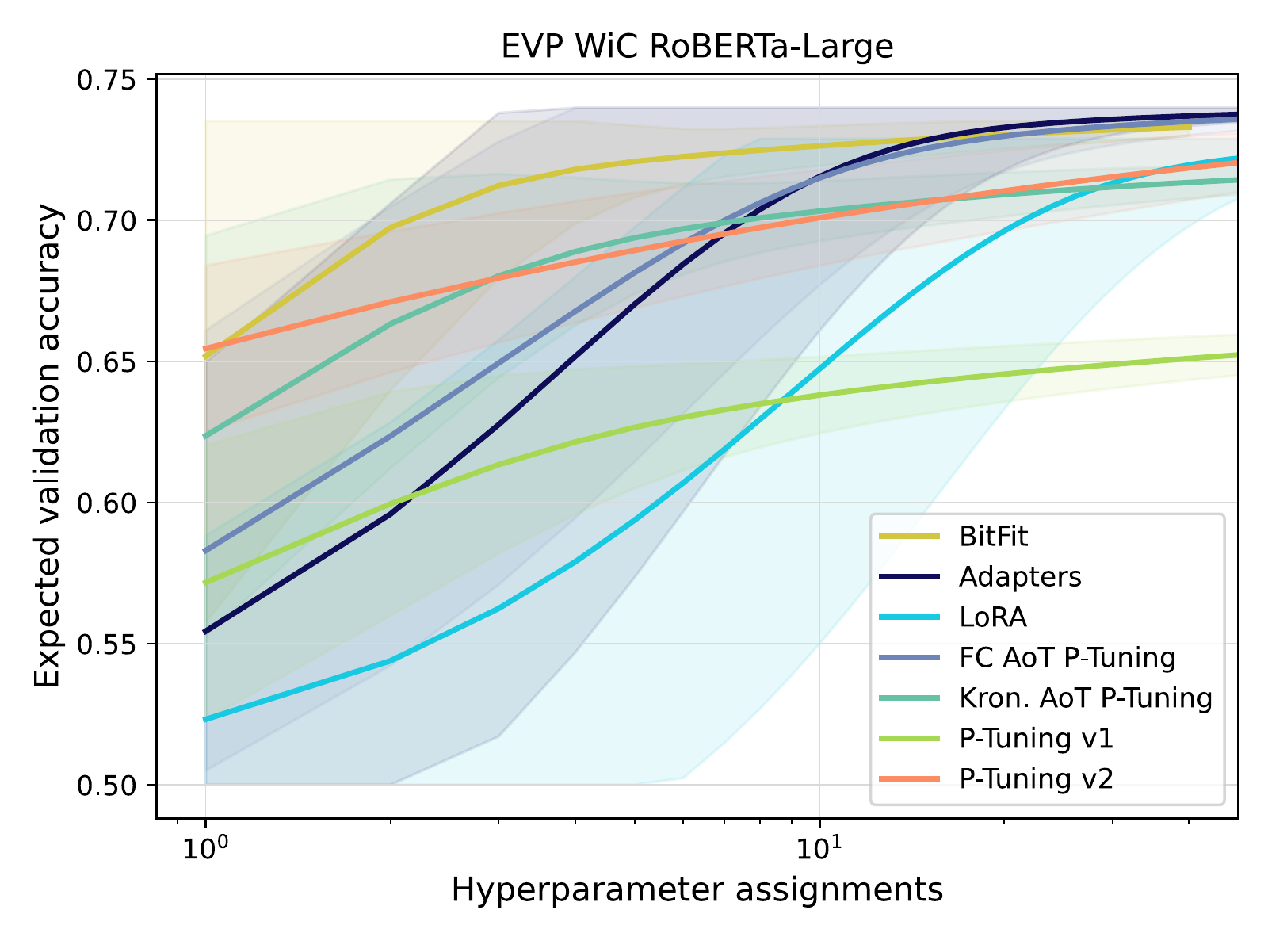}
    \caption{}
  \end{subfigure}
      \begin{subfigure}[t]{.26\linewidth}
    \centering\includegraphics[width=\linewidth]{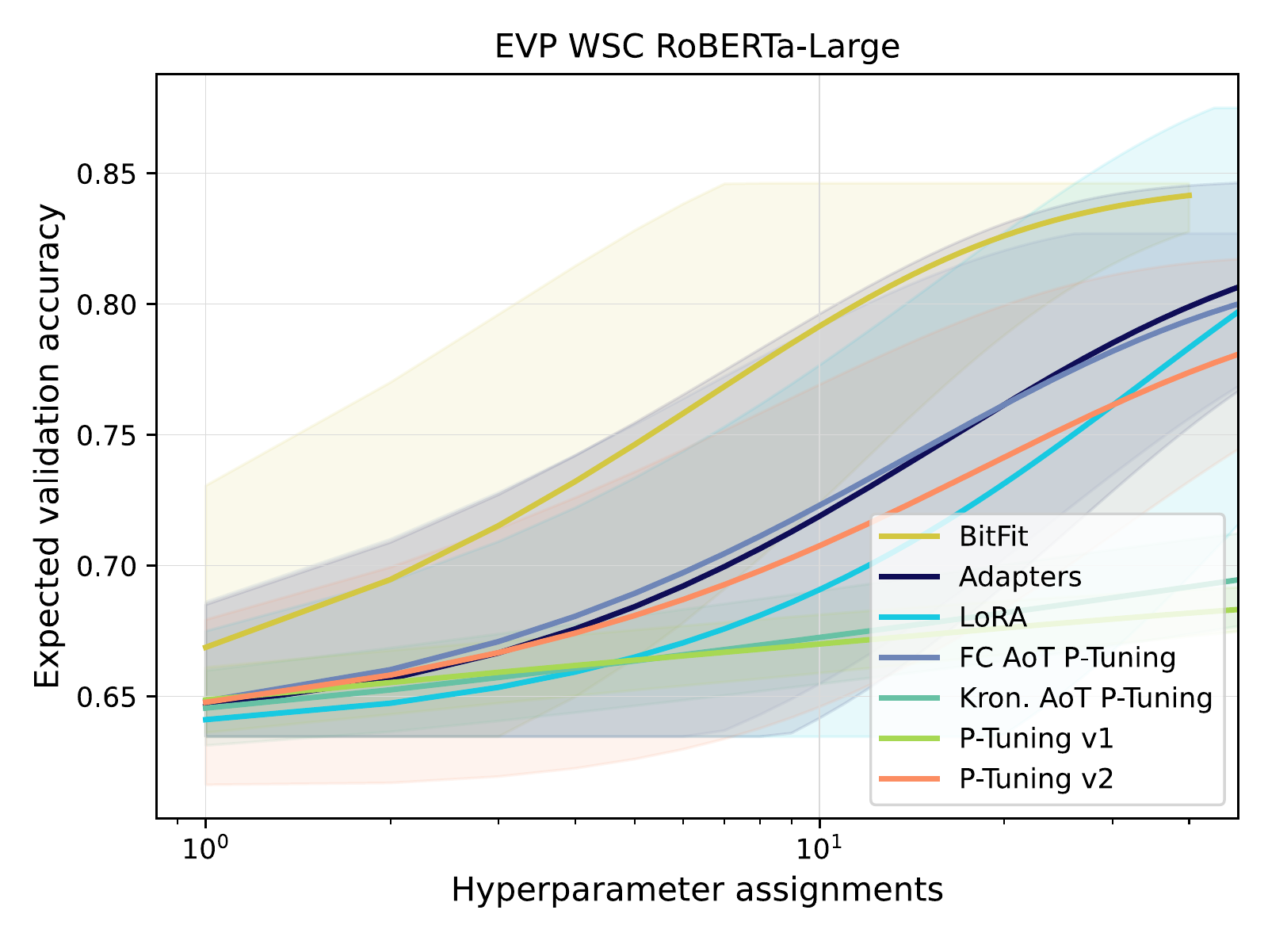}
    \caption{}
  \end{subfigure}
      \begin{subfigure}[t]{.26\linewidth}
    \centering\includegraphics[width=\linewidth]{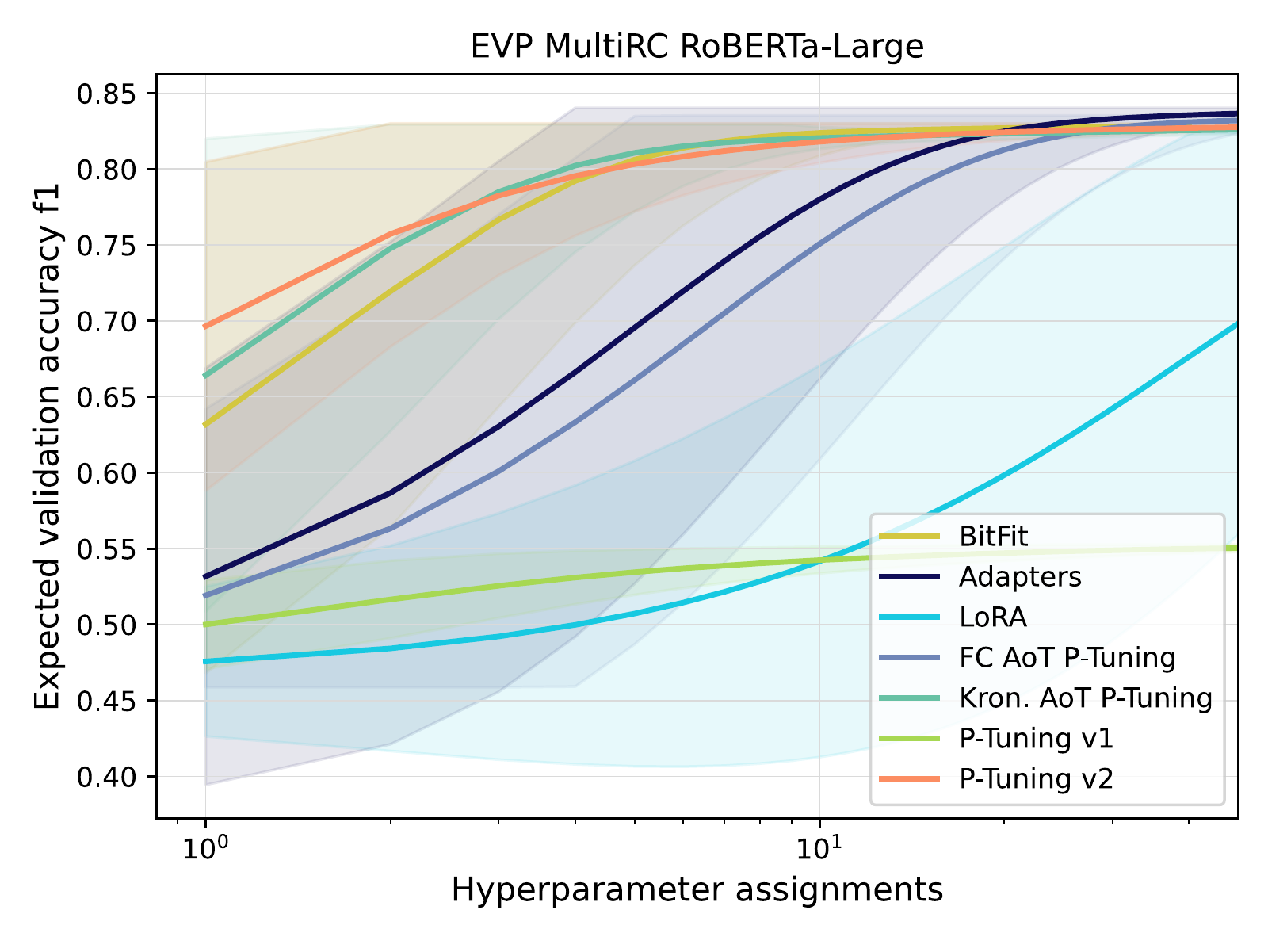}
    \caption{}
  \end{subfigure}
    \begin{subfigure}[t]{.24\linewidth}
    \centering\includegraphics[width=\linewidth]{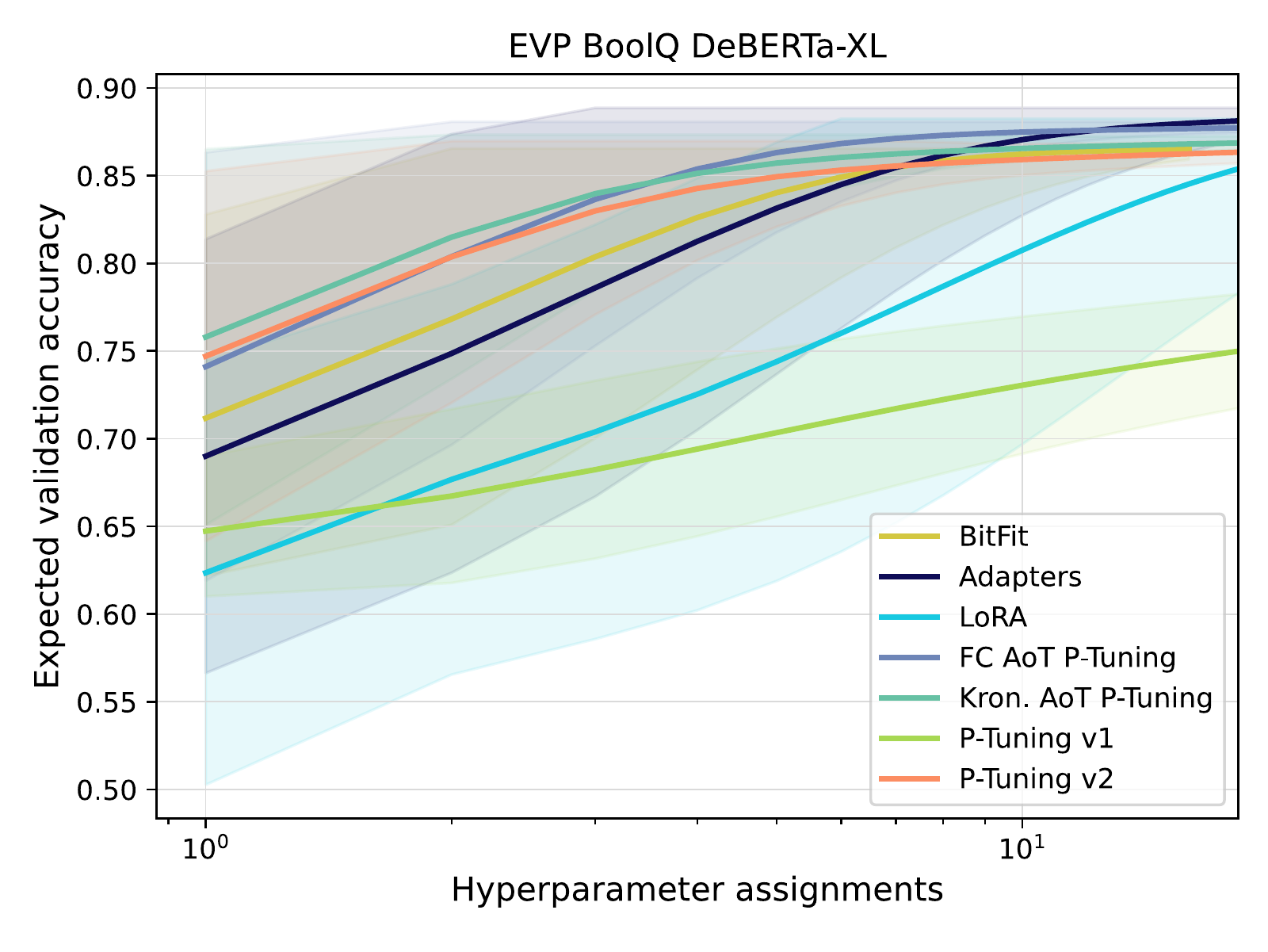}
    \caption{}
  \end{subfigure}
    \begin{subfigure}[t]{.24\linewidth}
    \centering\includegraphics[width=\linewidth]{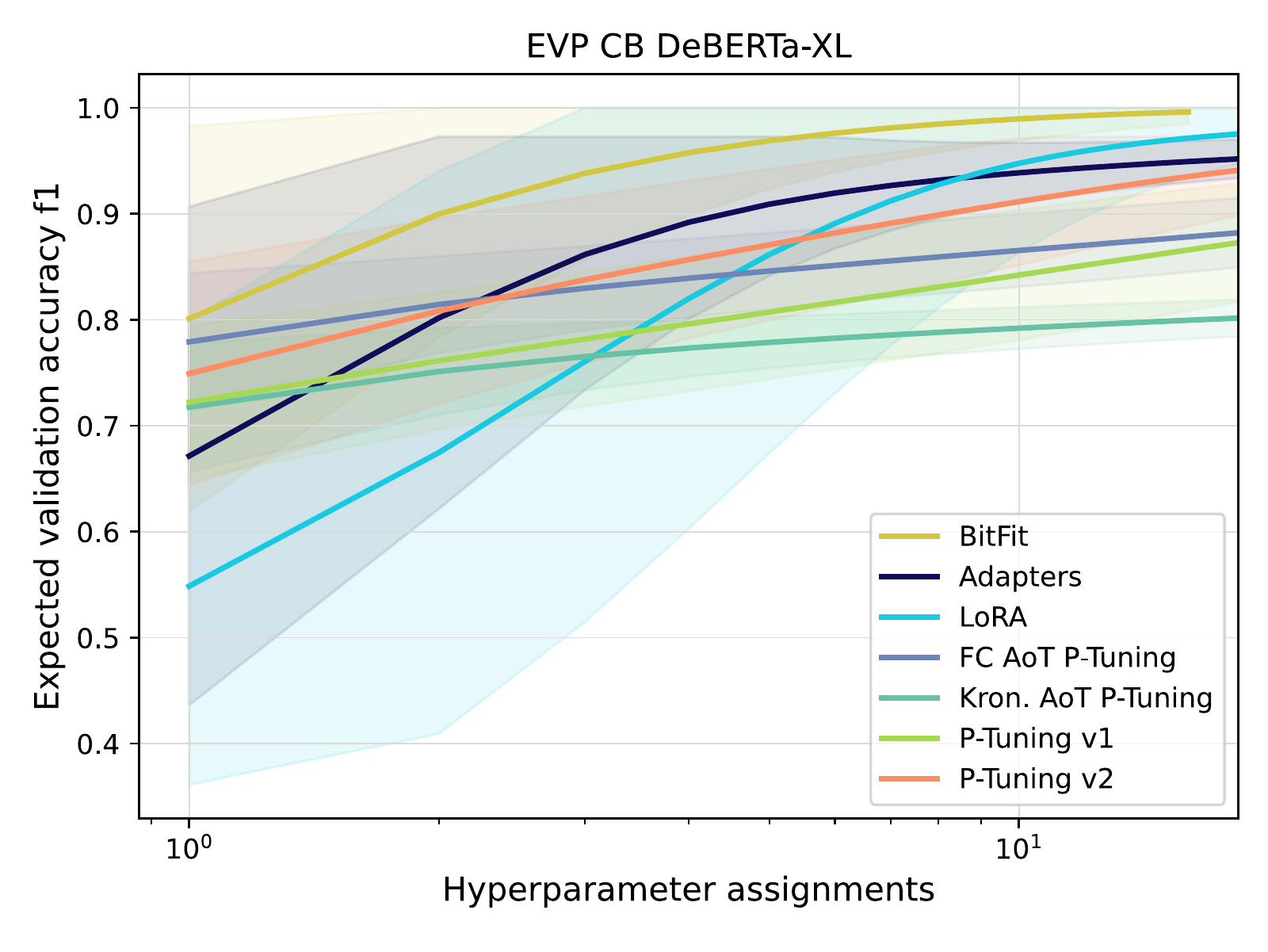}
    \caption{}
  \end{subfigure}
    \begin{subfigure}[t]{.24\linewidth}
    \centering\includegraphics[width=\linewidth]{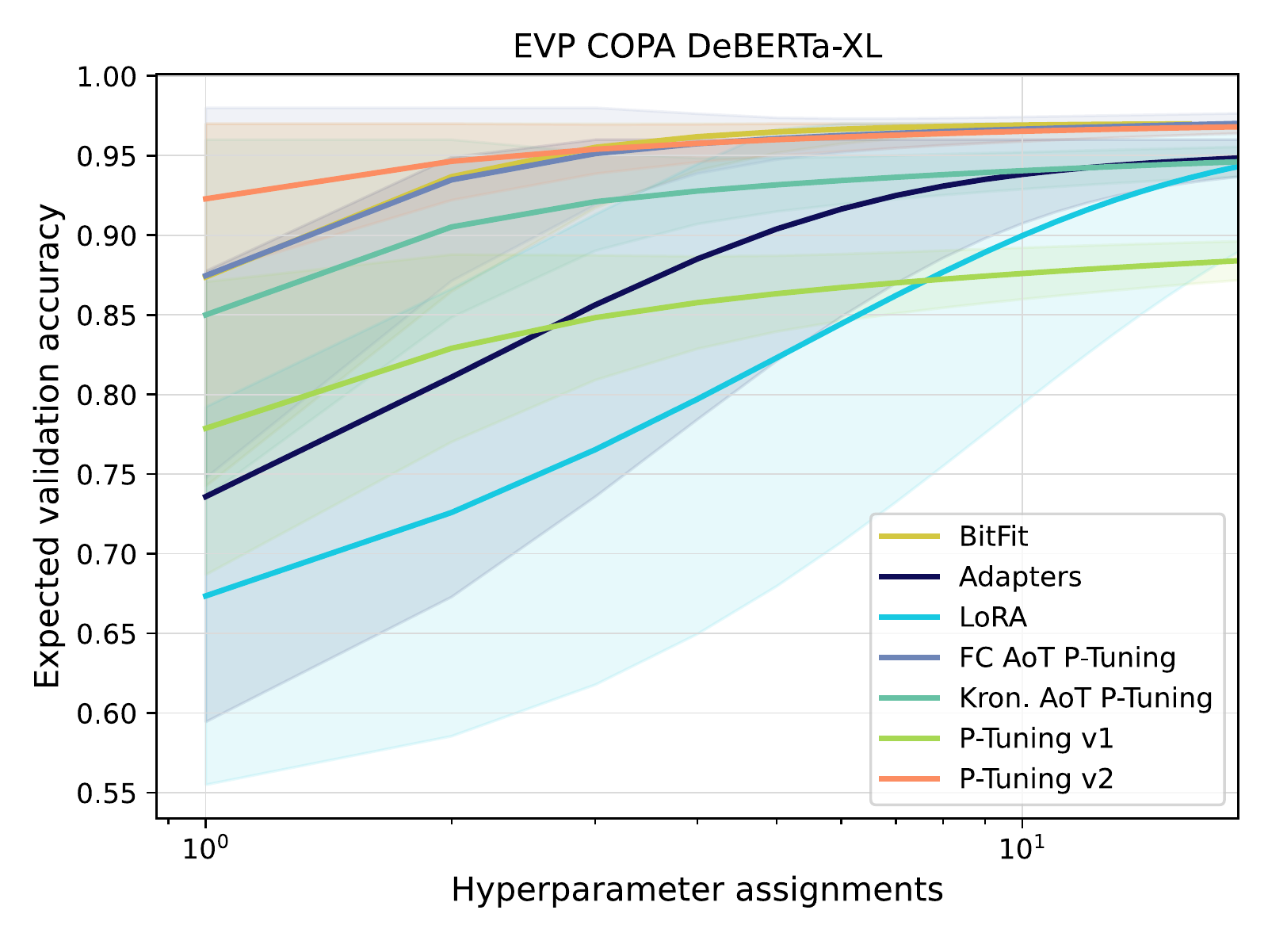}
    \caption{}
  \end{subfigure}
      \begin{subfigure}[t]{.24\linewidth}
    \centering\includegraphics[width=\linewidth]{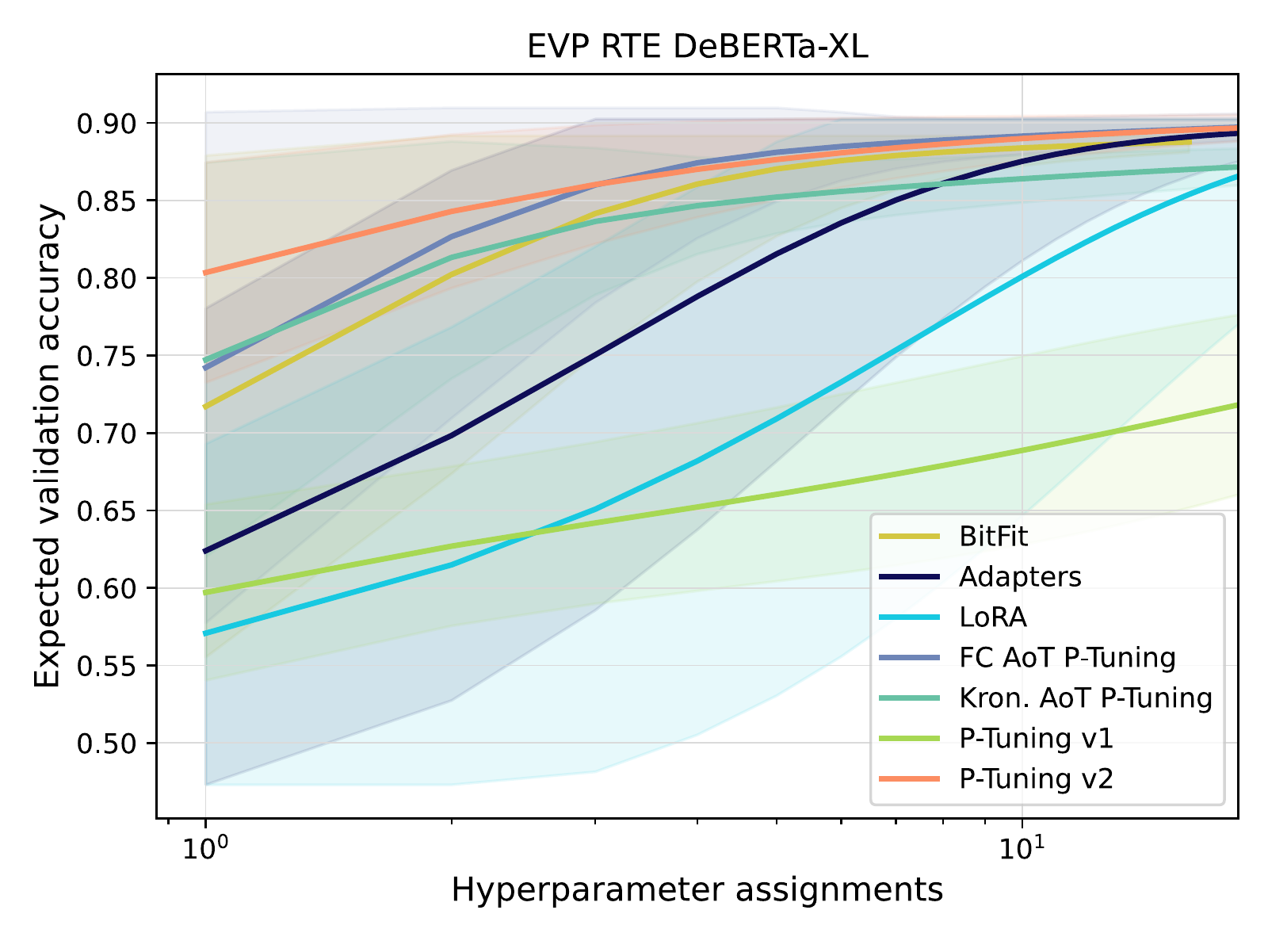}
    \caption{}
  \end{subfigure}
      \begin{subfigure}[t]{.26\linewidth}
    \centering\includegraphics[width=\linewidth]{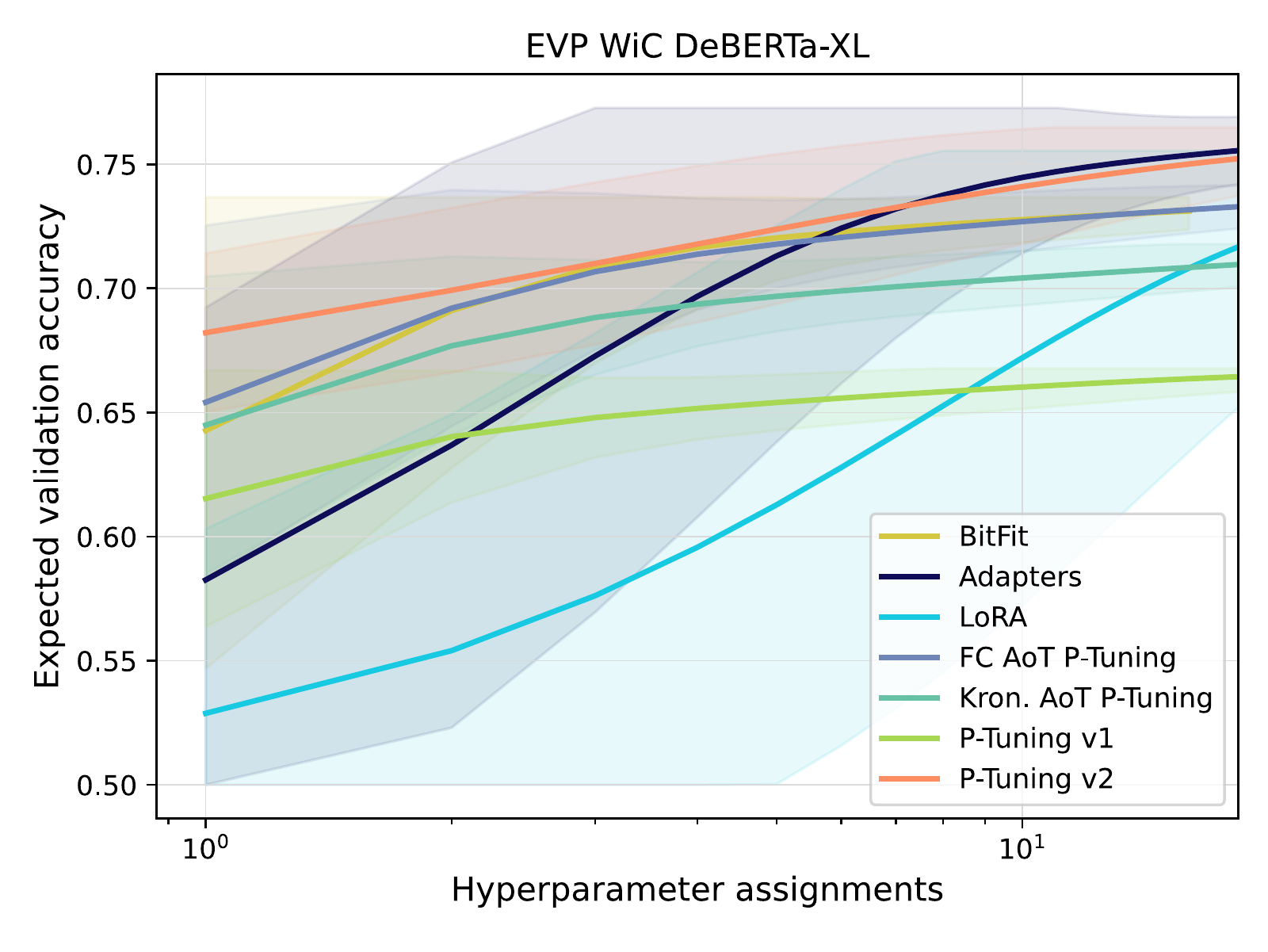}
    \caption{}
  \end{subfigure}
      \begin{subfigure}[t]{.26\linewidth}
    \centering\includegraphics[width=\linewidth]{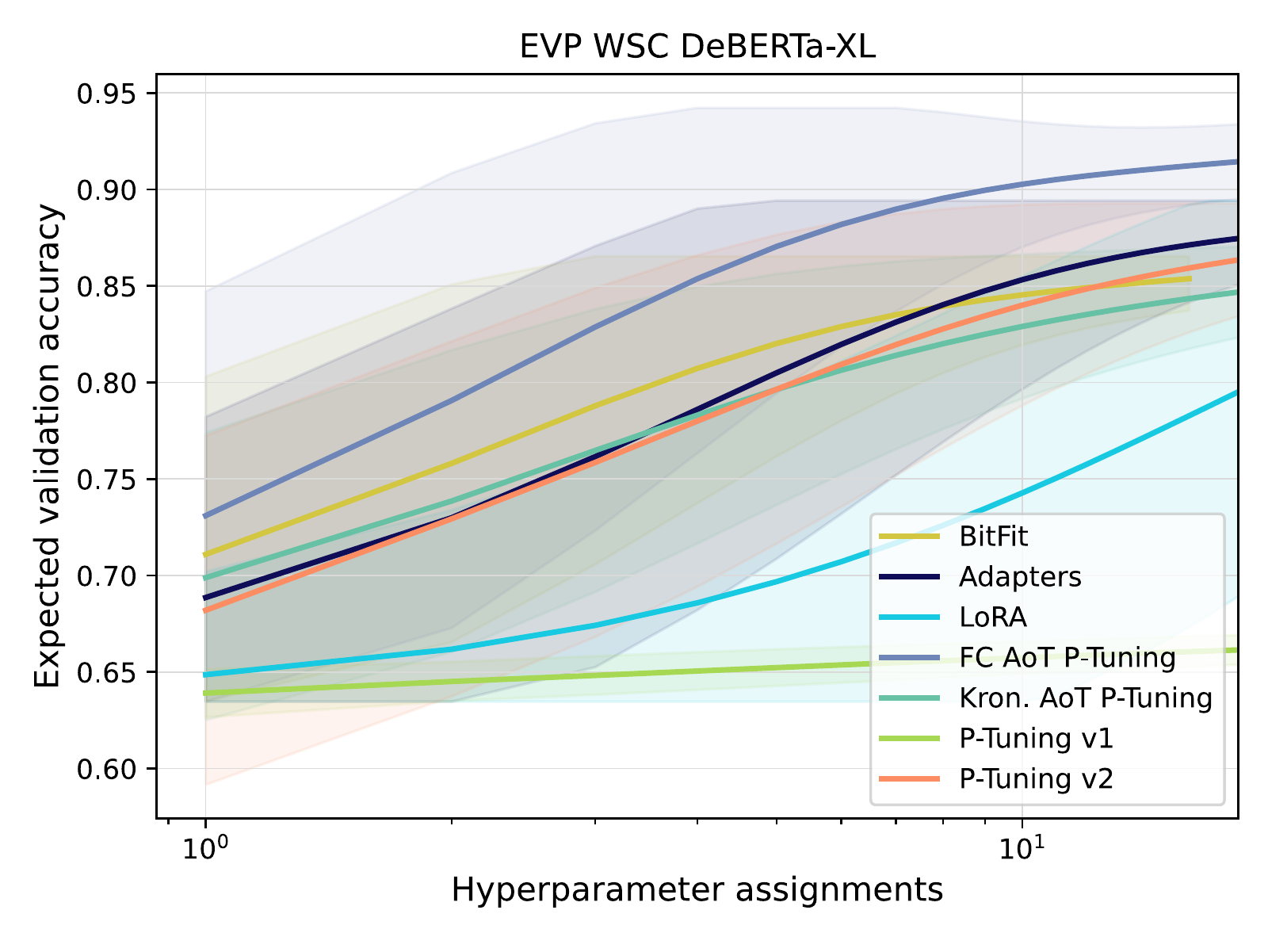}
    \caption{}
  \end{subfigure}
      \begin{subfigure}[t]{.26\linewidth}
    \centering\includegraphics[width=\linewidth]{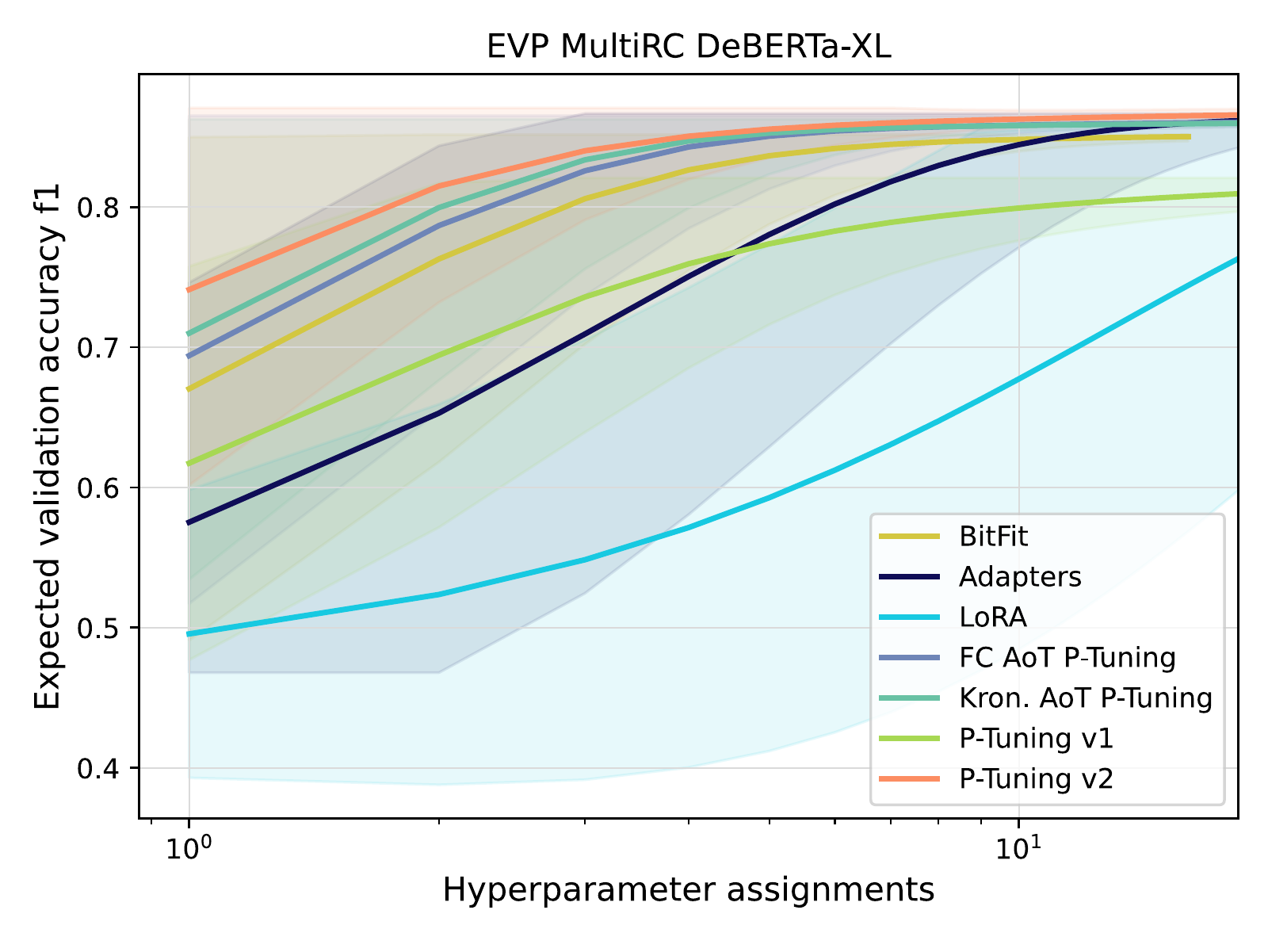}
    \caption{}
  \end{subfigure}
  \caption{Expected Validation Performance of trained models with SuperGLUE Benchmarking Datasets for RoBERTa-Base (a-g) and RoBERTa-Large (h-n). See Section 4.2 for more details.}
  \label{figure-evp-superglue}
\end{figure*}

\begin{figure*}[h!]
  \centering
      \medskip
      
      \begin{subfigure}[t]{.32\linewidth}
    \centering\includegraphics[width=\linewidth]{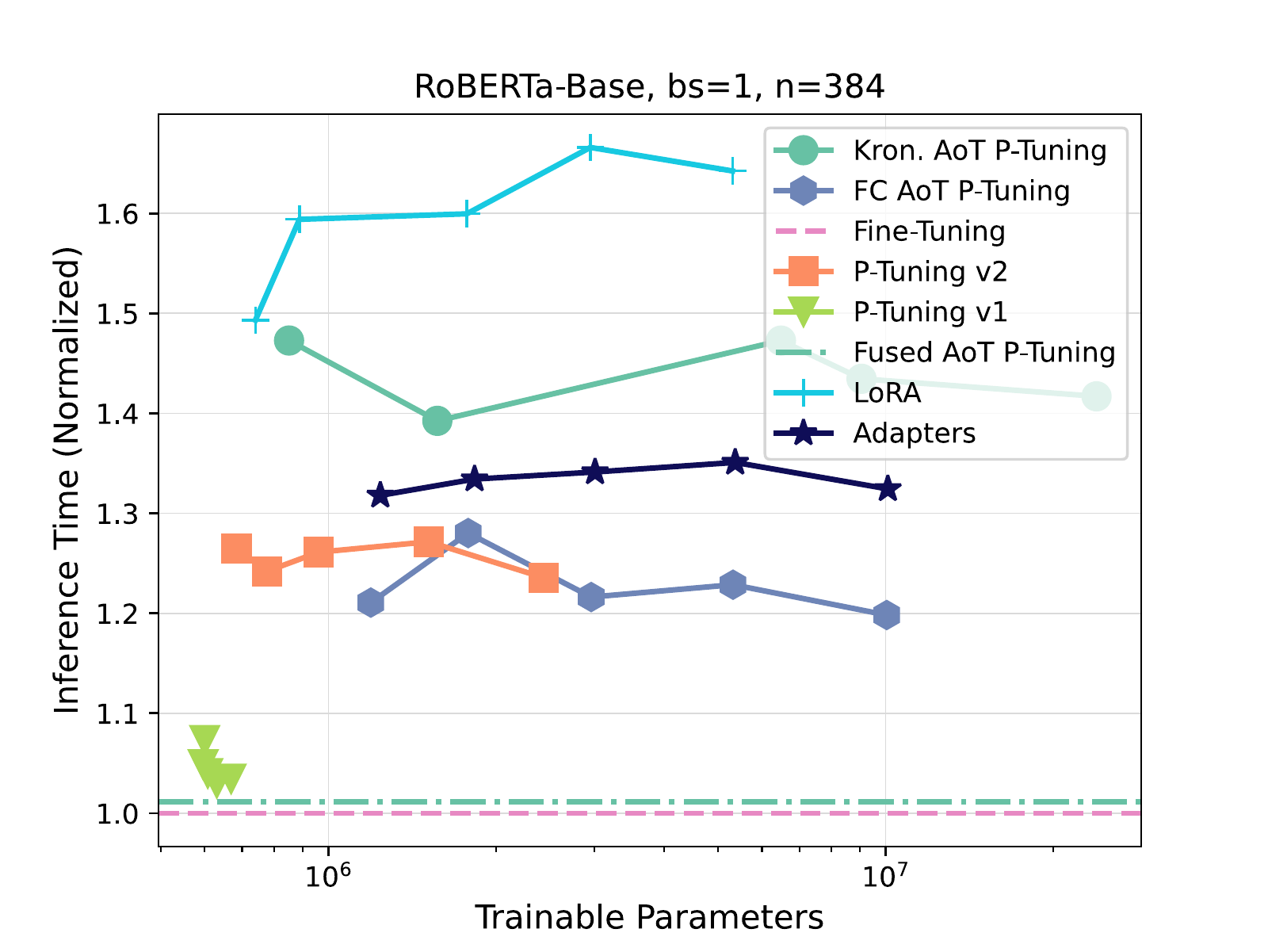}
    \caption{}
  \end{subfigure}
    \begin{subfigure}[t]{.32\linewidth}
    \centering\includegraphics[width=\linewidth]{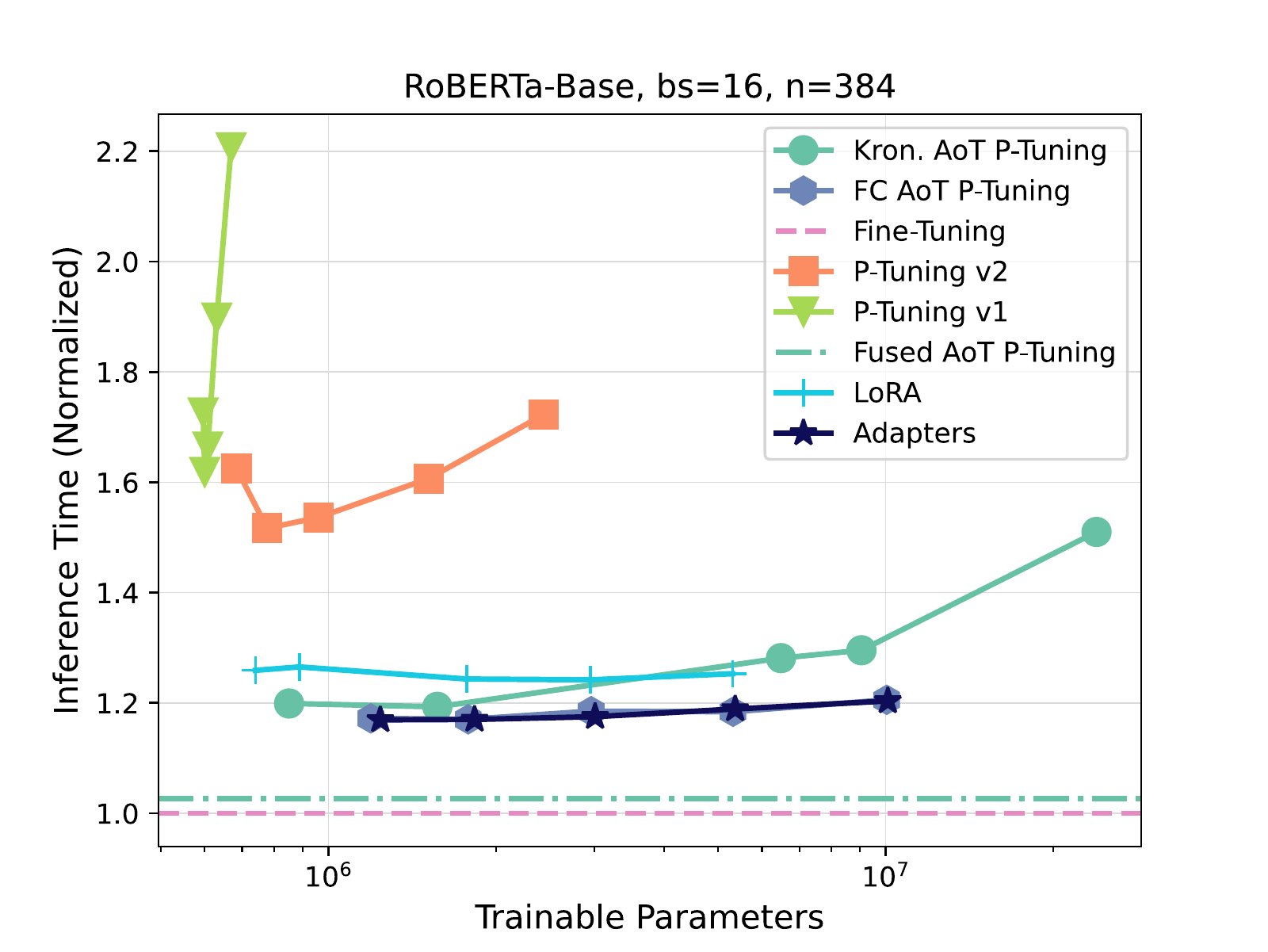}
    \caption{}
  \end{subfigure}
        \begin{subfigure}[t]{.32\linewidth}
    \centering\includegraphics[width=\linewidth]{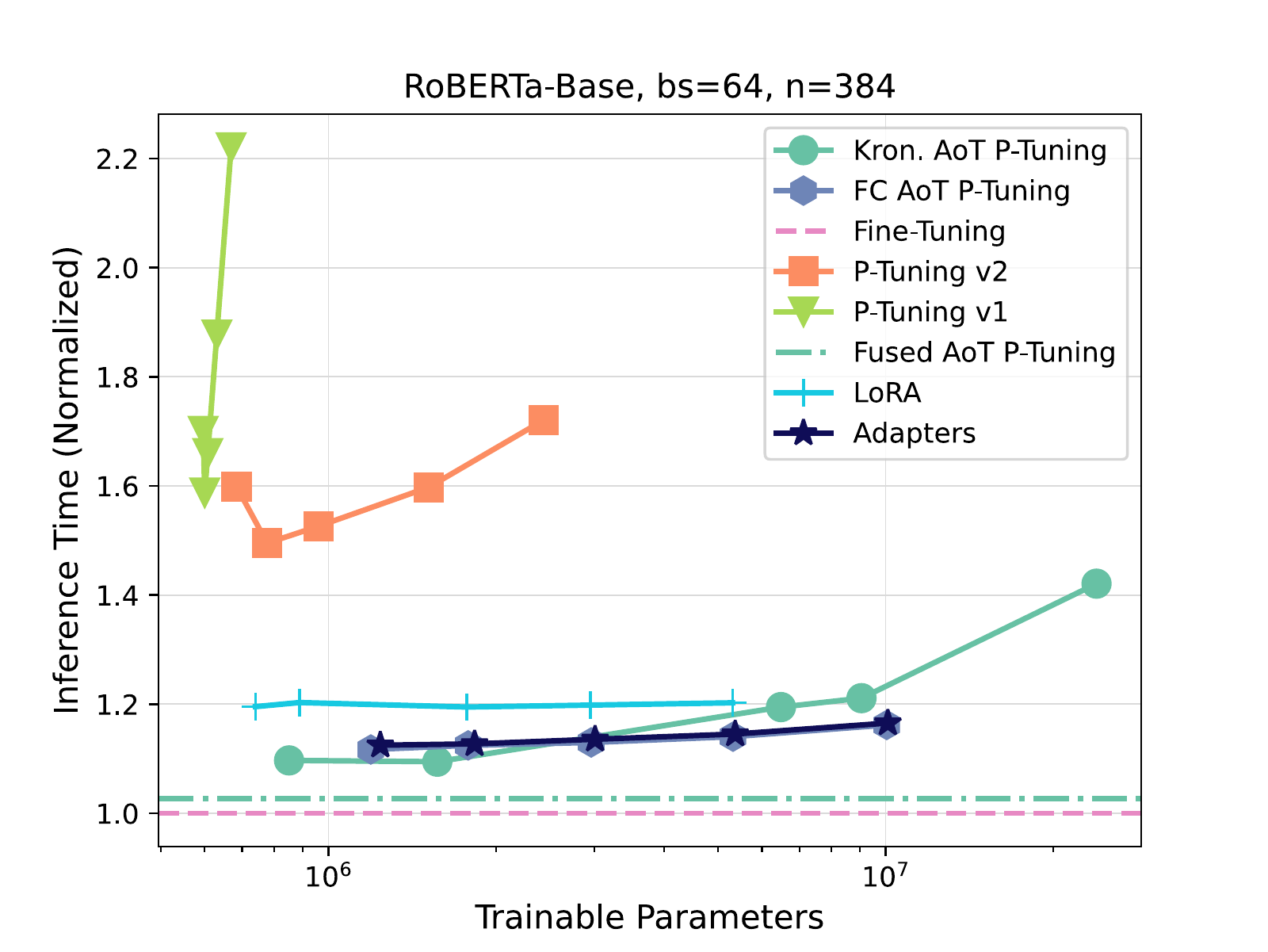}
    \caption{}
  \end{subfigure}
        \begin{subfigure}[t]{.32\linewidth}
    \centering\includegraphics[width=\linewidth]{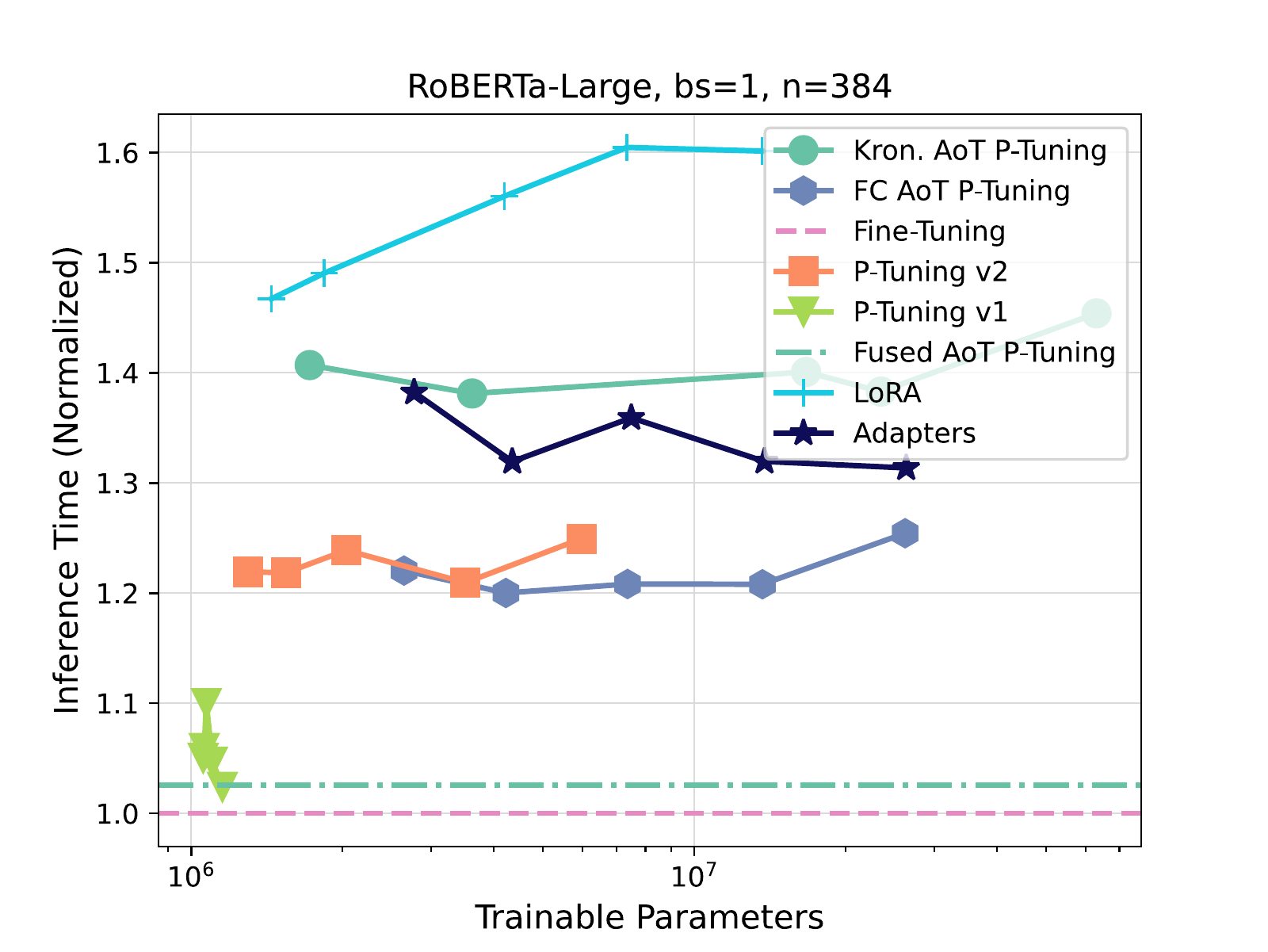}
    \caption{}
  \end{subfigure}
    \begin{subfigure}[t]{.32\linewidth}
    \centering\includegraphics[width=\linewidth]{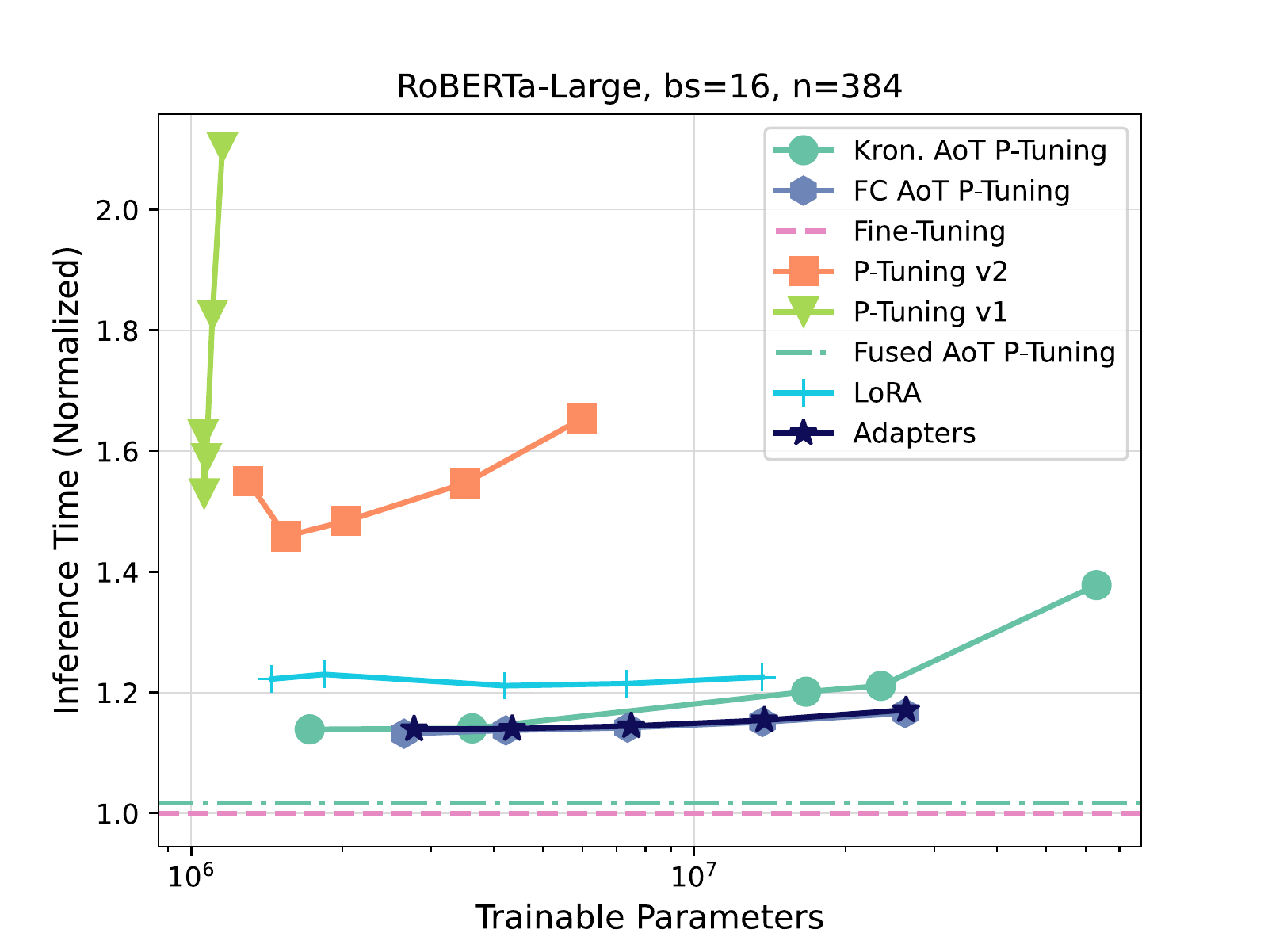}
    \caption{}
  \end{subfigure}
        \begin{subfigure}[t]{.32\linewidth}
    \centering\includegraphics[width=\linewidth]{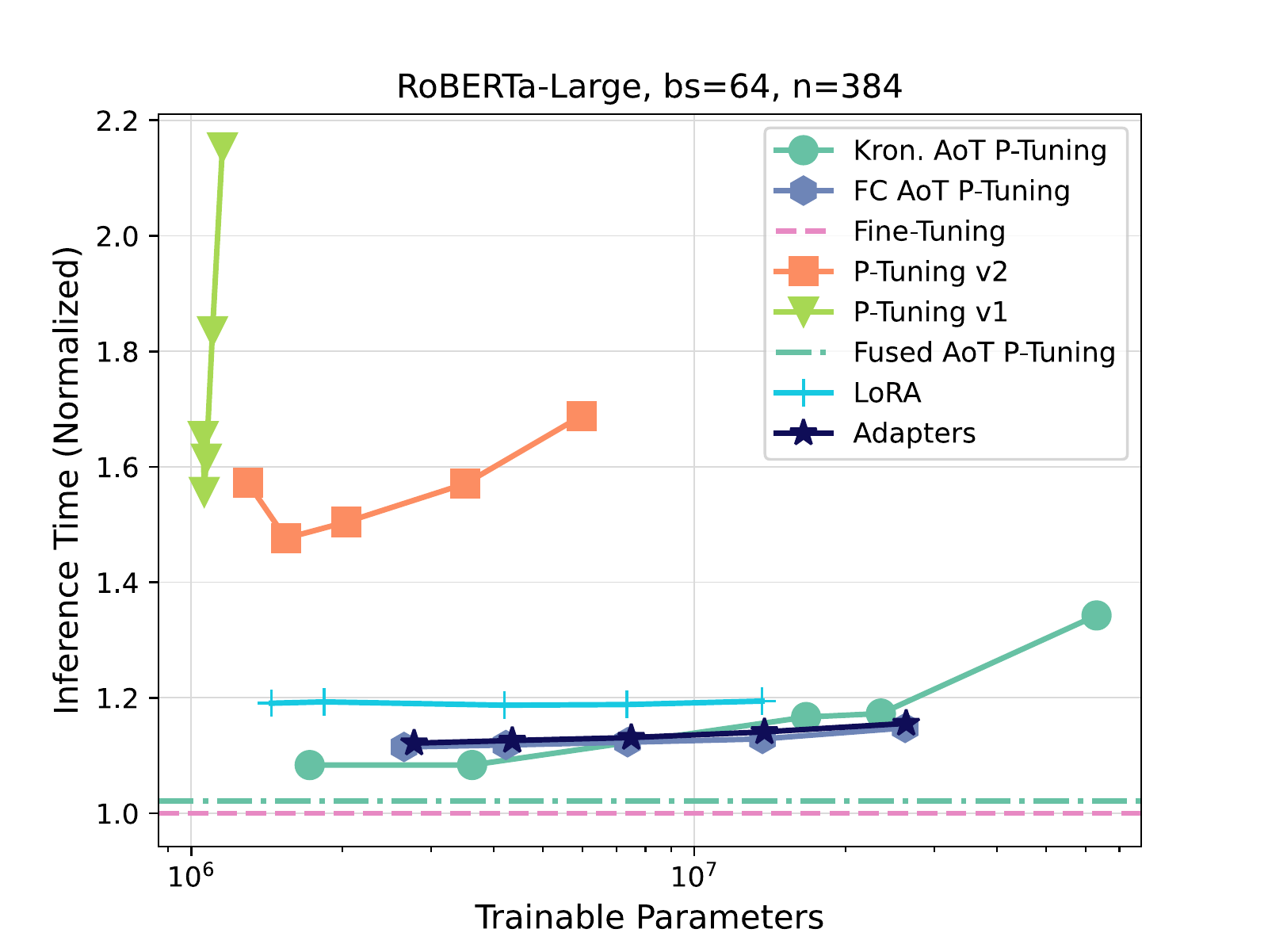}
    \caption{}
  \end{subfigure}
  
  \caption{Speed measurements for baseline methods with sequence length equal to $384$ for different back-bone models. See Section 4.4 for details.}
  \label{figure-speed-base-large}
\end{figure*}

\begin{figure*}[h!]
  \centering
      \medskip
      
      \begin{subfigure}[t]{.29\linewidth}
    \centering\includegraphics[width=\linewidth]{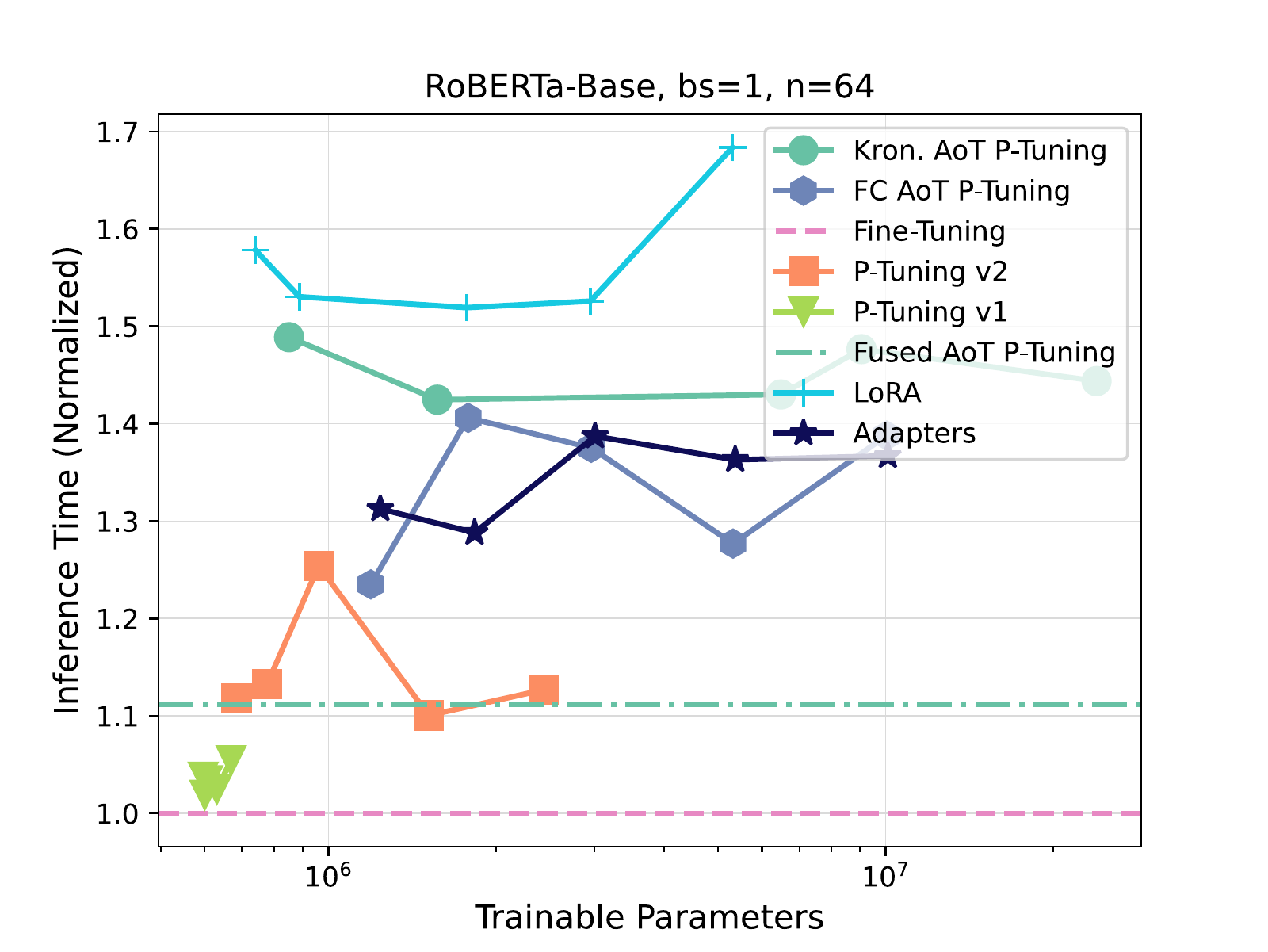}
    \caption{}
  \end{subfigure}
    \begin{subfigure}[t]{.29\linewidth}
    \centering\includegraphics[width=\linewidth]{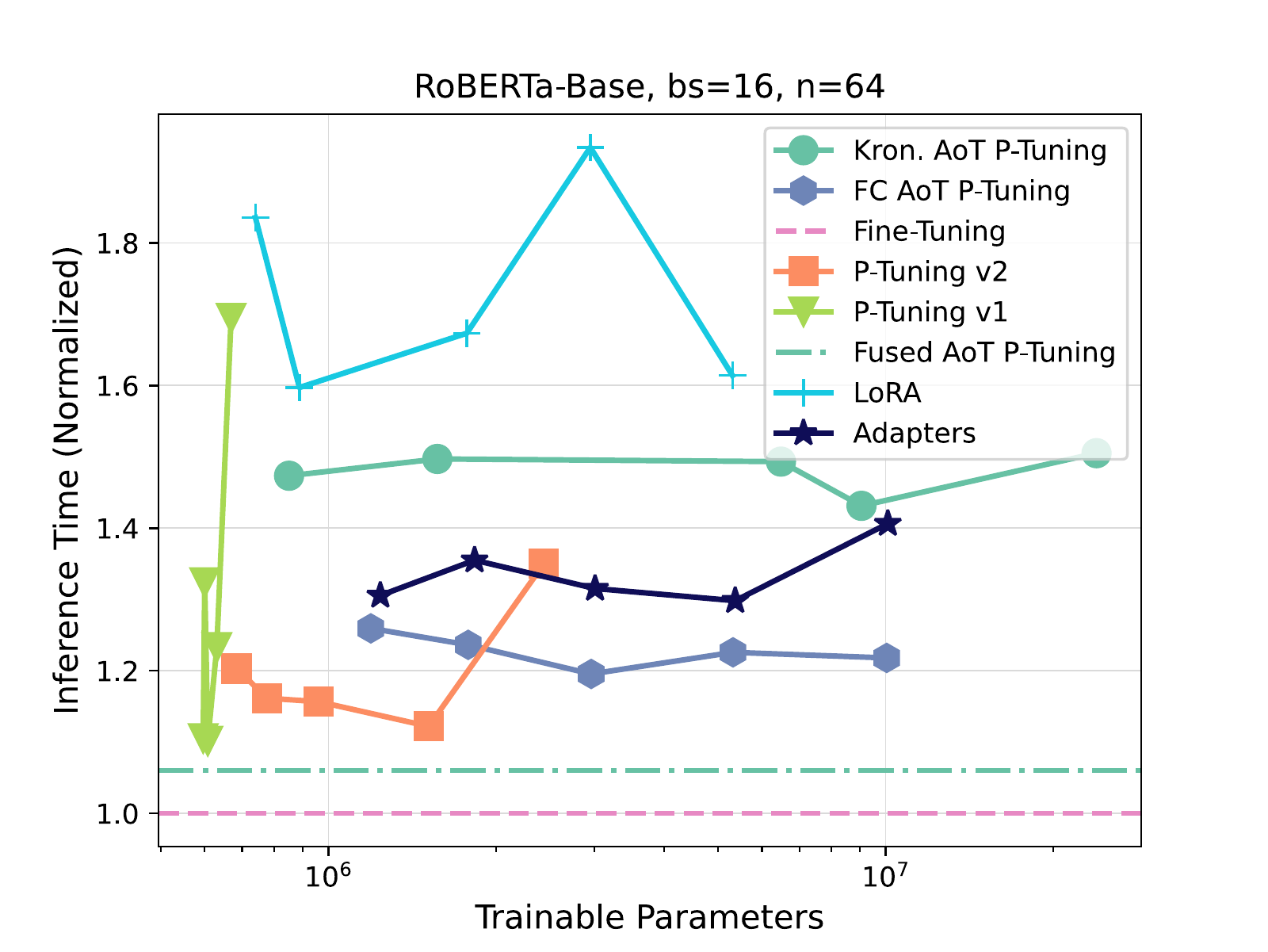}
    \caption{}
  \end{subfigure}
        \begin{subfigure}[t]{.29\linewidth}
    \centering\includegraphics[width=\linewidth]{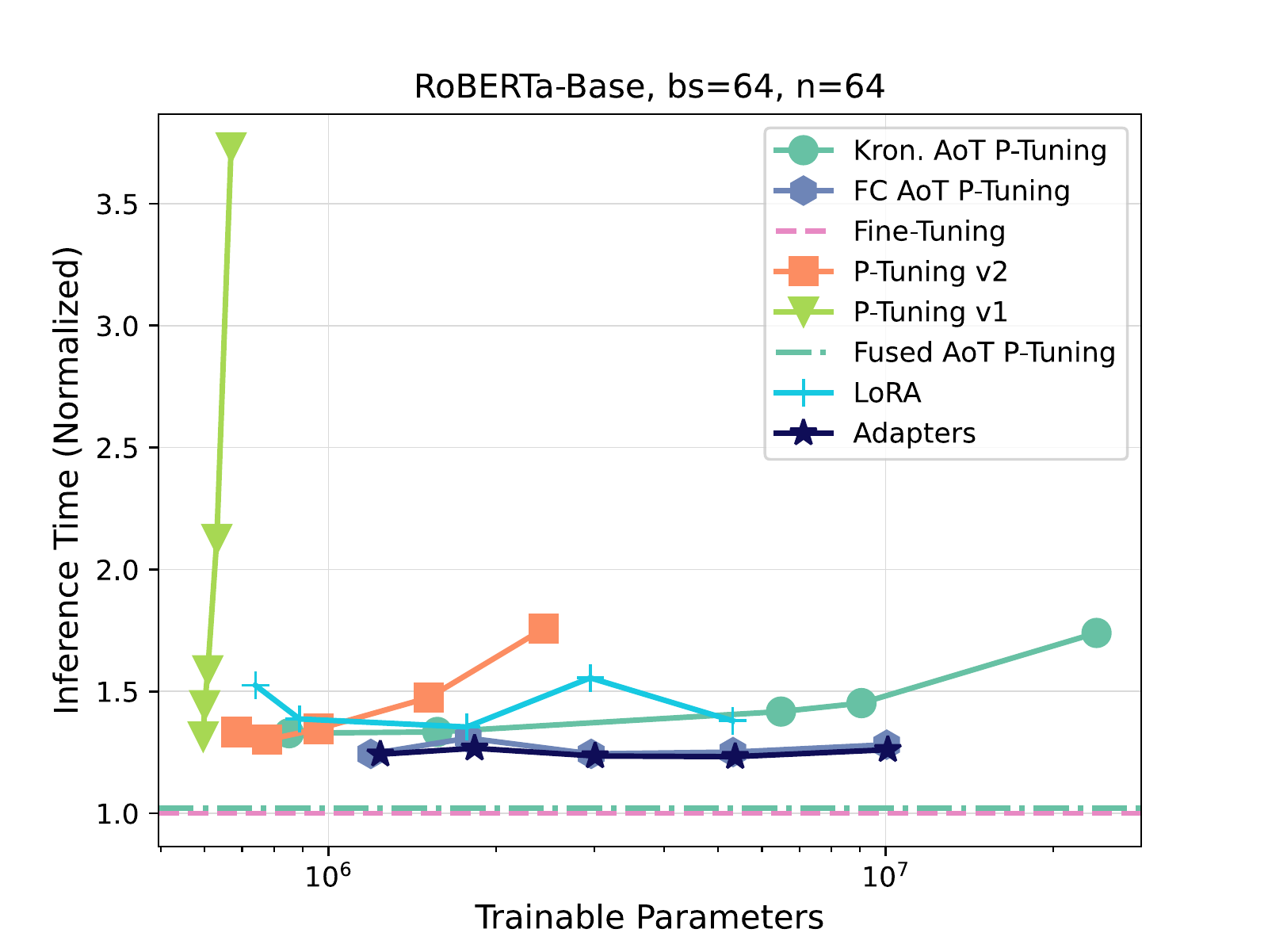}
    \caption{}
  \end{subfigure}
        \begin{subfigure}[t]{.29\linewidth}
    \centering\includegraphics[width=\linewidth]{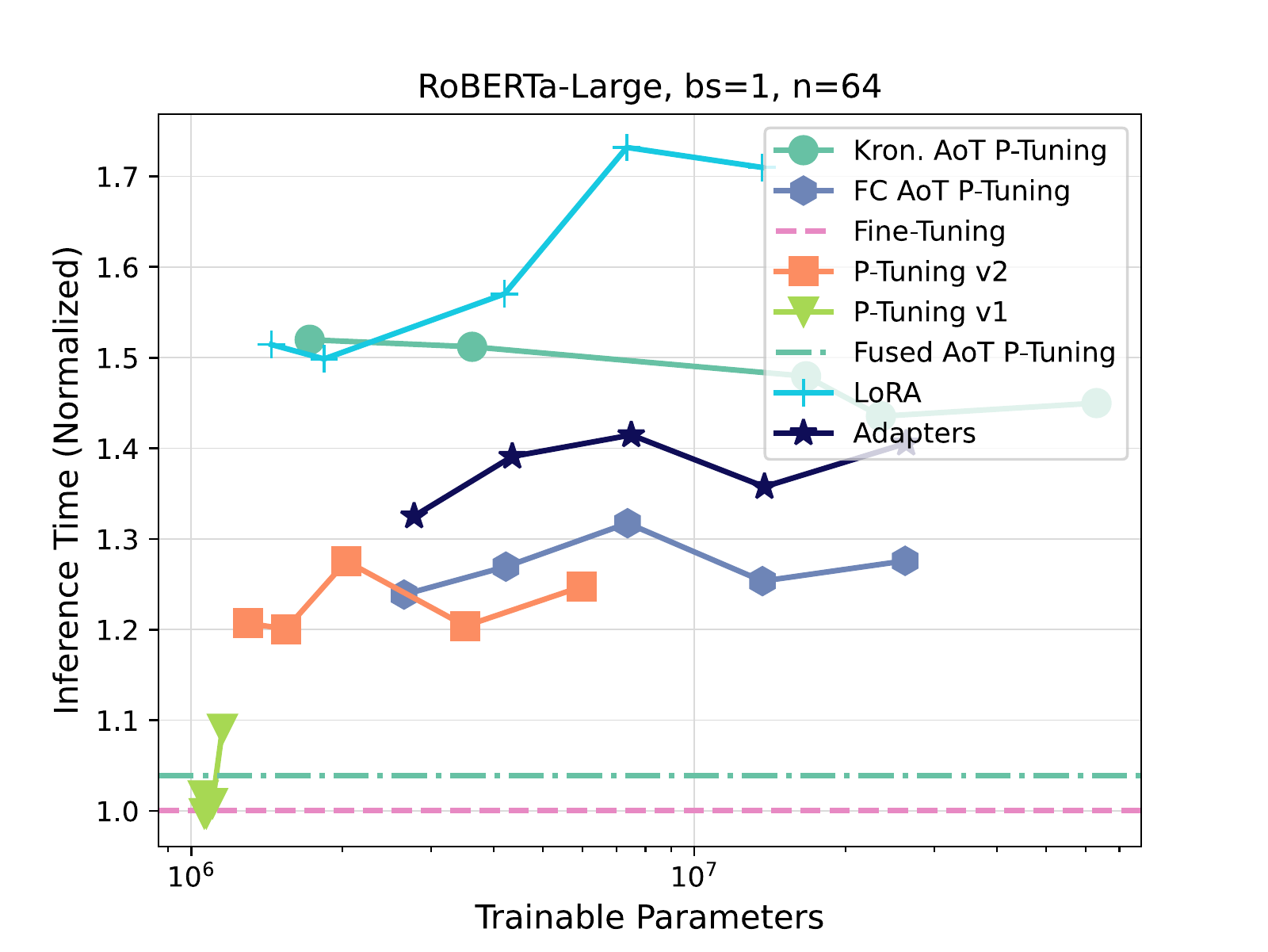}
    \caption{}
  \end{subfigure}
    \begin{subfigure}[t]{.29\linewidth}
    \centering\includegraphics[width=\linewidth]{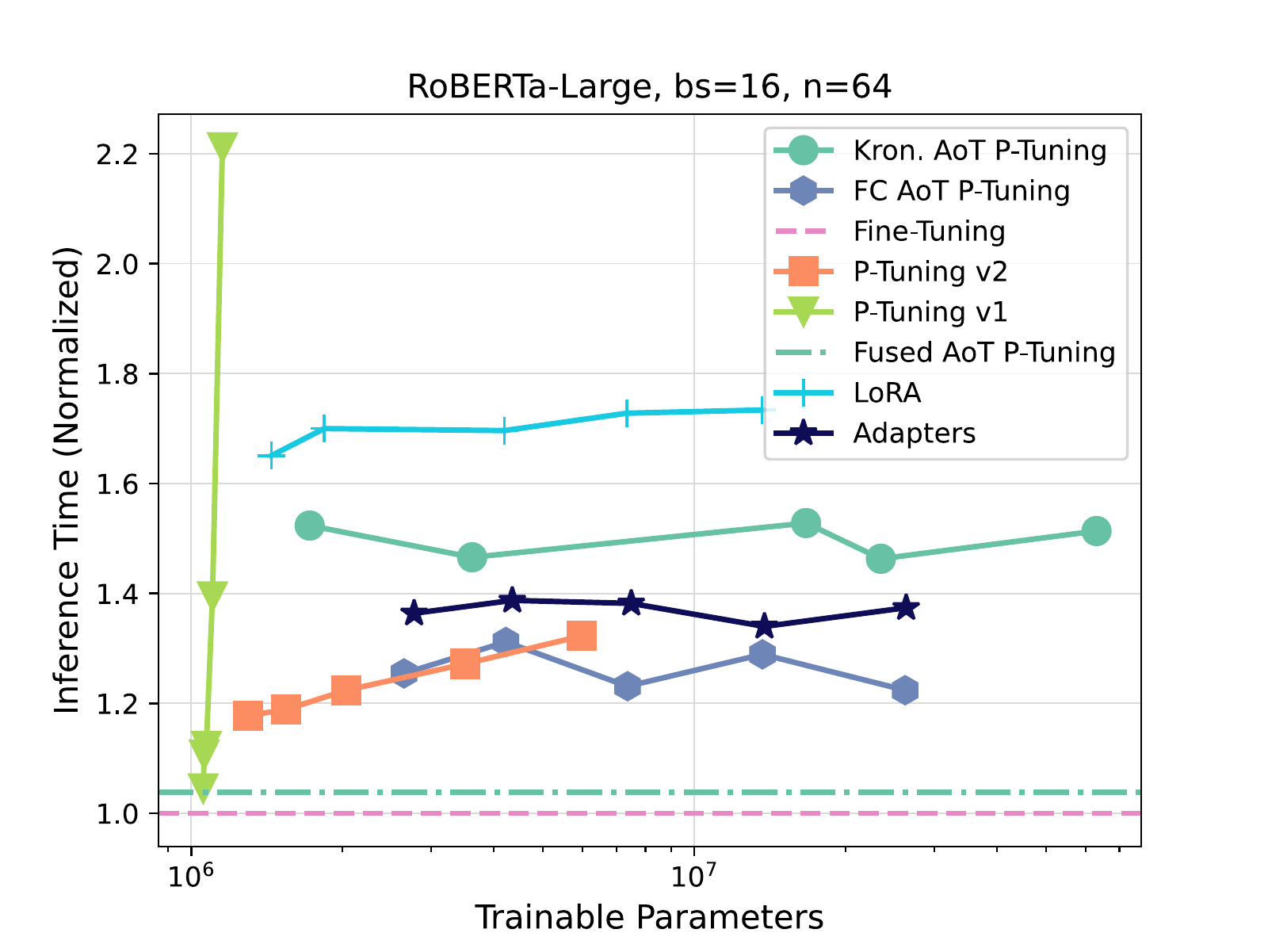}
    \caption{}
  \end{subfigure}
        \begin{subfigure}[t]{.29\linewidth}
    \centering\includegraphics[width=\linewidth]{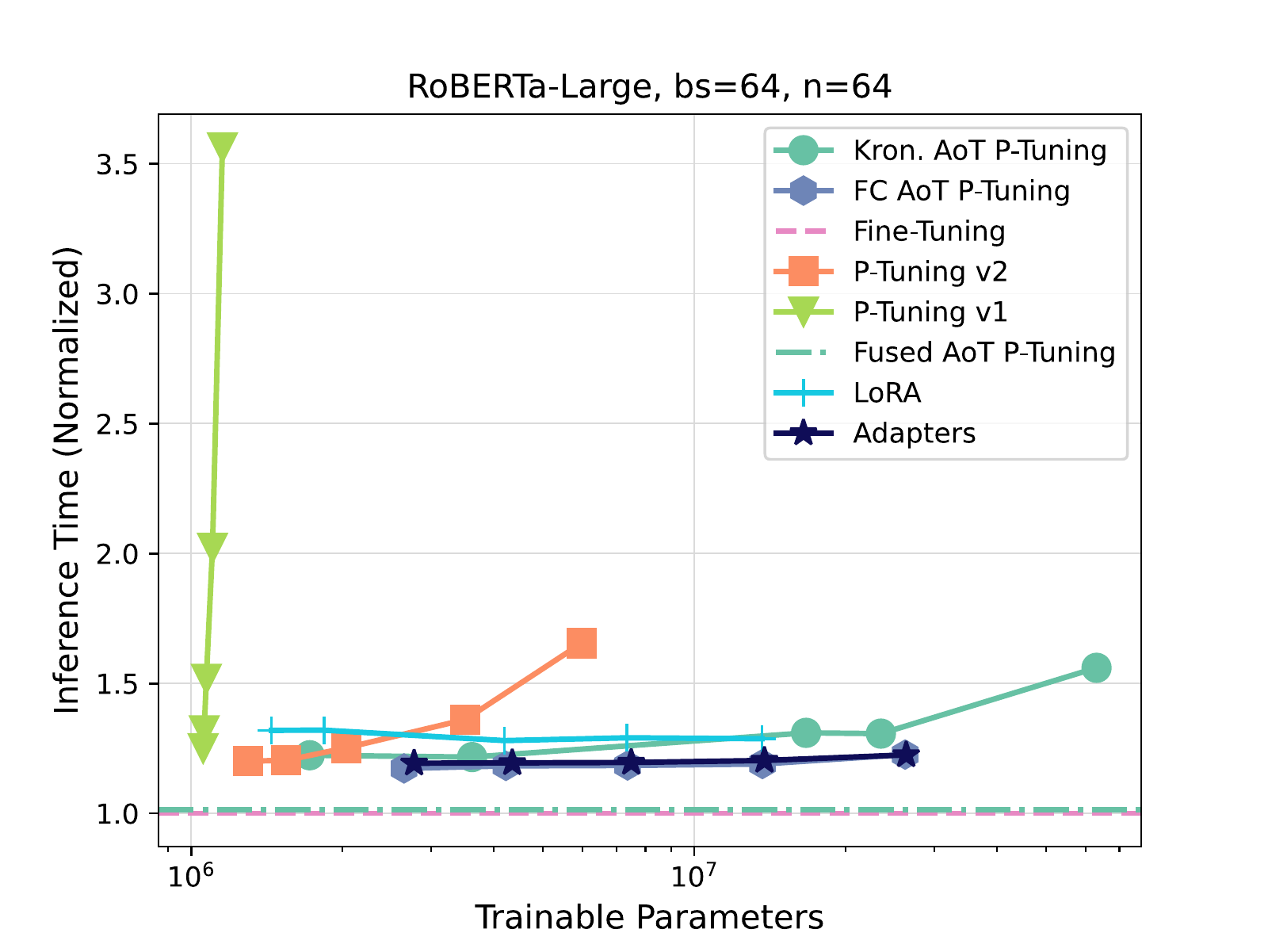}
    \caption{}
  \end{subfigure}
        \begin{subfigure}[t]{.29\linewidth}
    \centering\includegraphics[width=\linewidth]{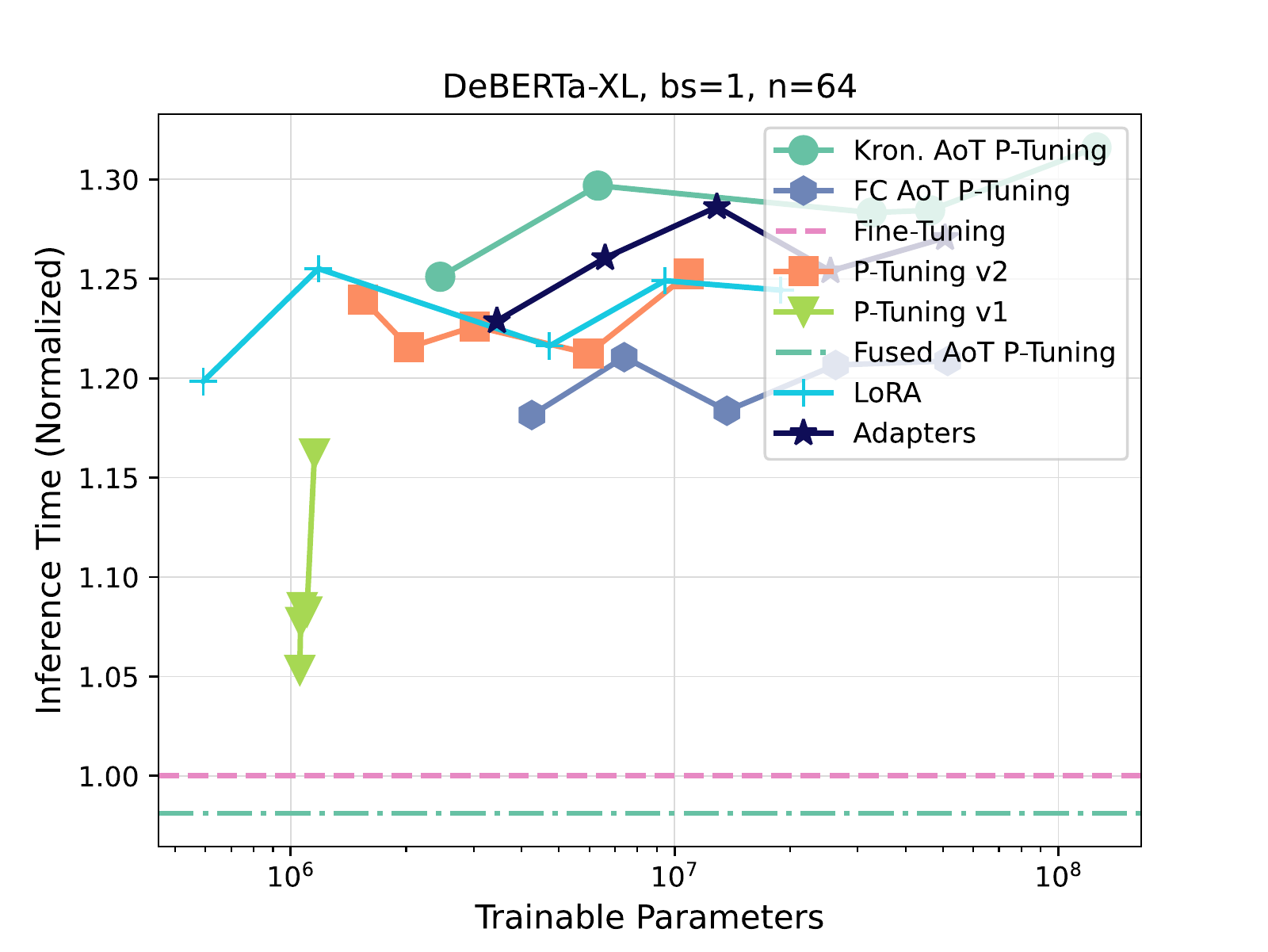}
    \caption{}
  \end{subfigure}
    \begin{subfigure}[t]{.29\linewidth}
    \centering\includegraphics[width=\linewidth]{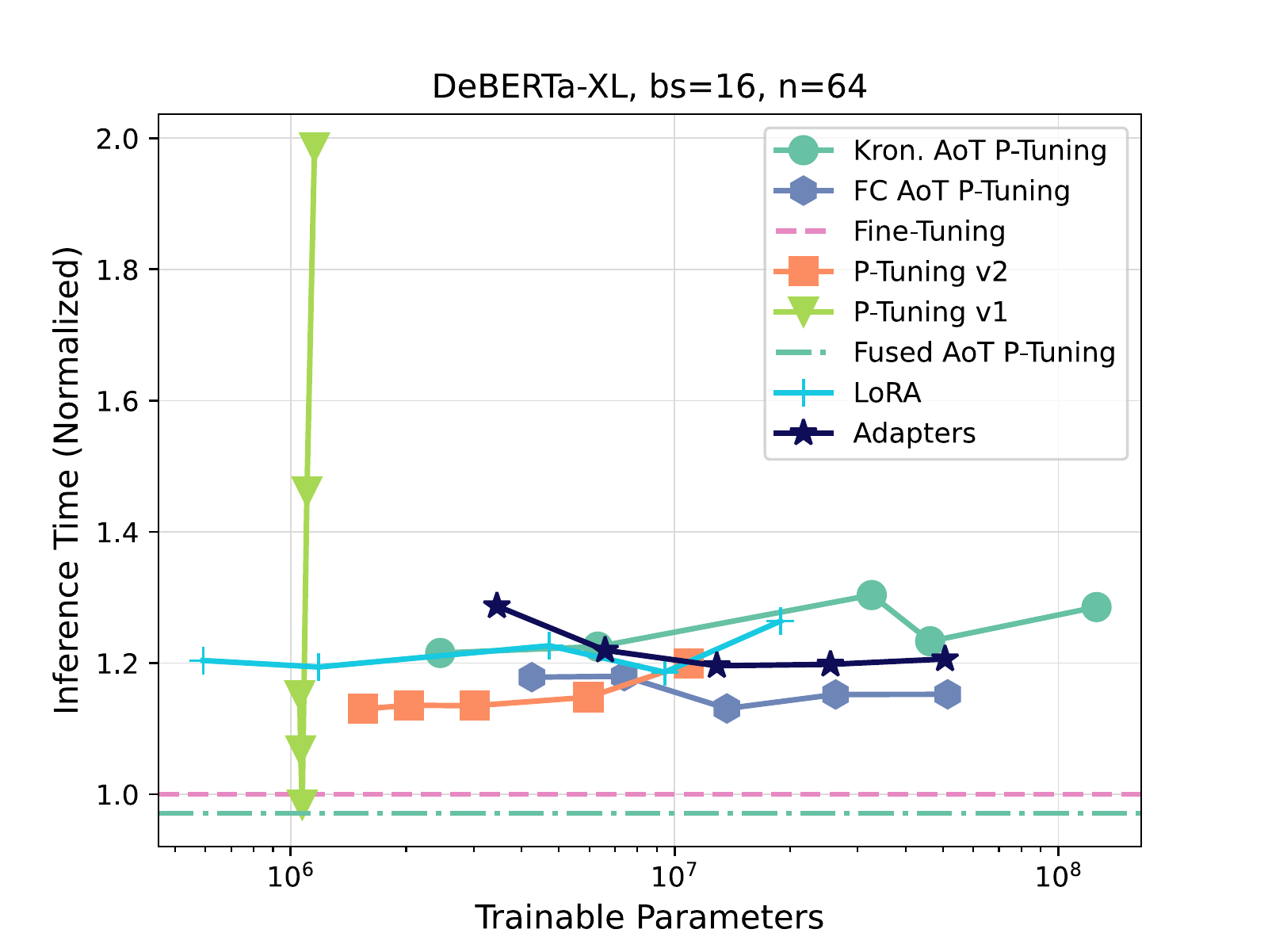}
    \caption{}
  \end{subfigure}
        \begin{subfigure}[t]{.29\linewidth}
    \centering\includegraphics[width=\linewidth]{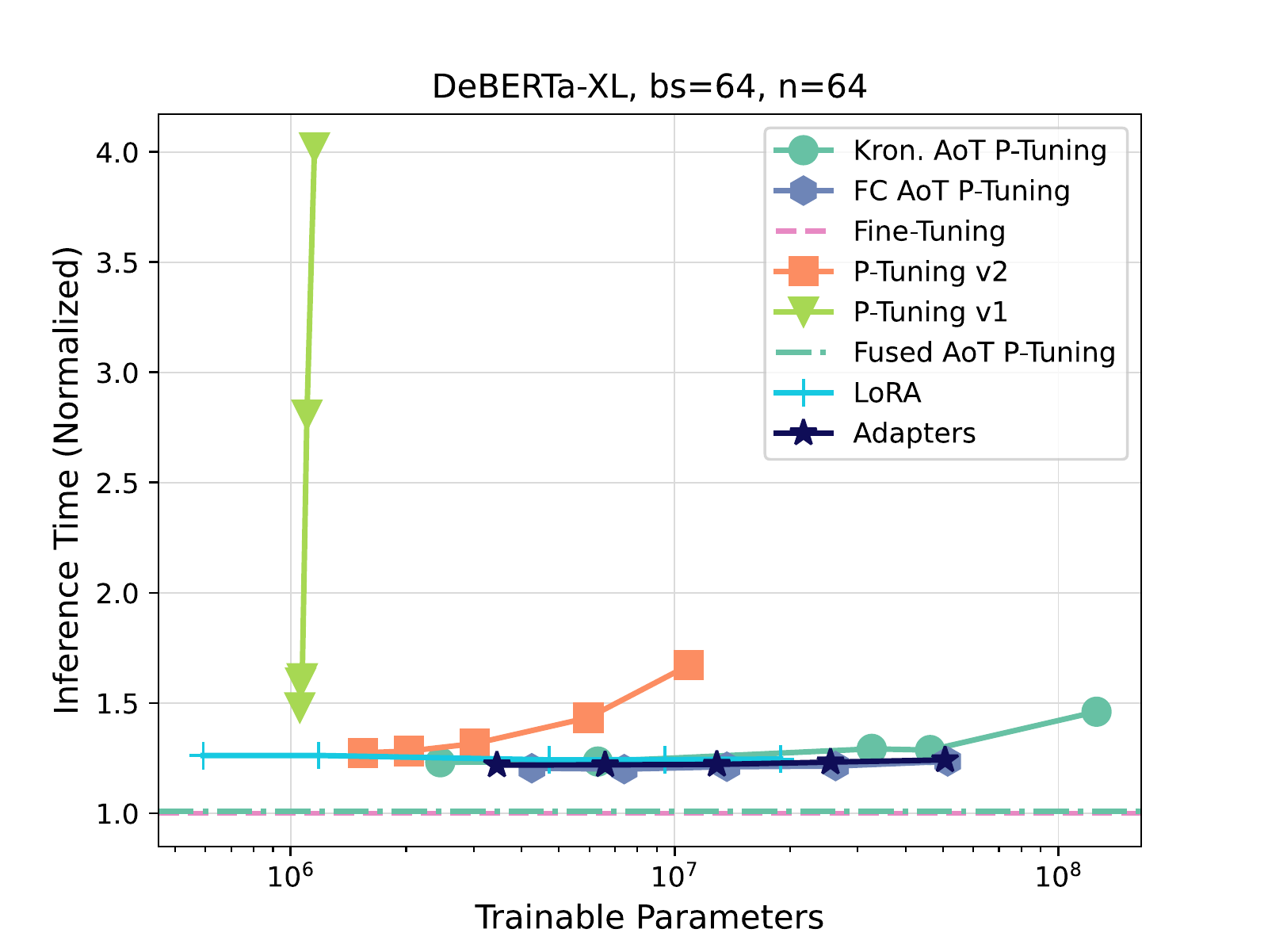}
    \caption{}
  \end{subfigure}
        \begin{subfigure}[t]{.29\linewidth}
    \centering\includegraphics[width=\linewidth]{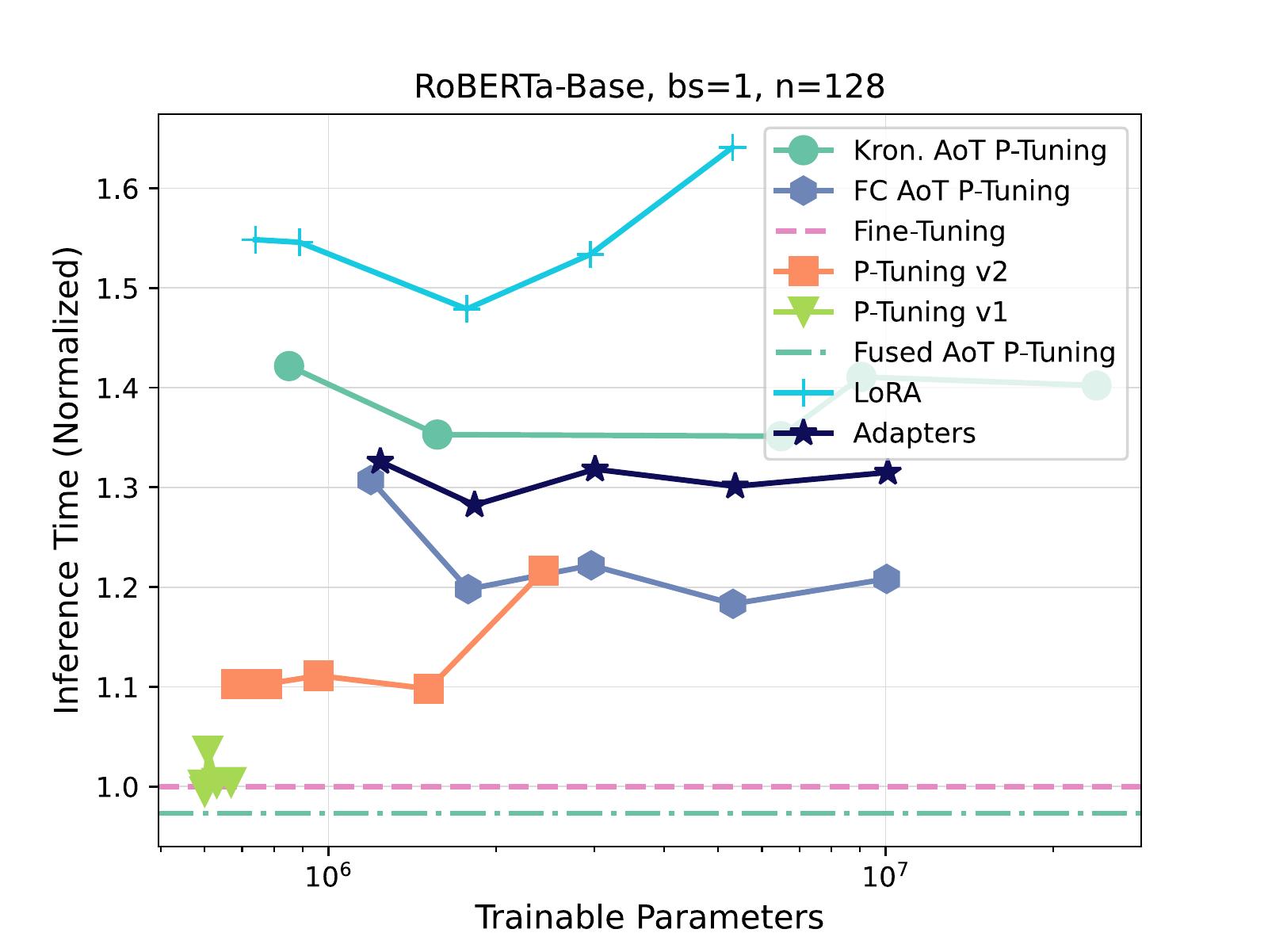}
    \caption{}
  \end{subfigure}
    \begin{subfigure}[t]{.29\linewidth}
    \centering\includegraphics[width=\linewidth]{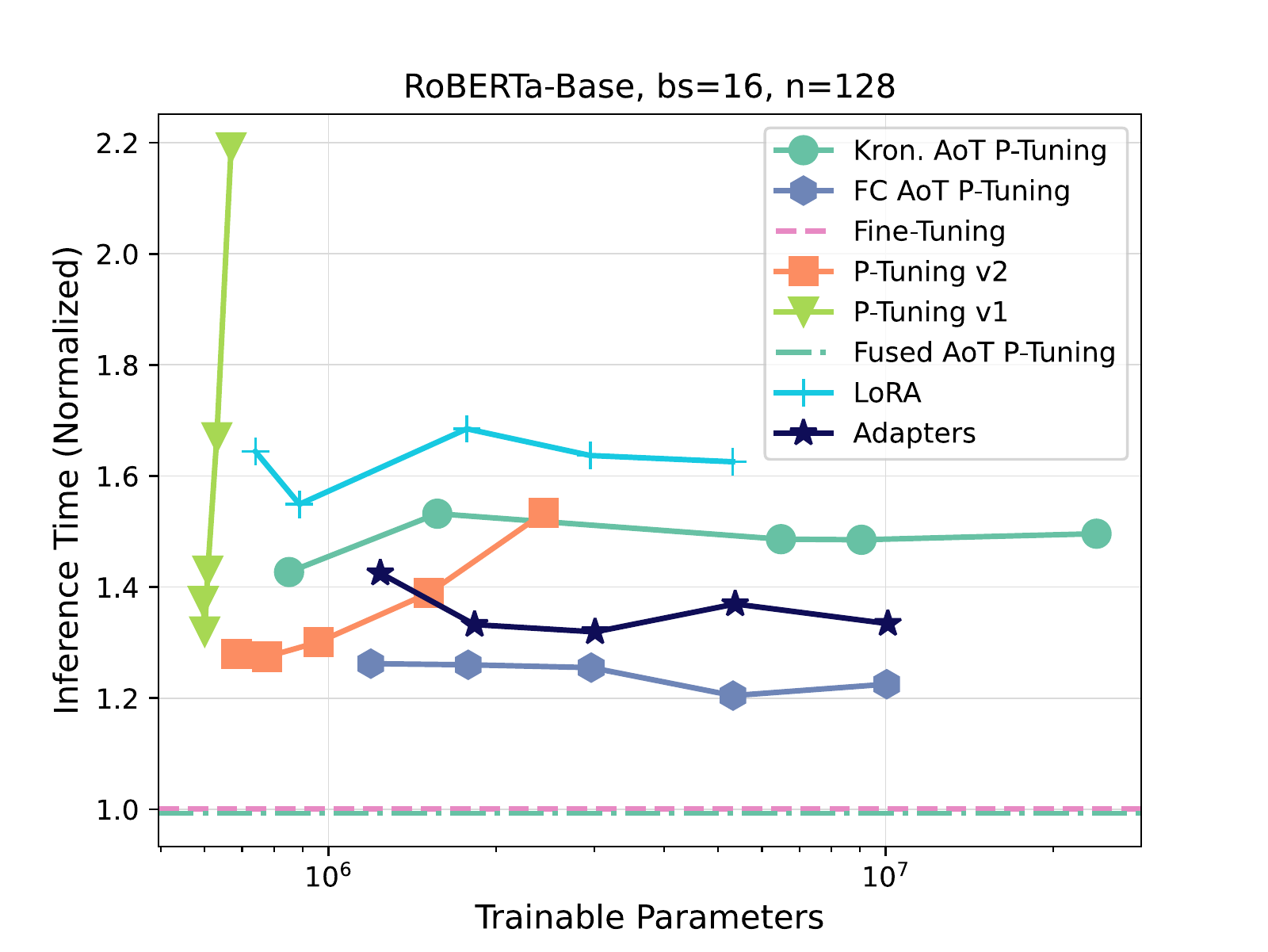}
    \caption{}
  \end{subfigure}
        \begin{subfigure}[t]{.29\linewidth}
    \centering\includegraphics[width=\linewidth]{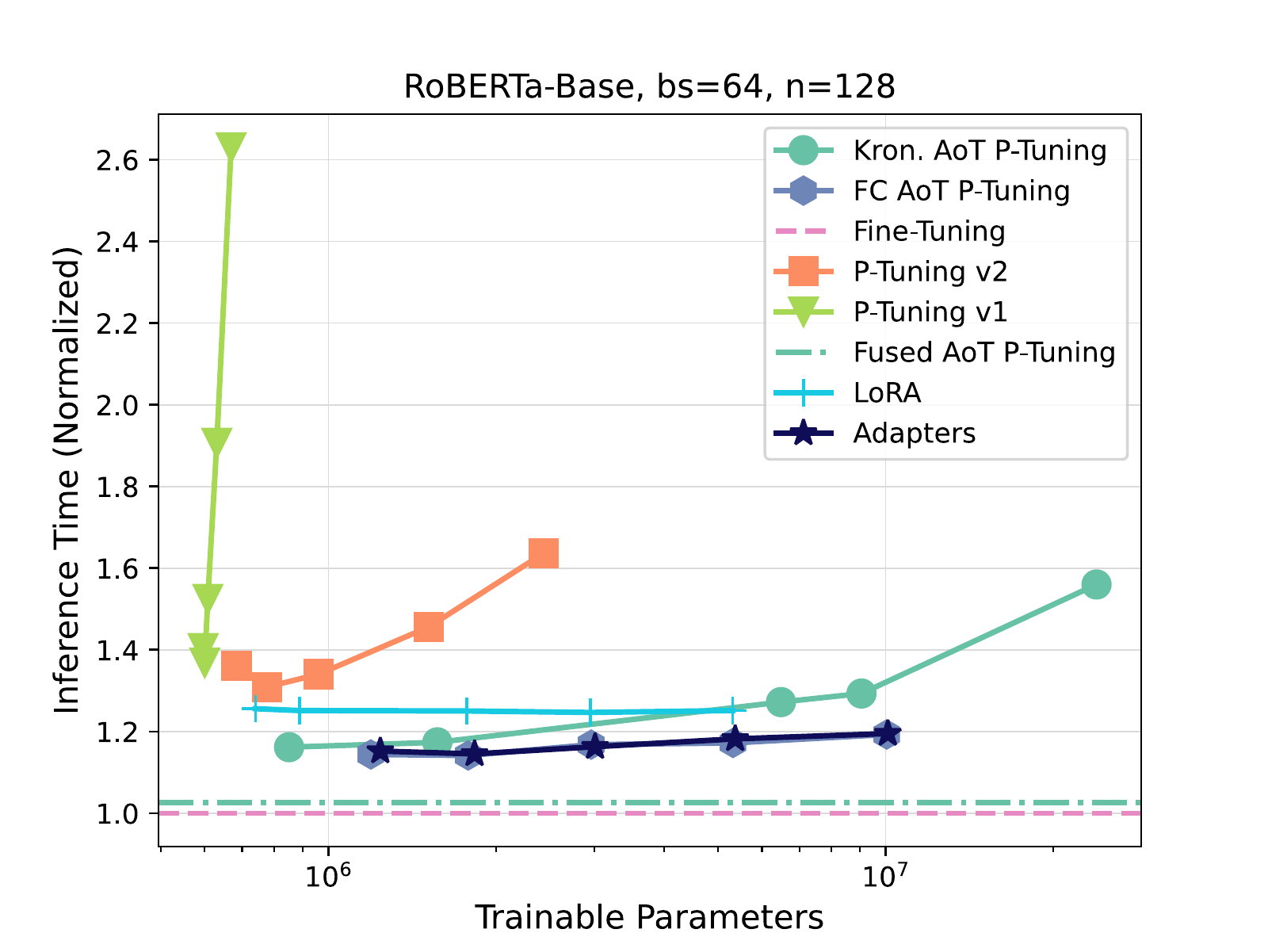}
    \caption{}
  \end{subfigure}
        \begin{subfigure}[t]{.29\linewidth}
    \centering\includegraphics[width=\linewidth]{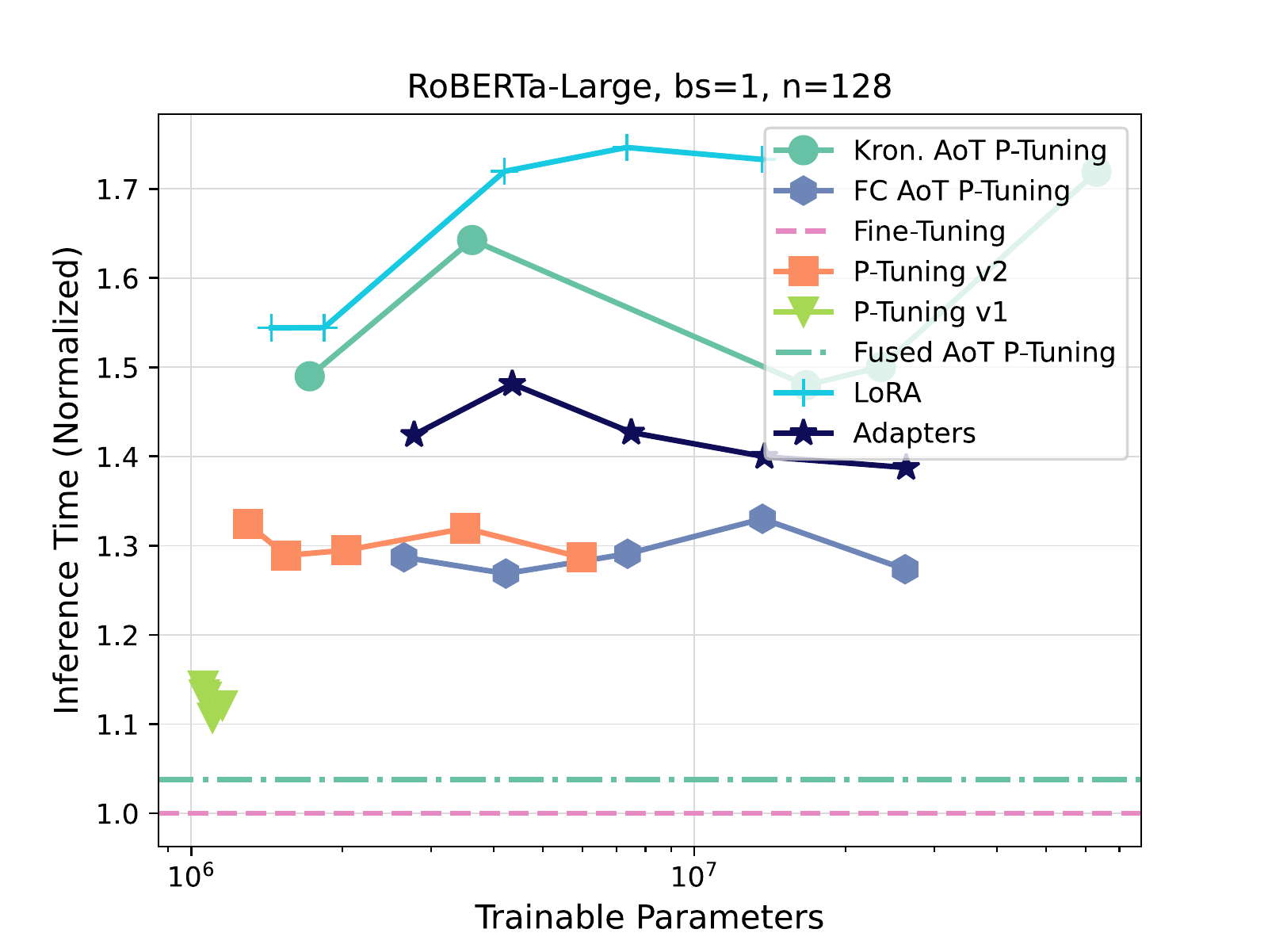}
    \caption{}
  \end{subfigure}
    \begin{subfigure}[t]{.29\linewidth}
    \centering\includegraphics[width=\linewidth]{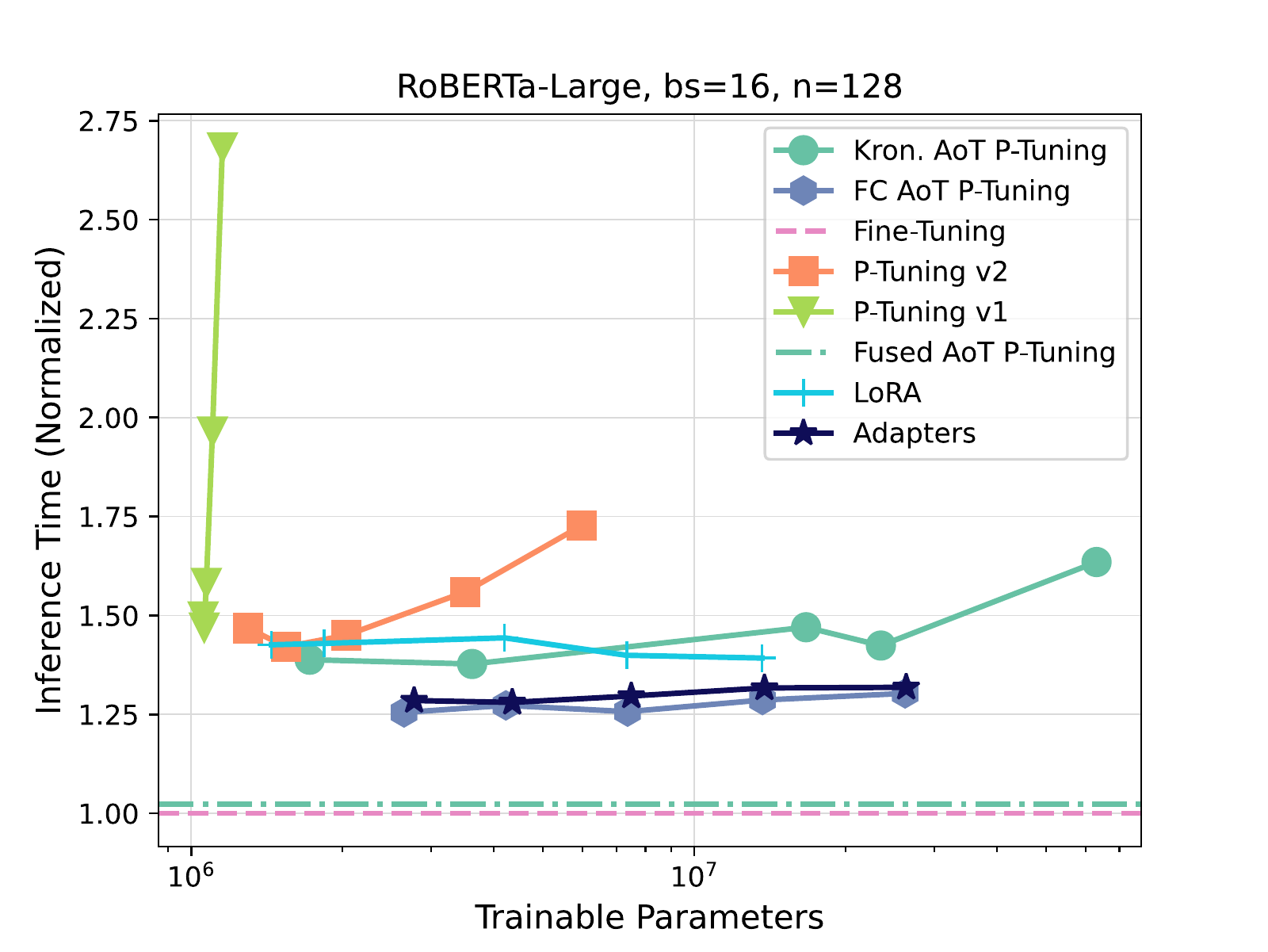}
    \caption{}
  \end{subfigure}
        \begin{subfigure}[t]{.29\linewidth}
    \centering\includegraphics[width=\linewidth]{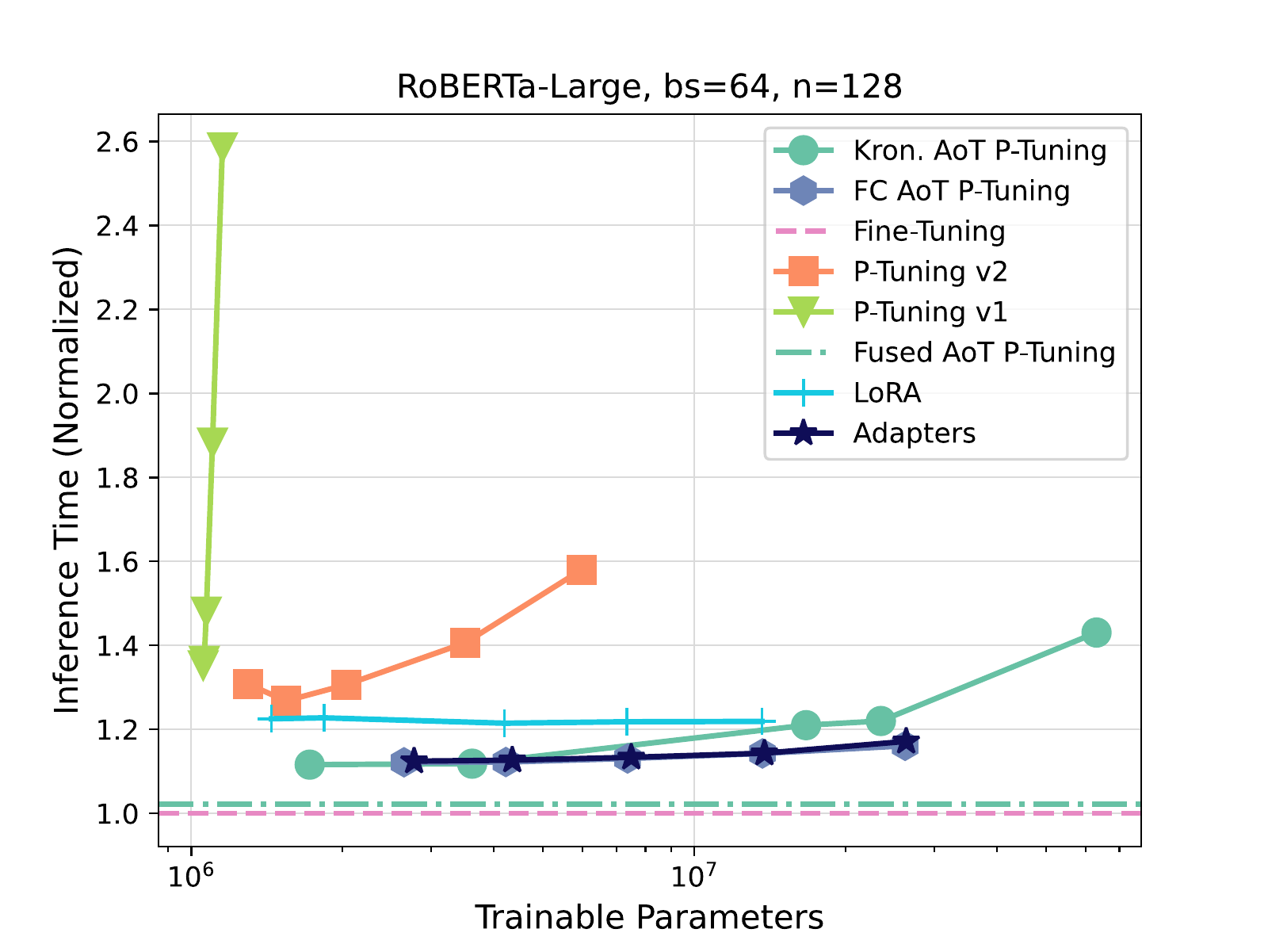}
    \caption{}
  \end{subfigure}
        \begin{subfigure}[t]{.29\linewidth}
    \centering\includegraphics[width=\linewidth]{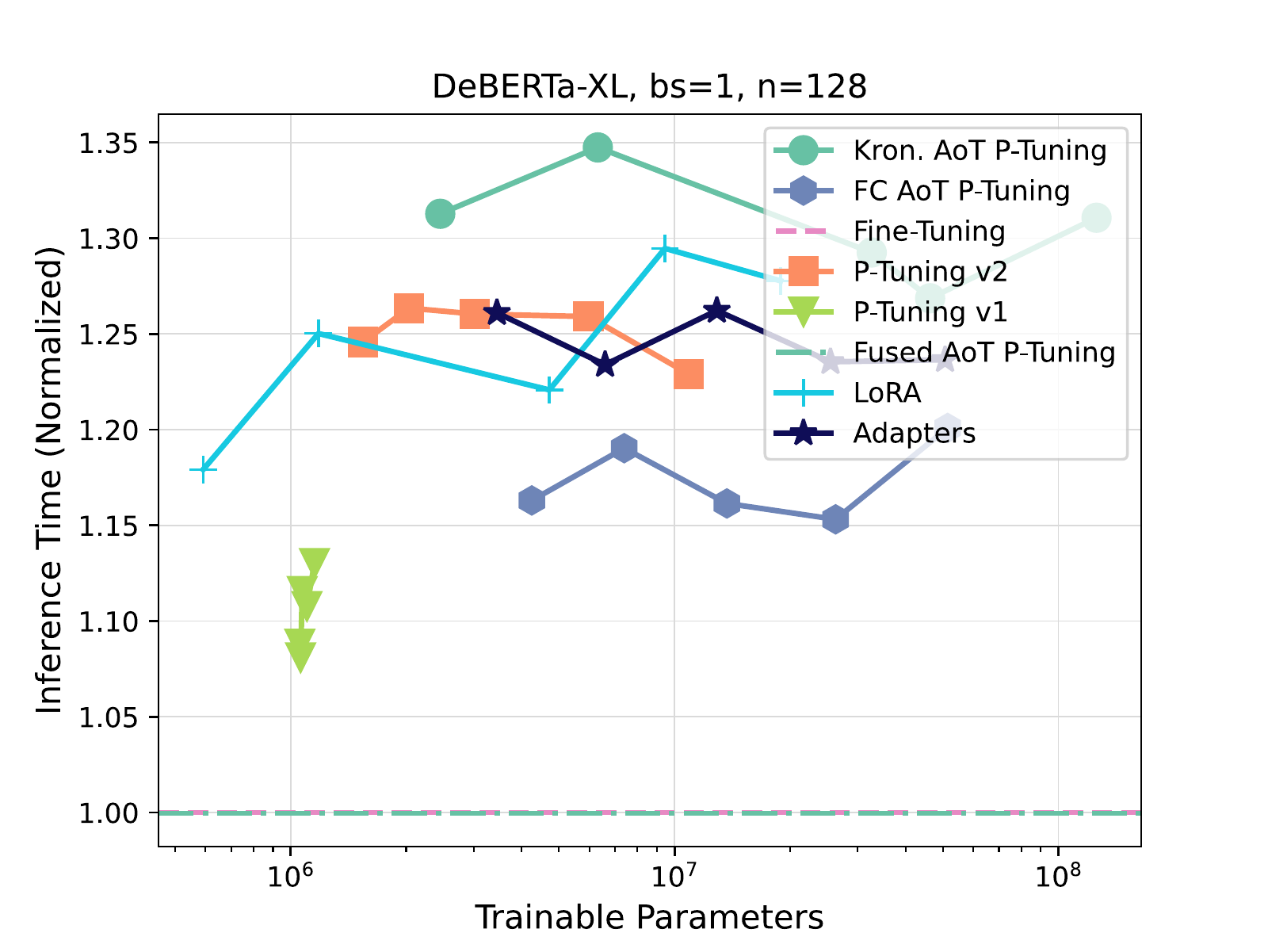}
    \caption{}
  \end{subfigure}
    \begin{subfigure}[t]{.29\linewidth}
    \centering\includegraphics[width=\linewidth]{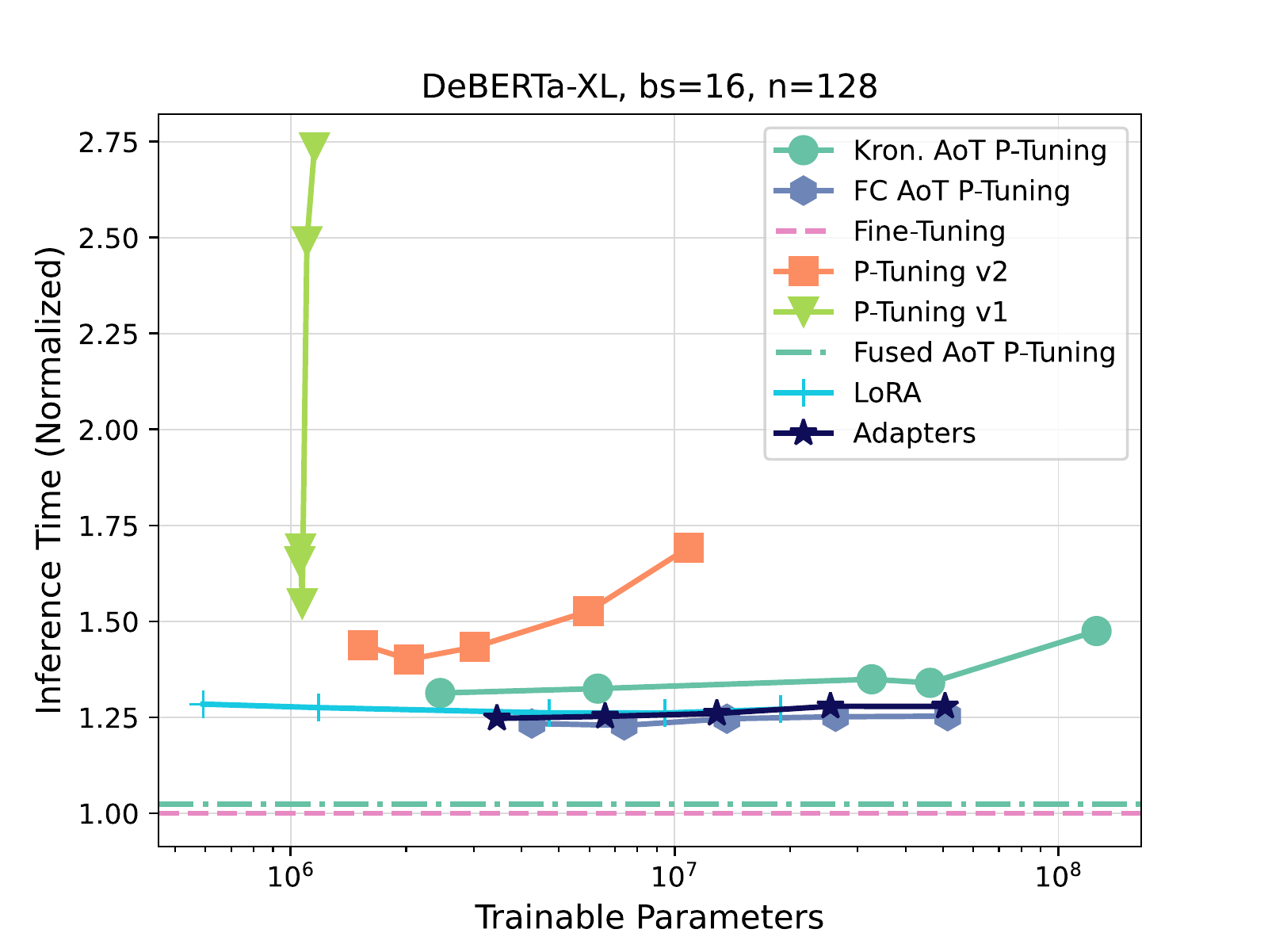}
    \caption{}
  \end{subfigure}
        \begin{subfigure}[t]{.29\linewidth}
    \centering\includegraphics[width=\linewidth]{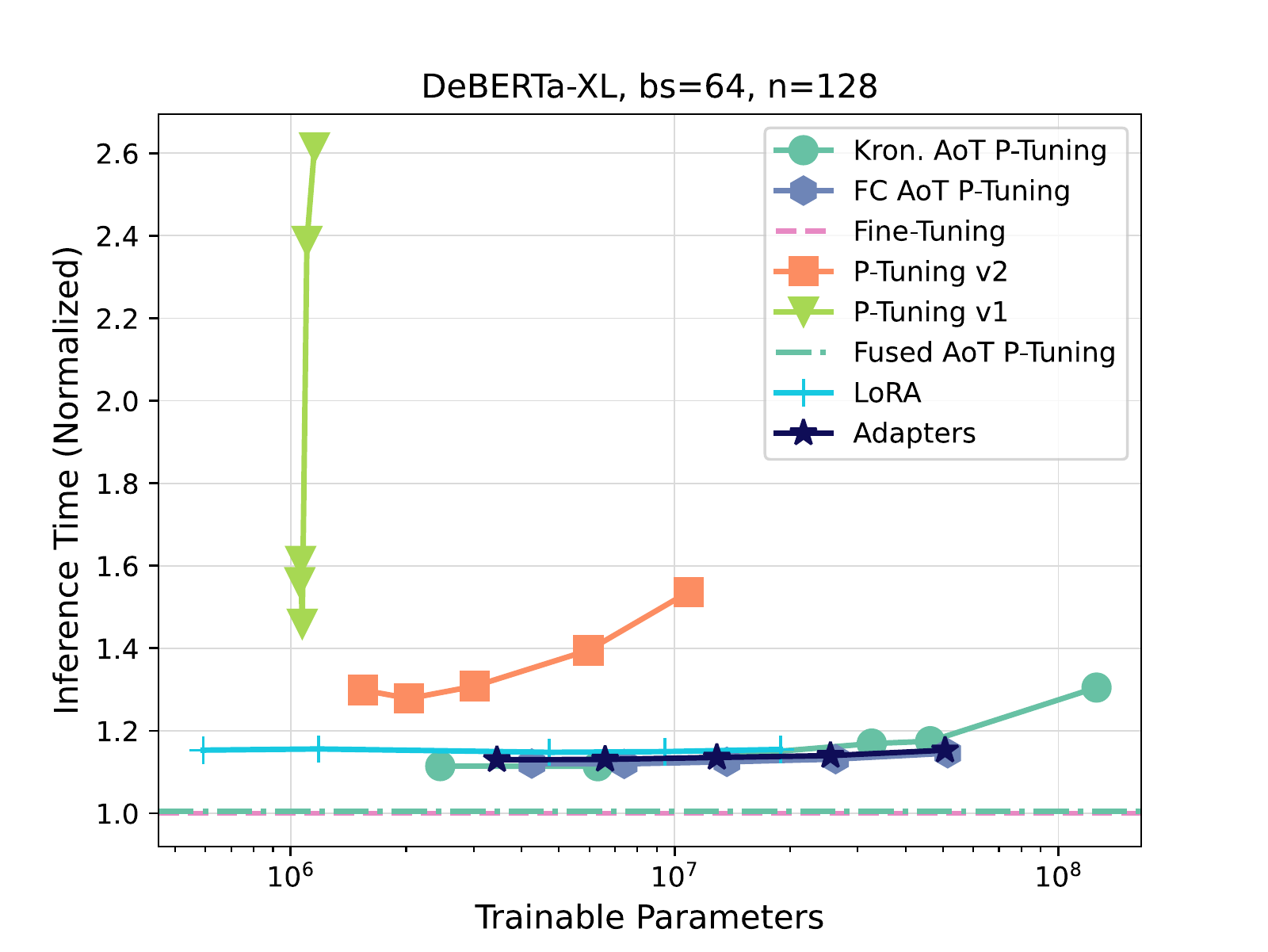}
    \caption{}
  \end{subfigure}
  
  \caption{Speed measurement for baseline methods with sequence length $\in [16, 64]$ for different backbone models. See Section 4.4 for details.}
  \label{appendix-figure-speed}
\end{figure*}

\begin{table*}[]
\begin{tabular}{r|l}
\toprule
$l$\# & Tokens $x$ with largest norm $||\mP_x||_2$                                                           \\
\toprule
0 & \makecell[l]{likes, a, is, loves, was, to, as, wants, s, ., pony, \\ eded, himself, Man, were, and, I, has, I, are, Frank, ., hates, \\ As, A, A, like, It, crop, Frank, After, ,, joins, As, Eric, \\ Likes, It, just, would, onna, him, To, behaving, after, in, because, behaves, \\ Is, We, Like}\\
\midrule
5 & \makecell[l]{,, ., narrower, doorway, backdoor, window, lousy, shortest, nicer, checkpoint, knob, \\ thinner, narrowing, oub, quieter, BAD, ;, VID, rectangle, tighter, crappy, intruder, tongues, \\ fing, rimination, blocker, and, raiding, detector, unmarked, sharper, knife, coolest, thicker, hoops, \\ DOWN, lightsaber, asshole, millisec, KEY, sharp, token, slashing, Defenders, jug, Donna, slider, \\ wedge, dding, kb}\\
\midrule
10 & \makecell[l]{her, Her, herself, him, above, she, out, hers, him, HER, She, \\ she, care, HER, above, Her, bold, CARE, cared, over, harder, louder, Above, \\ smarter, sooner, her, cares, better, Out, vind, stronger, She, taller, tougher, Him, \\ ahead, so, HIM, Susan, happier, up, Harry, aloud, higher, Above, SHE, could, \\ apart, barking, inem}\\
\midrule
22 & \makecell[l]{., there, dry, for, There, Her, her, sword, the, arse, wy, \\ dry, duc, The, it, took, cr, Rig, og, There, landing, the, wide, \\ centrally, red, grass, sw, oa, above, engine, FT, spir, cd, Coun, Ross, \\ there, ws, guy, starter, mans, aniel, green, freely, d, wide, stall, far, \\ artz, THERE, didn}\\
\midrule
32 & \makecell[l]{it, me, olit, Polit, Pat, him, Private, Susan, pat, he, her, \\ Self, Ins, Doc, Coun, Ang, Aut, Sil, ochond, me, Nob, IT, Senator, \\ Professional, Dri, itized, Je, Capt, Hillary, Whe, He, Kid, Registered, itious, Michelle, \\ Political, It, Shut, Phot, BIT, Politics, Bit, Jacob, ruct, Young, HE, Tu, \\ them, Mot, itu}\\
\midrule
37 & \makecell[l]{him, they, it, they, her, them, their, his, he, it, its, \\ hers, theirs, was, he, Susan, old, older, THEY, They, ITS, forth, Georg, \\ Thom, Tom, erved, Carl, Anna, nob, anos, itans, to, Eric, itcher, Harry, \\ Tim, Jen, them, Kid, Jeremy, JOHN, Jennifer, hands, Todd, put, Thomas, she, \\ Dan, Michelle, s}\\
\midrule
46 & \makecell[l]{chased, erved, house, houses, life, chasing, market, self, chase, raised, chester, \\ hunt, castle, HOU, atics, Singer, western, ogenous, rounded, stretched, esian, essed, omorphic, \\ horse, SER, central, ledge, hole, asio, Self, Self, iverse, oker, Judd, DF, \\ aday, paced, ourced, erness, Barkley, scape, sey, ationally, owned, landers, ded, study, \\ directed, OWN, produced}\\

\toprule
\end{tabular}
\caption{Tokens with the largest $L_2$ norm of $\mP$ entries for the WSC task. See Section 4.3 for more details.}
\label{wsc}
\end{table*}

\begin{table*}[]
\begin{tabular}{r|l}
\toprule
$l$\# & Tokens $x$ with largest norm $||\mP_x||_2$                                                           \\
\toprule
0 & \makecell[l]{fit, Loud, as, Air, Upon, Sets, Bound, Apart, scratched, sets, fit, \\ Upon, hosted, Shot, Unt, Host, fitt, Sight, atri, Ocean, ceed, ashore, set, \\ enture, underwent, planes, boats, Waves, Ali, shi, Active, Set, Atmosp, Airways, Host, \\ chat, Endless, pelled, rew, ached, unct, fitted, Proud, flu, itable, anson, Bound, \\ Assets, host, sets}\\
\midrule
5 & \makecell[l]{set, Set, sets, Set, SET, Setting, setting, SET, set, padd, Setting, \\ Sets, sets, bed, Cause, setting, the, cause, tread, itch, paddle, cause, thirsty, \\ Khe, he, anned, this, ?, of, Cause, What, bidding, This, This, what, \\ What, a, his, lic, The, wish, fugitive, they, Bed, Air, wake, conscience, \\ ., crowd, Let}\\
\midrule
10 & \makecell[l]{?, ?, ?!, ??, ., ?", "?, )?, ???, '?, set, \\ .?, !?, ?), !, ????, ...?, set, ?', ,, Set, Set, ed, \\ to, ??, ????????, as, ?????, ?)., setting, ?,, ?'", ?], sets, ..., \\ -, lt, —, :, lic, ???, led, ur, …, punching, of, t, \\ ?"., sets, um}\\
\midrule
22 & \makecell[l]{What, set, What, out, Set, sets, on, to, '?, what, Set, \\ in, WHAT, Sets, Setting, WHAT, from, Setting, dropped, of, Dig, Got, ?', \\ set, Exper, Gets, Ground, ...?, happened, Whatever, Your, decom, Getting, Got, overlooked, \\ Crack, )?, He, police, !?, happens, Suc, sets, what, Detective, GOT, Whatever, \\ SET, Getting, Flying}\\
\midrule
32 & \makecell[l]{glued, hid, melted, ., sent, breaths, etz, breath, Breath, baptized, watch, \\ putting, tongues, braces, put, hid, bleach, icating, burying, aver, lifting, Illuminati, orneys, \\ melting, withdrawing, numb, radios, inserts, amins, avert, breathing, puts, informants, lifting, hide, \\ conscience, recommending, withdrawn, ransom, catch, Gael, Vern, roth, ears, Put, gins, breathed, \\ attorneys, loss, biblical}\\
\midrule
37 & \makecell[l]{hid, dropped, raided, fought, ungle, Hide, destruct, smuggled, abandoned, looted, attacked, \\ barric, slid, dodged, drop, shut, drowned, hide, destruct, buggy, battled, shutdown, Hide, \\ Attack, hid, rawl, inaccessible, avalanche, slipped, deleted, rawling, encrypted, withdrawn, Killed, dug, \\ dropping, hoard, weapon, swallowed, defensive, destroy, exited, destroy, fight, Fighting, lost, deny, \\ suppress, encrypt, aggressively}\\
\midrule
45 & \makecell[l]{hopped, chats, pumped, paints, backed, spun, tread, coached, reefs, privately, noodles, \\ buddies, malls, whisper, endorsements, squeezed, pals, blush, comed, edits, rallies, gigs, recol, \\ mocked, curs, Bare, bubbles, warmed, chat, profiles, emails, Dreams, pads, chalk, interviewed, \\ sneakers, rocked, Gloves, hubs, docs, shaved, Rise, primaries, listened, shy, essays, whispers, \\ leeve, girlfriends, socks}\\

\toprule
\end{tabular}
\caption{Tokens with the largest $L_2$ norm of $\mP$ entries for the COPA task. See Section 4.3 for more details.}
\label{copa}
\end{table*}

\begin{table*}[]
\begin{tabular}{r|l}
\toprule
$l$\# & Tokens $x$ with largest norm $||\mP_x||_2$                                                           \\
\toprule
0 & \makecell[l]{gression, rium, History, orer, aic, history, oration, ré, orative, amic, history, \\ version, ural, osa, avage, ory, lia, range, History, rica, nation, root, USE, \\ á, ination, ulation, mentation, issance, state, rum, adal, idden, jection, oly, ó, \\ esis, orean, discovery, ria, ada, uration, entry, ord, verse, inations, ugal, itus, \\ olics, ESSION, ativity}\\
\midrule
5 & \makecell[l]{., to, in, for, and, of, ,, s, the, be, 's, \\ by, on, or, from, at, or, with, :, ly, a, an, ;, \\ on, -, ., in, under, an, as, I, and, !, about, er, \\ In, but, ?, A, is, ed, a, that, o, ers, S, ing, \\ now, ), -}\\
\midrule
10 & \makecell[l]{., of, and, for, morph, votes, elector, with, uild, igraph, tatt, \\ Assignment, as, contribut, advant, are, hod, Voters, matically, Init, rede, olon, on, \\ rehabilit, neum, mog, looted, req, by, Claim, the, ynchron, dule, promot, socio, \\ portfolios, goto, vulner, vote, setup, nominate, anism, s, subscrib, iop, lihood, slot, \\ elist, ramid, ysc}\\
\midrule
22 & \makecell[l]{in, In, in, be, In, .", being, Straw, -, its, a, \\ Majority, of, a, Latest, the, Jack, ine, latest, it, Lawyers, Watts, "., \\ "-, Massachusetts, their, .', been, ure, Till, '.", Signs, .'", Seventh, ?", \\ Taxes, Atlanta, !", electric, at, IN, ide, Current, Ladies, KP, Jersey, Students, \\ Knights, it, Anders}\\
\midrule
32 & \makecell[l]{Se, Hum, Brazil, Mur, Hur, aver, Hum, Yugoslavia, Mour, jud, a, \\ Hawai, Pag, Kant, ibal, Malaysia, EFF, Hur, .", adj, mur, Islam, and, \\ Guinea, Britain, Sadd, Def, Niger, ,, Holland, amus, Hay, Ma, Appro, Mur, \\ Countries, Wid, Asians, Nor, else, Calendar, Hed, Ved, ldom, english, Hind, mur, \\ bury, Ded, hol}\\
\midrule
37 & \makecell[l]{[SEP], +., Sk, Ble, Gre, cloud, Else, ., +,, "., \\ uran, cs, Ever, 2048, Ble, Keefe, Hyp, athan, Lib, Fra, Exp, \\ bro, Edit, Ros, Bean, Bo, Beck, Shell, sit, !., Saud, Phys, -, \\ shell, Ol, BLIC, ‑, Over, ea, orthy, Shot, pn, pas, ester, Reviewed, \\ Spe, sell, 2024}\\
\midrule
45 & \makecell[l]{Chance, Sw, chance, Nine, Shares, Chance, Scientists, Tw, Besides, Prof, chances, \\ Sn, sw, TW, EFF, J, IJ, Besides, chance, Between, icist, GU, SW, \\ pan, Ja, Psy, tw, Between, xon, Bj, Conj, Shares, Moh, UTH, Prediction, \\ science, intend, Science, iov, Nine, jp, dds, NJ, Jr, y, Nin, etsy, \\ Ibid, ymm, Reporting}\\
\toprule
\end{tabular}
\caption{Tokens with the largest $L_2$ norm of $\mP$ entries for the RTE task. See Section 4.3 for more details.}
\label{rte}
\end{table*}

\begin{table*}[]
\begin{tabular}{r|l}
\toprule
$l$\# & Tokens $x$ with largest norm $||\mP_x||_2$                                                           \\
\toprule
0 & \makecell[l]{.'', didn, doesn, don, ''., ,'', Didn, Doesn, didn, doesn, Don, \\ wouldn, Wouldn, Does, '', couldn, DON, Did, hadn, But, Don, Isn, ).", \\ DON, shouldn, ``, Obviously, Obviously, Isn, don, hasn, ))., Does, "?, ].", \\ wasn, Did, ],", ,., Naturally, ...", ),", Would, ``, But, '';, Naturally, \\ ]., ,), DOES}\\
\midrule
5 & \makecell[l]{,, 't, !,, .,, ?,, didn, , not, to, +,, *,, \\ ),, the, ., ],, considered, doesn, in, , ,), a, /,, ,[, \\ you, don, ,,, ,., shouldn, (),, hasn, ;, for, thought, weren, hadn, \\ thought, wasn, NOT, hair, ', .;, aren, `,, Said, ,, couldn, \\ isn, .—, idered}\\
\midrule
10 & \makecell[l]{Shant, Georg, Expect, Led, Assistant, Amph, Registered, Ear, McA, THEIR, Prev, \\ Emb, -,, Called, Gw, Alc, Until, Rhod, Introduced, that, Lat, Unt, Ul, \\ Sv, Gh, to, of, Fernand, ,, elta, jac, unch, Ov, Sebast, apologised, \\ JOHN, !"., Ll, hid, Somewhere, Been, Recently, and, Somebody, Fram, Coh, ')., \\ Sty, Elsewhere, Unt}\\
\midrule
22 & \makecell[l]{'s, 're, A, 've, the, a, her, ', be, A, a, \\ have, DOES, LIKE, ?", "?, 'd, )?, ?, s, ABOUT, ", Like, \\ Pant, didnt, 'm, '?, E, The, doesnt, Was, re, :, ie, \\ Surely, 'll, Corinth, At, Across, your, their, ?,, THEY, ...?, or, \\ Fra, HOW, )/}\\
\midrule
32 & \makecell[l]{I, '?, he, "?, ?', )?, 't, ''., .?, ...?, .'', \\ ?"., ':, He, !?, ?,, He, I, and, ?'", ?!", ?", ?)., \\ !', he, .:, ?!, she, +., )!, '., .', !?", \\ ,'', ')., ???, !., ).", we, CLOSE, `., "!, .], .--, ????, \\ '/, 're, .'"}\\
\midrule
37 & \makecell[l]{., 's, 't, ?, ', :, I, -, ', ,, of, \\ he, B, B, in, I, and, 'm, -, s, ", 'd, by, \\ for, ;, b, on, you, !, ", He, to, /, 've, y, \\ 're, ed, with, ., 'll, a, back, the, b, she, \\ He, E, C}\\
\midrule
45 & \makecell[l]{'t, not, NOT, the, not, Not, never, Not, 's, 're, NOT, \\ 've, you, of, [SEP], t, nt, Never, The, in, NEVER, he, to, \\ the, [CLS], hardly, never, neither, I, ,, 'm, cannot, no, The, annot, \\ it, Their, me, didnt, He, and, doesnt, Ear, a, ., Never, none, \\ if, on, nobody}\\
\toprule
\end{tabular}
\caption{Tokens with the largest $L_2$ norm of $\mP$ entries for the CB task. See Section 4.3 for more details.}
\label{cb}
\end{table*}

\end{document}